%% file: main.tex
\documentclass[10pt]{article} 
\usepackage{graphicx}

\usepackage[accepted]{tmlr}

\input{math_commands.tex}

\usepackage{hyperref}
\usepackage{url}

\usepackage{booktabs}
\usepackage{makecell}
\usepackage{multirow}
\usepackage{tabularx}
\usepackage{adjustbox}
\usepackage{wrapfig}
\usepackage{float}
\usepackage{pifont}
\usepackage{fontawesome5}
\usepackage{siunitx}
\usepackage[table]{xcolor}
\definecolor{customyellow}{RGB}{255,245,157}
\definecolor{homepageblue}{HTML}{1E88E5}
\definecolor{githubblack}{HTML}{181717}
\usepackage{subcaption}

\title{LibMoE: A Library for Comprehensive Research on Mixture of Experts in Large Language Models}

\author{\name Nam V. Nguyen$^{*,\diamond}$ \email namnguyen16ai@gmail.com \\
      \addr FPT Software AI Center
      \AND
      \name Thong T. Doan$^*$ \email doantienthongbku@gmail.com \\
      \addr FPT Software AI Center
      \AND
      \name Luong Tran \email luongtk@fpt.com\\
      \addr FPT Software AI Center
      \AND
      \name Van Nguyen \email vannth19@fpt.com \\
      \addr FPT Software AI Center
      \AND
      \name Quang Pham$^\diamond$ \email hqpham.2017@phdcs.smu.edu.sg \\
      \addr Independent Researcher}

\newcommand{\projectlink}[4]{%
  \href{#1}{%
    \sffamily\begin{tabular}{@{}c@{}}
      {\Large\textcolor{#2}{#3}}\\[-1pt]
      \textcolor{#2}{\textbf{#4}}
    \end{tabular}%
  }%
}

\def\month{06}  
\def\year{2026} 
\def\openreview{\url{https://openreview.net/forum?id=PB2ju8tq0n}}

\begin{document}

\maketitle
\begingroup
\renewcommand{\thefootnote}{\fnsymbol{footnote}}
\footnotetext[1]{Equal contribution.}
\renewcommand{\thefootnote}{\ensuremath{\diamond}}
\footnotetext[2]{Corresponding author.}
\endgroup

\begin{abstract}
Mixture-of-experts (MoE) architectures have become a cornerstone for scaling up and are a key component in many recent large language models such as GPT-OSS, DeepSeek-V3, Llama-4, and Gemini-2.5. However, systematic research on MoE remains severely constrained by the prohibitive computational costs of training and evaluation, limiting access to large-scale studies for most researchers. We introduce LibMoE, a unified framework for reproducible, efficient, and extensible MoE research that supports both pretraining and sparse-upcycling regimes. Beyond unified implementations, the framework provides transparent analytical tools for probing routing and expert dynamics. Leveraging this foundation, we conduct a comprehensive analysis along three dimensions: (i) routing dynamics, covering expert selection patterns, routing stability and optimality, and how routing entropy reveals task specialization and expert diversity; (ii) the effect of lightweight initialization on load balancing, demonstrating how subtle changes in router initialization shape early expert utilization; and (iii) training regime differences, revealing how sparse upcycling and full pretraining exhibit distinct routing patterns and stability profiles. By lowering the barrier to entry and standardizing evaluation, along with our comprehensive analysis, LibMoE broadens access to MoE research and establishes a reliable benchmark to guide future innovations. 
\end{abstract}

\begin{center}
\small
\projectlink{https://fsoft-aic.github.io/fsoft-LibMoE.github.io/}{homepageblue}{\faIcon{home}}{Homepage}
\hspace{0.75em}
\projectlink{https://github.com/Fsoft-AIC/LibMoE}{githubblack}{\faGithub}{GitHub}
\end{center}

\input{contents/1_Introduction/introduction}
\input{contents/2_Related_works/related_works}

\input{contents/3_Desining_LibMoE/designing_libmoe}

\input{contents/4_Training_LibMoE/training_libmoe}
\input{contents/5_Analysis/analysis}
\input{contents/6_Rethinking_LibMoE/rethinking_libmoe}
\input{contents/7_Conclusion/concusion}

\bibliography{main}
\bibliographystyle{tmlr}

\appendix
\input{contents/A_add_exps/additional_exp}

\end{document}

%% file: math_commands.tex
\usepackage{amsmath,amsfonts,amssymb,bm}

\def\eqref#1{equation~\ref{#1}}

\def\1{\bm{1}}

\def\vs{{\bm{s}}}

\def\vx{{\bm{x}}}

\DeclareMathAlphabet{\mathsfit}{\encodingdefault}{\sfdefault}{m}{sl}
\SetMathAlphabet{\mathsfit}{bold}{\encodingdefault}{\sfdefault}{bx}{n}

\def\gD{{\mathcal{D}}}

\def\gR{{\mathcal{R}}}



%% file: contents/1_Introduction/introduction.tex
\section{Introduction}
\label{sec:introduction}

Recent years have witnessed a dramatic surge in the adoption and success of deep learning models, fueled by their scalability with massive data and compute resources and their state-of-the-art performance across diverse domains. Yet, this rapid scaling has introduced significant challenges in computational efficiency. A prominent strategy to address these challenges is the sparse mixture of experts (SMoE) architecture, which expands model capacity without a proportional increase in computation. 

First introduced by \citep{jacobs1991adaptive}, SMoE has since achieved remarkable success across multiple fields, including large language models (LLMs) \citep{shazeer2017outrageously, fedus2022switch, comanici2025gemini, liu2024deepseekv3, jiang2024mixtral}, multimodal learning \citep{yun2024flex, lin2024moe, han2024fusemoe, meta2025llama4}, and computer vision \citep{riquelme2021scaling, fan2022m3vit, han2024vimoe}. By activating only a sparse subset of parameters per input, SMoE substantially improves training efficiency, enabling models with hundreds of billions of parameters to be trained at manageable computational cost \citep{fedus2022switch}. Interestingly, SMoE algorithms often achieved superior performances compared to their dense counterpart \citep{shazeer2017outrageously}, which has gathered an increased interest in the community to develop more advanced algorithms and better theories to understand their behaviour \citep{muennighoff2024olmoe,xue2024openmoe,kang2025flame,zoph2022st,nguyen2025deepseekmoe,nguyen2024sigmoid}. Consequently, SMoE has become a cornerstone methodology driving the development of next-generation intelligent systems. 

However, despite rapid algorithmic advances, progress in MoE research remains fundamentally hindered by the lack of standardized frameworks. Large-scale MoE experiments are typically feasible only for groups with access to massive computational resources: training OLMoE-1B-7B, for example, required 256 H100 GPUs \citep{muennighoff2024olmoe}, and MOMA relied on 256 A100 GPUs \citep{lin2024moma}. In contrast, the majority of researchers are confined to small-scale studies \citep{nielsen2025tight,han2024fusemoe,teo2024momentumsmoe,csordas2023approximating,csordas2024moeut,nguyen2025camex,teo2025molex} or even synthetic benchmarks \citep{nguyen2024sigmoid,nguyen2024least,yan2024understanding}, resulting in fragmented and often incomparable results. This persistent discrepancy severely undermines the potential of MoE architectures, whose true strengths emerge only in large-scale training regimes. As a result, even prominent deployed models such as Phi-3 \citep{abdin2024phi3tr}, Skywork-MoE \citep{wei2024skyworkmoe}, and GPT-OSS \citep{openai2025gptoss120bgptoss20bmodel} tend to rely on conventional SMoE designs \citep{fedus2022switch}, with recent advances in routing and expert selection algorithms seeing little adoption in practice \citep{do2023hyperrouter,zhou2022_expert_choice_routing,dai2022stablemoe,wang2024remoe}. These challenges underscore a critical need for a unified, reproducible framework that lowers the barrier to meaningful SMoE research under realistic resource constraints and enables rigorous, systematic evaluation of both established and emerging methods.

To address these challenges, we present LibMoE, a unified framework that brings together state-of-the-art SMoE algorithms, standardized training pipelines, and transparent analytical tools for rigorous investigation of routing dynamics. Our primary objectives are twofold: (1) to provide an accessible and streamlined toolkit that enables meaningful SMoE research on large language models (LLMs) under realistic resource constraints; and (2) to enable comprehensive, reproducible analysis of SMoE routing mechanisms, yielding deeper insights into expert behaviors. With LibMoE, we aim to accelerate progress and foster a more collaborative, open-source SMoE research community.

\textbf{Designing the LibMoE Framework.}
To address persistent challenges in accessible and reproducible MoE research, we introduce LibMoE, a unified framework for systematic evaluation of SMoE methods across both full pretraining and sparse-upcycling regimes \citep{komatsuzaki2022sparse}. LibMoE accommodates a broad spectrum of model sizes and algorithmic variants, supporting rigorous comparisons from small-scale language models (0.15B and 0.68B parameters) to large-scale vision–language models (5.67B parameters). The framework integrates seven recent state-of-the-art SMoE algorithms and provides standardized pipelines for both pretraining and evaluation, enabling consistent benchmarking and cross-domain analysis. All MoE algorithm experiments are conducted under realistic resource constraints, with training times ranging from \colorbox{yellow!25}{6 to 44 hours on $4 \times$ H100 GPUs} (see Table~\ref{tab:time_and_resource} for details). While our results reveal that current SMoE algorithms yield only marginal performance improvements under these settings, LibMoE substantially lowers the barrier to entry, making advanced and reproducible experimentation feasible even for research groups with limited computational resources.

\textbf{Comprehensive Analysis of Mixture-of-Experts Dynamics.}
Beyond providing unified implementations, LibMoE is designed as a transparent and extensible infrastructure for systematically analyzing the behavior of SMoE models. The framework exposes modular instrumentation for monitoring routing decisions, expert utilization, load balancing, and inter-expert interactions, enabling fine-grained inspection of routing behavior throughout both training and inference. Using this analytical capability, we conduct a large-scale, cross-domain empirical study of seven recent SMoE algorithms across multiple training regimes, including small-scale pretraining (0.15B and 0.68B parameters) and sparse upcycling at the 5.67B scale~\citep{komatsuzaki2022sparse}. Our analysis uncovers consistent and previously under-characterized differences along three dimensions: (i) routing dynamics, reflected in expert selection stability, entropy evolution, and specialization patterns; (ii) the impact of lightweight router initialization on early-stage load balancing; and (iii) systematic divergences in routing behavior between full pretraining and sparse upcycling regimes. Together, these results demonstrate how design choices in routing and initialization influence training stability and expert utilization, and establish a reproducible analytical basis for principled comparison of SMoE algorithms.

%% file: contents/2_Related_works/related_works.tex
\section{Related Works}
\label{sec:related_work}

\subsection{Mixture of Experts}

Originally introduced by \citet{jacobs1991adaptive}, the MoE framework was first formulated as an ensemble method that combines multiple specialized models through an adaptive gating network. Early variants, including Jacobs’ gating network and the Hierarchical MoE (HME) \citep{jordan1994hierarchical}, improved learning algorithms for multiclass classification \citep{chen1999improved}, and achieved success in domains such as speech recognition \citep{gales2006product}. \cite{eigen2013learning} extends the MoE to a layer in neural network, which consists of a set of experts (neural networks) and a trainable gate. These early efforts collectively shaped the theoretical and empirical foundation of MoE, setting the stage for the large-scale sparse MoE architectures that gained prominence after 2017. In 2017, \cite{shazeer2017outrageously} introduced the SMoE model, which incorporated MoE layers into long short-term memory (LSTM) networks to dramatically scale model capacity without increasing computational cost. Building on this idea, GShard \citep{lepikhin2020gshard} extended MoE to Transformers by replacing the feed-forward network (FFN) layers in the T5 \citep{raffel2020exploring} architecture with MoE layers. Switch Transformers \citep{fedus2022switch} further simplified the routing mechanism and scaled training to 1.6 trillion parameters, setting a milestone for large language models (LLMs). Since then, substantial progress has been made in advancing sparsely-gated MoE architectures for LLM development, including novel routing techniques \citep{chi2022representation,roller2021hash,zuo2021taming,zhong2024lory,muqeeth2023soft,wu2024yuanm32,wang2024remoe,nielsen2025tight,nguyen2025competesmoe,csordas2023approximating}, stability-oriented strategies \citep{zoph2022st,dai2022stablemoe}, fine-grained expert segmentation \citep{park2024monet,dai2024deepseekmoe,he2024mixture}, shared experts \citep{rajbhandari2022deepspeed,liu2024deepseekv2,liu2024deepseekv3,dai2024deepseekmoe,qwen_moe}, special-purpose experts \citep{jin2024moe++,yan2025tc}, gradient update modification \citep{panda2025dense,yang2024solving}, and sparse upcycling \citep{komatsuzaki2022sparse,nakamura2025drop,chen2025automatic,hui2024upcycling}; with extensions also demonstrated in multimodal learning \citep{shen2023scaling,han2024fusemoe,li2024cumo,lin2024moe,wu2024deepseek,yun2024flex}. Despite the encouraging progress, a clear discrepancy remains in current MoE research. In particular, despite a proliferation of advanced algorithms and theoretical analyses, the most capable and widely deployed LLMs~\citep{openai2025gptoss120bgptoss20bmodel, comanici2025gemini, liu2024deepseekv2, liu2024deepseekv3, dai2024deepseekmoe, xai2024grok1, yang2024qwen2tr, jiang2024mixtral, meta2025llama4, he2024upcycling, abdin2024phi3tr, dbrx_2024, wu2024yuanm32} still rely on SMoE variants closely aligned with the original formulation~\citep{shazeer2017outrageously, fedus2022switch}. This indicates that our current theoretical understanding remains largely confined to academic settings and has yet to meaningfully influence real-world deployments precisely the domain where SMoE could be most impactful. We attribute this gap primarily to the high entry barrier posed by massive datasets and computational requirements, which place large-scale SMoE experimentation out of reach for most research groups. Therefore, we develop LibMoE to enable training and evaluation settings that closely reflect real-world practice, including both early-stage pretraining and late-stage sparse upcycling under limited resources. This design allows for fair, comprehensive, and large-scale benchmarking of SMoE algorithms, while making state-of-the-art methods more accessible to the broader research community.

\subsection{Analyzing and Understanding Mixture-of-Experts Models}

Since the introduction of the sparsely-gated MoE layer~\citep{shazeer2017outrageously}, research has explored both the promise of conditional capacity and the challenge of expert imbalance. This has driven the development of stability-oriented methods (e.g., StableMoE~\citep{dai2022stablemoe}), novel routing designs (e.g., Expert-Choice~\citep{zhou2022_expert_choice_routing}), and system-level optimizations (e.g., MegaBlocks~\citep{gale2022_megablocks}). Differentiable routing mechanisms, including DSelect-k~\citep{hazimeh2021_dselectk}, BASE layers~\citep{lewis2021_base_layers}, and ReMoE~\citep{wang2024remoe}, have further increased flexibility, while AdaMoE~\citep{zeng2024_adamoe} and loss-free balancing~\citep{wang2024auxiliarylossfree} have enhanced efficiency and adaptability. Beyond algorithmic innovation, a growing body of theoretical and empirical work has examined MoE dynamics. For example,~\citet{zhao2024_sparse_moe_generalization} derived sparsity-aware generalization bounds, while~\citet{fan2024_moe_design_choices} systematically ablated routing granularity and expert count, revealing that token-level routers tend to capture syntactic rather than semantic patterns. Other studies have characterized router behavior (e.g.,~\citet{nguyen2024_cosine_router}) and highlighted linguistic specialization (e.g.,~\citet{antoine2024_pos_router_moe}), collectively connecting routing design to generalization, specialization, and stability. Efforts to democratize MoE research have yielded open-source frameworks such as OpenMoE~\citep{xue2024openmoe}, OLMoE~\citep{muennighoff2024olmoe}, and FLAME-MoE~\citep{kang2025flame}, which provide models, training frameworks, and diagnostic tools. However, the adoption of these resources is often limited by substantial computational requirements, restricting reproducibility and practical experimentation. Moreover, most prior studies focus on behaviors specific to their own MoE variants, rather than providing systematic cross-method comparisons. In contrast, our work aims to provide a unified and comprehensive empirical analysis of leading MoE algorithms under a standardized evaluation protocol. By systematically comparing a diverse set of methods and probing the effects of key design parameters, we offer deeper insights into the factors driving performance and specialization in modern MoE architectures.

\subsection{Mixture of Experts Toolkits}

Several open-source toolkits, including FastMoE~\citep{he2021fastmoe}, OpenMoE+t5x~\citep{xue2024openmoe}, and Tutel~\citep{hwang2023tutel}, support the implementation of SMoE algorithms. However, these frameworks present notable limitations for contemporary research. Tutel and OpenMoE+t5x are primarily designed for large-scale pretraining on hundreds of GPUs, restricting accessibility for groups with limited resources. FastMoE, while more lightweight, lacks support for recent LLM architectures and advanced distributed training libraries such as DeepSpeed~\citep{Lian2024UniversalCE}. In contrast, LibMoE is specifically designed to lower these barriers: researchers can train models with as few as $1$B tokens for vision–language tasks and $6$B tokens for language modeling representing a $1{,}000\times$ reduction in data requirements compared to OpenMoE while benefiting from support for modern LLMs and modular distributed training.

%% file: contents/3_Desining_LibMoE/designing_libmoe.tex
\section{Designing LibMoE}

\subsection{Preliminary: Mixture of Experts}
The standard SMoE layer~\citep{shazeer2017outrageously} consists of a router $\gR(\cdot, W_r)$, parameterized by $W_r$, and a set of $N$ experts $\{ g(\cdot, W_{e_i}) \}_{i=1}^N$, each with parameters $W_{e_i}$ for $i \in [N]$. Given an input token $\vx$, the router computes an affinity score vector over all experts as: $\vs_{\gR} = \sigma\left( \mathrm{TopK}_{-\infty}(\vx^{\top}W_r) \right)$, where $\sigma$ is a scoring function, typically implemented as a softmax or sigmoid. The operator $\mathrm{TopK}_{-\infty}$ retains the top-$K$ values and sets the remaining entries to negative infinity ($-\infty$), enforcing sparsity. The SMoE output is then calculated as a weighted sum of expert outputs, modulated by the affinity scores:
\begin{equation} \label{eqn:moe}
    \hat{y} = \sum_{i = 1}^N \vs_{\gR}^i \cdot g(\vx; W_{e_i}).
\end{equation}

In practice, $K$ is often chosen such that $K < N$ to reduce computational cost while preserving performance.

\input{contents/3_Desining_LibMoE/3_2_vlm}
\input{contents/3_Desining_LibMoE/3_1_pretrain_llm}
\input{contents/3_Desining_LibMoE/3.3_evaluation_benchmarking}
\input{contents/3_Desining_LibMoE/3_4_design_modulars}

%% file: contents/3_Desining_LibMoE/3_2_vlm.tex
\subsection{Vision-Language Model: LLaVA Architecture}

A key challenge in training SMoE models is obtaining a massive dataset and a large amount of compute. Thus, beyond the traditional pre-training setting, we propose to incorporate SMoE training into any existing dense LLM checkpoints via the Sparse Upcycling technique~\citep{komatsuzaki2022sparse}, which duplicates the original model to create experts and continue training them on a downstream dataset as a normal SMoE. Consequently, we can bypass the expensive pre-training step and evaluate SMoE algorithms with the most advanced public LLMs.

\begin{figure*}[ht]
    \centering
    \includegraphics[width=1\linewidth]{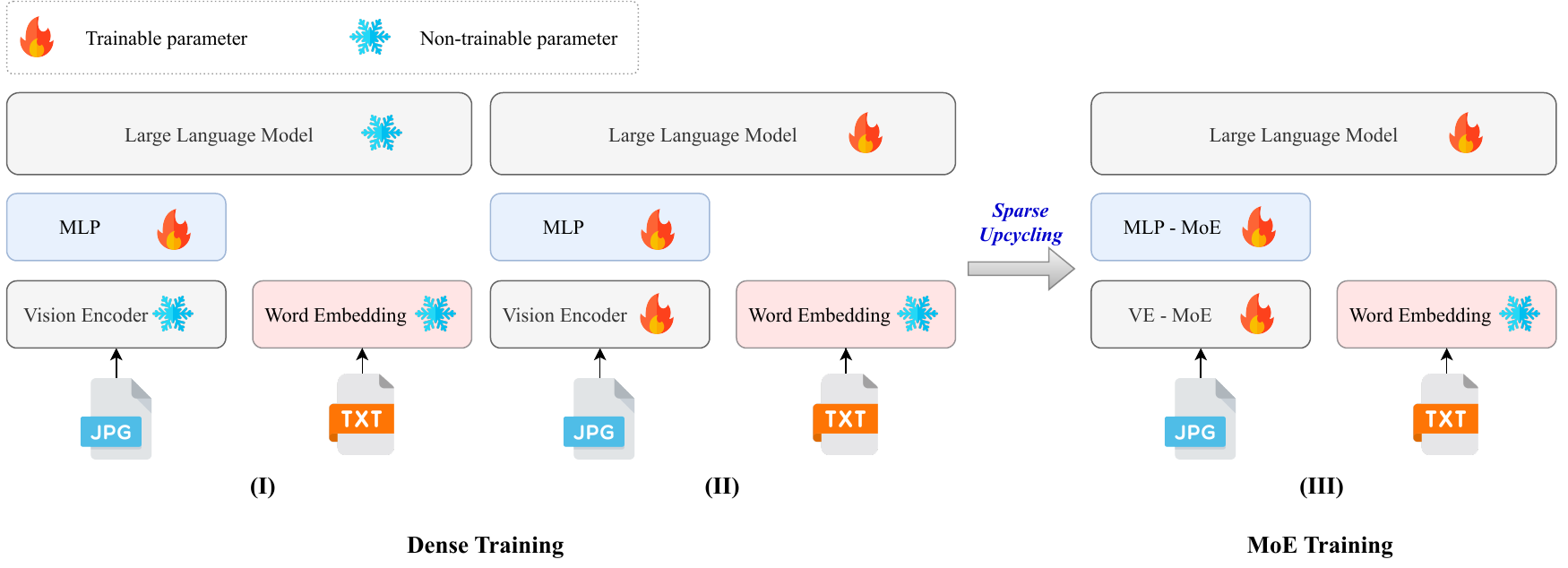}
    \caption{Overview of the LibMoE-VLM architecture and training process. In the first stage of Dense Training, only the MLP is trained to improve alignment. In the second stage, all parameters are trained. During SMoE Training, the feed-forward networks (FFNs) of the Vision Encoder (VE) and MLP Connector are used to initialize the experts within the SMoE framework, and all parameters continue to be trained. }
    \label{fig1_2}
\end{figure*}
\paragraph{Training pipeline}
We adopt a vision--language pretraining task, which represents a challenging multimodal learning setting while requiring a relatively modest amount of data to initiate training (approximately 1.4B tokens). Following the CUMO framework~\citep{li2024cumo}, we upcycle the LLaVA model~\citep{liu2023llava}, which consists of a pretrained visual encoder, a pretrained large language model (LLM), and a randomly initialized vision--language MLP connector. 
Our training procedure follows a two-stage paradigm, comprising \emph{dense training} and \emph{sparse MoE training}. During dense training, we adopt a two-step strategy consisting of \emph{pre-training} and \emph{pre-finetuning}. In the pre-training stage, only the vision--language MLP connector is optimized, while both the visual encoder and the LLM remain frozen. This stage aligns visual representations with the LLM embedding space and establishes a stable multimodal interface without perturbing the pretrained backbones. Subsequently, in the pre-finetuning stage, all model parameters are unfrozen and jointly optimized using high-quality image--caption data. After dense training, we perform visual instruction tuning by converting the model into an SMoE architecture via sparse upcycling, as illustrated in Figure~\ref{fig1_2}. Following prior findings~\citep{li2024cumo}, sparse upcycling is restricted to the MLP connector and visual encoder, since upcycling a dense LLM has been shown to be less effective than directly employing an SMoE LLM. Importantly, the dense training stage is agnostic to SMoE-specific design choices, allowing the resulting checkpoints to be reused across different SMoE algorithms. In our largest-scale experimental setting using 4$\times$H100 GPUs, the complete training pipeline requires slightly over 54 hours in total, with the sparse MoE (visual instruction tuning) stage accounting for approximately 17 hours on the LLaVA-665K dataset or 35 hours on the OneVision (1.2M) dataset. Notably, the dense training stage (approximately 19 hours) is performed only once and reused when comparing different MoE algorithms, which significantly reduces the marginal cost of benchmarking. These results demonstrate that the proposed training setup is computationally accessible. Full specifications of model scales, datasets, and hyperparameters are provided in Appendices~\ref{appendix:exp_setting_vlm} and~\ref{appendix:hyper_settings_vlm}.

%% file: contents/3_Desining_LibMoE/3_1_pretrain_llm.tex
\subsection{Pre-training LLM: Decoder-only Switch Transformer}

A key design of our language modeling experiment is to establish a standardized yet flexible setup that enables fair comparison across different MoE algorithms while maintaining computational efficiency. We adopt a decoder-only Transformer architecture in the style of Switch Transformer \citep{fedus2022switch}, where the feedforward sublayer is replaced by sparsely activated MoE modules. This design supports scaling from small (0.15B) to large (0.68B) models and provides a unified backbone for evaluating MoE variants. To ensure consistency, we keep the architecture and optimization pipeline fixed across experiments, varying only the routing strategies.




\textbf{Training pipeline.} Our experimental setup builds upon the framework of \citet{csordas2023approximating}, with adjustments tailored to our model configurations and minor refinements to the baseline hyperparameters. Our codebase supports all major forms of parallelism, including data, tensor, pipeline, and expert parallelism, ensuring scalability across model sizes. MoE variants are implemented by modularly replacing the MoE component, while keeping other training components identical. This design ensures consistency and reproducibility across experiments, allowing us to isolate the impact of routing. Full specifications of model scales, datasets, and hyperparameters are provided in Appendix~\ref{subappx:lm_setting} and Appendix~\ref{subappx:lm_hyperparams}.

%% file: contents/3_Desining_LibMoE/3.3_evaluation_benchmarking.tex
\subsection{Comprehensive Evaluation of SMoE Algorithms}

Beyond model scaling, rigorous evaluation plays a crucial role in identifying SMoE algorithms that translate to real-world impact. Traditional machine learning studies typically follow the training--validation--testing paradigm, where separate models are trained and evaluated on individual benchmarks. Many existing SMoE works~\citep{chi2022representation,do2023hyperrouter} adopt this framework, training a dedicated model per benchmark.

In contrast, large language models (LLMs) have introduced a paradigm shift, wherein a single model trained on massive corpora can generalize across benchmarks without explicit fine-tuning, a setting commonly referred to as zero-shot evaluation. To align SMoE development with this emerging paradigm and real-world deployment scenarios, LibMoE is designed to support zero-shot evaluation across both language and vision-language domains.

Specifically, for vision-language modeling, we extend the LMMS-Eval framework~\citep{zhang2024lmmsevalrealitycheckevaluation} to evaluate the final checkpoints of various SMoE algorithms. We carefully select 12 widely used benchmarks from LMMS-Eval and report their performance, while also providing a LibMoE model loader to enable future researchers to explore nearly 100 supported benchmarks with minimal effort.

In the language modeling setting, we incorporate nine widely recognized zero-shot benchmarks, including HellaSwag~\citep{zellers2019hellaswag} and ARC-Challenge~\citep{clark2018think}, which serve as canonical examples of standard evaluation datasets. These benchmarks are frequently used in the assessment of state-of-the-art models such as GPT-4~\citep{openai2023gpt4} and DeepSeek-V3~\citep{liu2024deepseekv3}, making them highly representative of real-world evaluation practices. By adopting these standardized tasks, our evaluation aligns closely with established protocols in the LLM community.

Overall, LibMoE is designed to facilitate comprehensive, zero-shot evaluation of SMoE algorithms under realistic conditions, bridging the gap between research experimentation and deployment-oriented assessment.

%% file: contents/3_Desining_LibMoE/3_4_design_modulars.tex
\subsection{Design Principles}

LibMoE is designed with a strong emphasis on modularity, extensibility, and practical usability to accelerate research on SMoE algorithms. At its core, LibMoE consists of three major modules. The MoE module provides a unified abstraction for routing strategies, supporting customizable designs such as gating mechanisms, load balancing losses, and expert selection logic. The training module handles optimization workflows across both LLM and VLM tasks, allowing researchers to integrate pretrained backbones, apply sparse upcycling or pretraining from scratch, and utilize distributed strategies such as data parallelism and model sharding. Finally, the \textit{evaluation module} enables seamless plug-and-play evaluation, supporting a wide range of benchmarks and task formats through standardized APIs.

In addition to providing baseline implementations of leading SMoE algorithms, LibMoE emphasizes ease of experimentation: researchers can rapidly prototype new routing designs or modify existing pipelines with minimal code changes. Its modular infrastructure allows each component routing, training, and evaluation to be extended or swapped independently. Moreover, by abstracting common functionality across tasks and model types, LibMoE lowers the engineering burden typically associated with SMoE experimentation, especially in resource-constrained settings.

%% file: contents/4_Training_LibMoE/training_libmoe.tex
\section{Training LibMoE}

\subsection{SMoE Algorithms}

To demonstrate that LibMoE is well-suited for SMoE research, we implement and benchmark seven state-of-the-art SMoE algorithms. First, we consider methods that modify or improve the router network. These include the original SMoE~\citep{fedus2022switch}, which uses a softmax router; the sigmoid-gated SMoE~\citep{csordas2023approximating}, which has been shown to achieve faster convergence; and XMoE~\citep{chi2022representation}, which mitigates representation collapse and enhances routing capacity through input dimension reduction and cosine-normalized routing. Second, we incorporate recent designs that utilize shared expert mechanisms, such as DeepSeek-V2~\citep{liu2024deepseekv2} with softmax routing and DeepSeek-V3~\citep{liu2024deepseekv3} with a sigmoid router. Inspired by these designs, we implement SharedE-V2 and SharedE-V3, which adopt the shared-expert principles of DeepSeek-V2 and DeepSeek-V3, respectively. Lastly, we include MoE++\citep{jin2024moe++}, which augments standard SMoE with zero-computation experts and pathway-aware routing to improve efficiency and throughput, and TC-MoE\citep{yan2025tc}, which introduces a ternary expert choice space and reward-based routing to reduce redundancy while enhancing model expressiveness. These diverse algorithms allow us to evaluate LibMoE’s compatibility with a wide range of routing strategies and architectural innovations. A concise overview of these benchmarked SMoE variants is provided in Appendix~\ref{appendix:moe_overview}.

\input{contents/4_Training_LibMoE/result_sparse_upcycling_tab}

\input{contents/4_Training_LibMoE/result_pretrain_tab}

\subsection{Performance Comparison}

Across both vision–language modeling and language modeling settings, we observe a consistent trend: performance differences among contemporary SMoE variants remain relatively modest under matched training conditions. In the VLM sparse upcycling regime, no single method consistently outperforms others, with only marginal variations observed across benchmarks. Similarly, in language model pretraining, while recent MoE designs exhibit incremental improvements, these gains do not amount to a decisive advantage over the standard SMoE formulation. Collectively, these results reflect current practical deployments of frontier models, where the original SMoE design is often preferred for its simplicity, stability, and scalability, as opposed to more complex routing mechanisms that yield limited empirical benefits under comparable training conditions.

%% file: contents/4_Training_LibMoE/result_sparse_upcycling_tab.tex
\begin{table}[ht]
\centering
\caption{Performance of SMoE variants with \textbf{6} versus \textbf{3} active experts on a ViT backbone (5.67B parameters). Bold values mark the best score in each column; $\downarrow$/$\uparrow$ indicate that lower/higher values are preferable, respectively.}
\scriptsize
\renewcommand{\arraystretch}{1.02}
\begin{adjustbox}{width=\textwidth,center}
\begin{tabular}{lcccccccccccc|cc}
\toprule
\textbf{\begin{tabular}[c]{@{}c@{}}MoE \\ Method\end{tabular}} & 
\textbf{AI2D} & 
\textbf{\begin{tabular}[c]{@{}c@{}}Text \\ VQA\end{tabular}} & 
\textbf{GQA} & 
\textbf{\begin{tabular}[c]{@{}c@{}}MM \\ Bench\end{tabular}} & 
\textbf{\begin{tabular}[c]{@{}c@{}}Hallusion \\ Bench\end{tabular}} & 
\textbf{\begin{tabular}[c]{@{}c@{}}Math \\ Vista\end{tabular}} & 
\textbf{MMMU} & 
\textbf{MMStar} & 
\textbf{Pope} & 
\textbf{MME} & 
\textbf{\begin{tabular}[c]{@{}c@{}}MME \\ RW\end{tabular}} & 
\textbf{\begin{tabular}[c]{@{}c@{}}OCR \\ Bench\end{tabular}} & 
\textbf{\begin{tabular}[c]{@{}c@{}}AVG \\ Acc\end{tabular}}$\uparrow$ & 
\textbf{\begin{tabular}[c]{@{}c@{}}AVG \\ Rank\end{tabular}}$\downarrow$ \\
\midrule
\multicolumn{15}{c}{\textit{LLAVA + OneVision / 1M2 samples}} \\
\midrule

SMoE         & 69.56          & 43.93          & 61.51 & 71.31          & 46.90          & 37.90 & 41.56          & 41.23          & 86.28          & 63.33          & 27.83          & 37.50          & 52.40          & 5.96          \\
XMoE         & 69.72          & 43.93          & 61.52 & 72.25          & \textbf{47.42} & 38.50 & 42.11          & \textbf{43.99} & 86.61          & 63.81          & 29.18          & \textbf{39.40} & 53.20          & 3.33          \\
$\sigma$-MoE & 69.79          & 44.69          & 61.70 & 71.74          & 47.00          & 38.40 & \textbf{43.11} & 42.08          & 86.69          & 63.80          & 29.70          & 38.40          & 53.09          & 3.33          \\
SharedE-V2   & 70.56          & \textbf{45.04} & 61.34 & 71.13          & 47.11          & \textbf{39.50} & 42.78          & 42.73          & 86.53          & 63.93          & 29.55          & 38.70          & 53.24          & 3.33          \\
SharedE-V3   & \textbf{71.92} & 44.59          & \textbf{61.93} & \textbf{72.59} & 46.37          & 38.20 & 42.33          & 43.30          & \textbf{86.78} & \textbf{64.61} & 29.29          & \textbf{39.40} & \textbf{53.44} & \textbf{2.46} \\
TC-MOE       & 70.08          & 43.75          & 61.89 & 71.05          & 45.74          & 38.10 & 41.89          & 43.64          & 86.76          & 63.05          & \textbf{31.84} & 38.30          & 53.01          & 4.50          \\
MOE++        & 70.13          & 43.37          & 61.52 & 71.39          & 46.16          & 38.60 & 40.78          & 43.24          & 86.60          & 63.26          & 28.19          & 37.50          & 52.56          & 5.08       \\
\midrule
\multicolumn{15}{c}{\textit{LLAVA / 665K samples}} \\
\midrule
SMoE         & 65.52          & 41.51          & 61.62          & 72.25          & 41.75          & 29.50          & 41.67          & \textbf{42.24} & 87.13          & 61.05          & 32.15          & 31.70 & 50.67          & 3.42          \\
XMoE         & \textbf{65.84} & 41.96          & 61.61          & 72.16          & 41.85          & \textbf{31.60} & 41.67          & 40.32          & 86.64          & 60.95          & 32.73          & \textbf{33.20} & \textbf{50.88} & \textbf{3.08} \\
$\sigma$-MoE & 65.52          & \textbf{42.08} & \textbf{61.74} & 71.22          & 40.80          & 30.30          & 41.22          & 41.99          & 86.64          & \textbf{61.44} & 32.31          & 32.50 & 50.65          & 3.71          \\
SharedE-V2   & 64.77          & 41.96          & 61.27          & 71.74          & 41.85          & 30.90          & \textbf{43.33} & 41.56          & \textbf{86.82} & 60.52          & \textbf{32.88} & 31.40 & 50.75          & 3.67          \\
SharedE-V3   & 65.58          & 42.06          & 61.26          & \textbf{72.42} & 41.43          & 30.60          & 42.44          & 41.75          & 86.81          & 60.93          & 31.47          & 32.60 & 50.86          & 3.33          \\
TC-MOE       & 65.50          & 40.70          & 61.19          & 71.22          & 42.06          & 29.30          & 41.22          & 41.10          & 86.53          & 60.30          & 31.89          & 33.00 & 50.34          & 5.42          \\
MOE++        & 65.03          & 41.44          & 60.61          & 71.74          & \textbf{42.69} & 30.30          & 43.00          & 40.32          & 86.58          & 60.11          & 31.32          & 31.20 & 50.36          & 5.38         \\

\bottomrule
\end{tabular}
\end{adjustbox}
\label{tab:vlm_moe_results}

\end{table}

%% file: contents/4_Training_LibMoE/result_pretrain_tab.tex
\begin{table}[ht]
\centering
\small
\caption{Performance of SMoE variants with \textbf{66} total experts and \textbf{8} active experts in the language pre-training setting, evaluated on small-scale (0.15B) and large-scale (0.68B) models. PPL denotes perplexity for language modeling. Bold values denote the best result in each column; $\downarrow$/$\uparrow$ indicate that lower/higher values are preferable, respectively.}
\resizebox{\textwidth}{!}{
\begin{tabular}{c c c ccccccccc |cc}
\toprule
 & \makecell{\textbf{MoE Method}} & \makecell{\textbf{PPL} $\downarrow$}
 & \makecell{\textbf{LAM}\\ \textbf{BADA}}
 & \makecell{\textbf{BLiMP}}
 & \makecell{\textbf{CBT}}
 & \makecell{\textbf{Hella}\\ \textbf{Swag}}
 & \makecell{\textbf{PIQA}}
 & \makecell{\textbf{ARC-}\\ \textbf{Easy}}
 & \makecell{\textbf{RACE}}
 & \makecell{\textbf{SIQA}}
 & \makecell{\textbf{Common}\\ \textbf{SenseQA}}
 & \makecell{\textbf{AVG} \\ \textbf{Acc} $\uparrow$}
 & \makecell{\textbf{AVG} \\ \textbf{Rank} $\downarrow$} \\
\midrule

 & SMoE        & 13.63 & 25.27 & \textbf{77.71} & 84.18 & \textbf{29.43} & 57.94 & 32.68 & 30.11 & 35.62 & 24.65 & 44.18 & 4.50 \\
 & XMoE        & 13.98 & 24.57 & 76.53 & 84.12 & 29.34 & 58.27 & 32.26 & 29.69 & 35.47 & 24.49 & 43.86 & 6.45 \\
 & $\sigma$-MoE& 13.61 & 25.43 & 77.38 & 84.23 & 29.13 & 58.92 & 32.73 & \textbf{31.05} & 34.90 & 24.90 & 44.30 & 4.30 \\
 & SharedE-V2  & 13.49 & 25.29 & 77.37 & 84.33 & 29.38 & \textbf{60.17} & \textbf{33.83} & 31.02 & 35.57 & 24.98 & \textbf{44.66} & \textbf{2.80} \\
 & SharedE-V3  & \textbf{13.42} & 25.49 & 77.20 & 84.40 & 29.38 & 59.14 & 32.52 & 30.60 & 35.57 & 25.47 & 44.42 & 3.20 \\
 & TC-MoE      & 13.51 & \textbf{25.60} & 76.91 & 84.68 & 29.27 & 59.03 & 33.02 & 30.63 & \textbf{36.03} & \textbf{26.37} & 44.62 & 2.90 \\
\multirow{-7}{*}{\makecell{Small Model\\(0.15B)}} 
 & MoE++       & 13.54 & 25.45 & 77.23 & \textbf{84.83} & 29.28 & 58.49 & 33.49 & 30.11 & 35.62 & 24.49 & 44.33 & 3.85 \\

\midrule

 & SMoE        & 9.51 & 37.13 & 80.47 & 89.83 & 37.49 & 64.36 & 38.22 & 33.03 & 37.41 & 26.54 & 49.39 & 5.15 \\
 & XMoE        & 9.66 & 35.25 & 80.38 & 89.35 & 37.19 & 64.20 & 38.99 & 32.95 & 37.77 & 28.34 & 49.38 & 5.75 \\
 & $\sigma$-MoE& 9.46 & 37.56 & 81.08 & 89.57 & 37.52 & 64.91 & 39.15 & 32.68 & 37.67 & \textbf{28.50} & 49.85 & 3.60 \\
 & SharedE-V2  & 9.52 & 37.11 & 80.98 & 89.93 & 37.14 & 64.36 & 38.06 & 33.17 & 36.95 & 27.35 & 49.45 & 5.25 \\
 & SharedE-V3  & 9.49 & 36.88 & \textbf{81.28} & 89.65 & 37.32 & \textbf{65.72} & 38.86 & 33.12 & \textbf{38.59} & 28.09 & 49.95 & 3.60 \\
 & TC-MoE      & \textbf{9.38} & 37.87 & 81.21 & \textbf{90.19} & \textbf{37.95} & 64.47 & 39.28 & 33.77 & 37.92 & 27.85 & 50.06 & 2.35 \\
\multirow{-7}{*}{\makecell{Large Model\\(0.68B)}} 
 & MoE++       & \textbf{9.38} & \textbf{38.80} & 80.88 & 89.77 & 37.70 & 64.64 & \textbf{39.37} & \textbf{34.02} & 37.97 & 28.34 & \textbf{50.16} & \textbf{2.30} \\
\bottomrule
\end{tabular}}
\label{tab:lm_result}
\end{table}

%% file: contents/5_Analysis/analysis.tex
\section{Analysis}
\label{sec:analysis}

Beyond empirical performance, we analyze how different MoE architectures route, specialize, and balance experts during training. We focus on expert-selection dynamics, capacity utilization, and the effects of initialization, architecture, and task domain. Our analysis spans both from-scratch pretraining (a 0.15B LLM trained on 6.55B tokens) and sparse upcycling (a VLM fine-tuned on LLaVA-665K), and additionally includes Qwen3-VL-30B-A3B~\cite{Yang2025Qwen3TR}, a representative large-scale SMoE model. Together, these settings provide a unified view of routing behavior across regimes.
\paragraph{a) How Stable Is Expert Routing Throughout Training in MoE Algorithms?}  \label{subsec:router_change_rate}
\begin{figure*}[ht]
    \centering
    \includegraphics[width=0.9\linewidth]{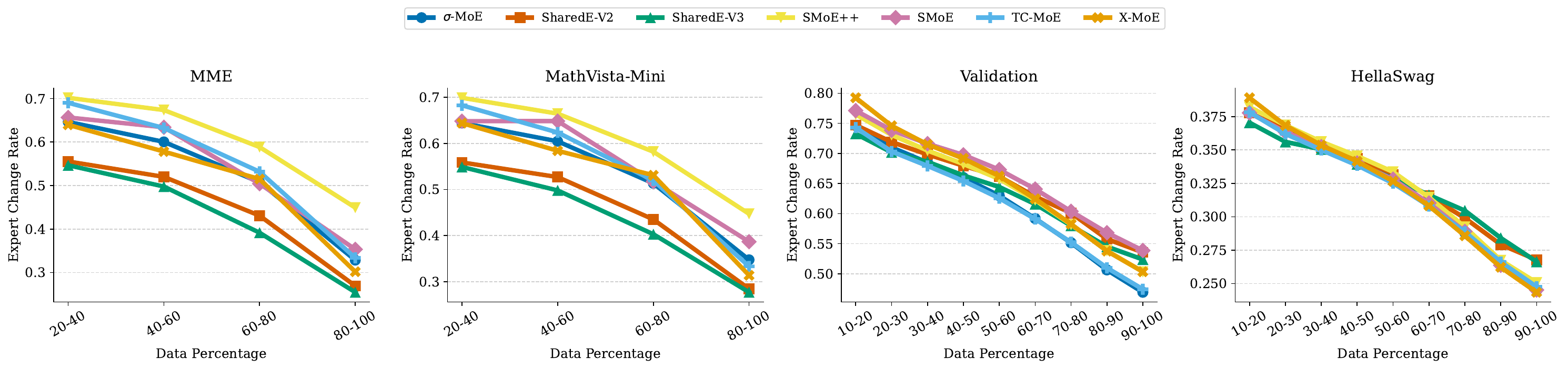}
    \caption{Impact of Training Data Percentage on Expert Selection.}
    \label{fig:expert_change_rate}
\end{figure*}

This experiment investigates how training-data scale influences expert-selection dynamics across SMoE algorithms. We quantify routing stability using the \emph{Expert Change Rate} (ECR). For a fixed evaluation set $\gD$, we compare consecutive checkpoints and, for each token at each MoE layer, record the selected expert. We define $\mathrm{ECR}(\gD)$ as the fraction of token-to-expert assignments on $\gD$ (aggregated across MoE layers) for which the selected expert differs between the two checkpoints. Lower ECR corresponds to more stable routing, where the router makes fewer reassignment decisions for identical inputs, whereas higher ECR indicates more volatile routing, with experts being reassigned more frequently over the course of training.

Figure~\ref{fig:expert_change_rate} reports ECR throughout training for both vision-language models (MME, MathVista; sparse upcycling) and language models (Validation, HellaSwag; from-scratch pretraining). Across all settings, we observe that SMoE algorithms exhibit a consistently decreasing ECR as training progresses. Among the evaluated methods, XMoE shows stronger late-stage convergence on VLM tasks. We attribute this to its cosine-normalized routing and the smaller learning rate, which makes this setup more efficient. In contrast, during language-model pretraining, XMoE exhibits a relatively high ECR (peaking around the $40\%$ training mark), consistent with the degraded performance reported in Table~\ref{tab:lm_result}. We hypothesize that this degradation arises from an interaction between cosine-normalized routing and the pretraining learning rate. By bounding the router logits (up to a temperature/scale), cosine/L2 normalization can reduce the effective logit margin and make softmax gating more sensitive to the chosen scale and step size \citep{agarwala_temperature}. Moreover, since normalization choices can affect gradient scaling, an overly large learning rate can exacerbate optimization instability and even trigger divergence accompanied by growing internal activations/norms \citep{xiong2020_layernorm_transformer,rybakakov2024stability}.

Furthermore, we observe a divergent trend in the behavior of shared expert architectures namely, SharedE-V2 and SharedE-V3 across different tasks. In vision language models (VLMs), both variants exhibit significantly higher routing stability compared to other methods, likely because $N_s$ shared experts are always active while the router only assigns tokens to $K_r$ routed experts, reducing the degrees of freedom in expert assignment relative to fully routed alternatives. 

\begin{wrapfigure}{r}{8.6cm}        
  \centering
  \includegraphics[width=\linewidth]{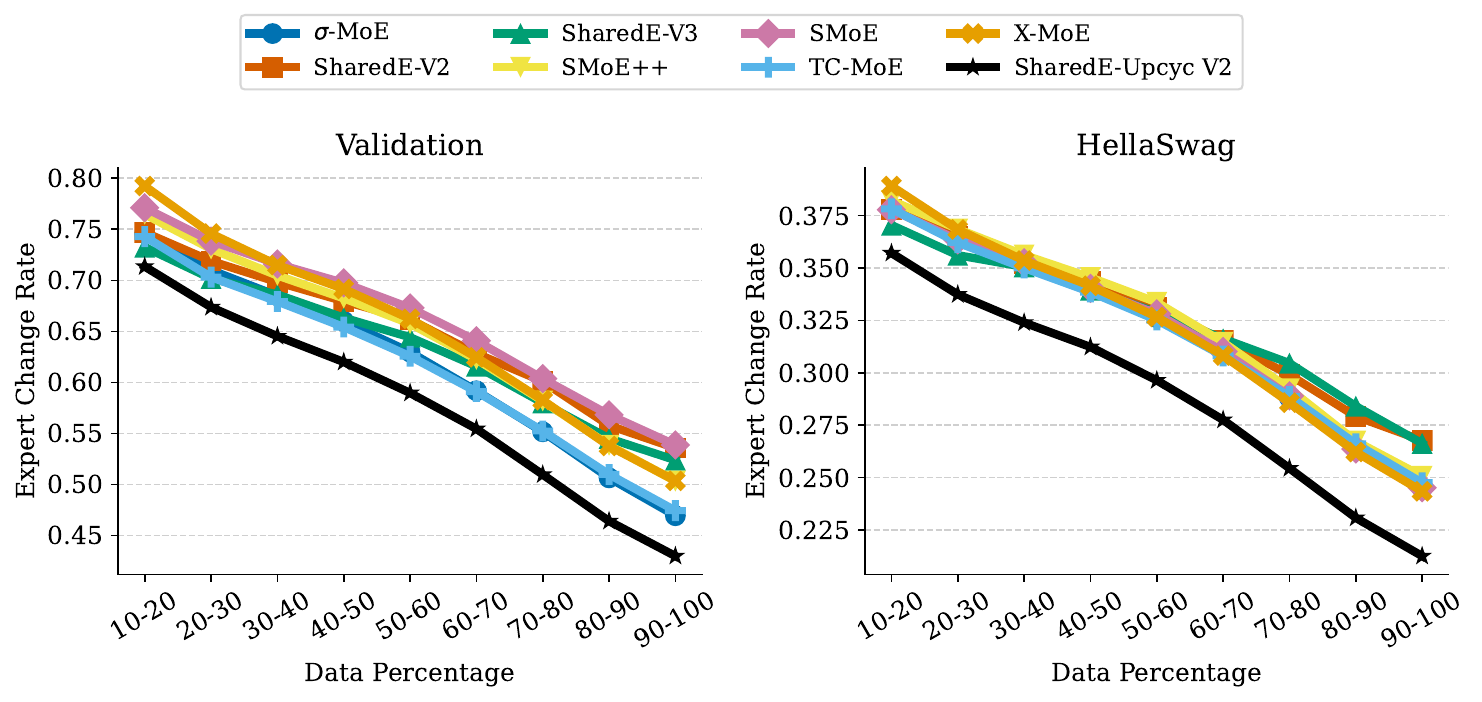}
  \caption{Effect of upcycled shared experts trained on prior tasks on routing behavior, measured by expert change rate during language model pretraining.}
  \label{fig:expert_change_rate_sharedexperts}
\end{wrapfigure}

However, this advantage disappears during language model pretraining. We hypothesize that this divergence arises from the distinct initialization schemes and functional roles of shared experts. In the VLM sparse upcycling setup, SMoE layers are initialized from a pretrained dense model, where shared experts already encode general purpose visual linguistic representations. These pretrained experts act as stable anchors, enabling the router to focus its capacity on distributing tokens among non-shared experts, thereby reducing volatility in expert selection. In contrast, language model pretraining starts from random initialization. Here, both the router and shared experts must co-adapt from scratch. As the shared experts' representations evolve over time, their influence on the final output changes accordingly. This compels the router to continuously revise its decisions not just for load balancing but also to adapt to the shifting behavior of shared experts in order to minimize the task loss. The result is a tight coupling between unstable shared experts and routing dynamics, which leads to higher expert change rates and delayed convergence.

To validate this hypothesis, we conduct a controlled experiment (Figure~\ref{fig:expert_change_rate_sharedexperts}) by weakening the shared expert’s influence: we upcycle only the shared expert from a pretrained checkpoint while randomly initializing the rest of the model, referred to as SharedE-Upcyc V2. The results show that when the shared expert is strong and stable, the router can focus on allocating inputs to task-specific experts leading to a lower expert change rate compared to other baselines. Interestingly, this stabilizing effect diminishes as training data increases. In large-scale regimes (e.g., Table~\ref{tab:vlm_moe_results}, trained on over 1M samples), SharedE-V2 and SharedE-V3 not only preserve routing stability but also achieve top performance. This confirms that shared expert architectures are particularly beneficial in data-rich scenarios, where their pre-initialized knowledge complements the router's optimization.

Taken together, these findings highlight that routing stability is shaped by the training regime. In from-scratch language-model pretraining, both the router and experts must co-adapt under random initialization, so early-stage stability is critical to prevent prolonged assignment churn. In contrast, VLM sparse upcycling starts from a pretrained dense checkpoint, where experts can serve as stable general-purpose anchors; consequently, routing tends to evolve more smoothly and robustness becomes more important than rapid early convergence. Therefore, effective sparse MoE designs should account for both regimes: fast stabilization during from-scratch pretraining and sustained robustness during upcycling.

\paragraph{b) Are the Selected Experts Truly Optimal in MoE Algorithms?}

\begin{figure*}[ht]
    \centering
    \includegraphics[width=0.9\linewidth]{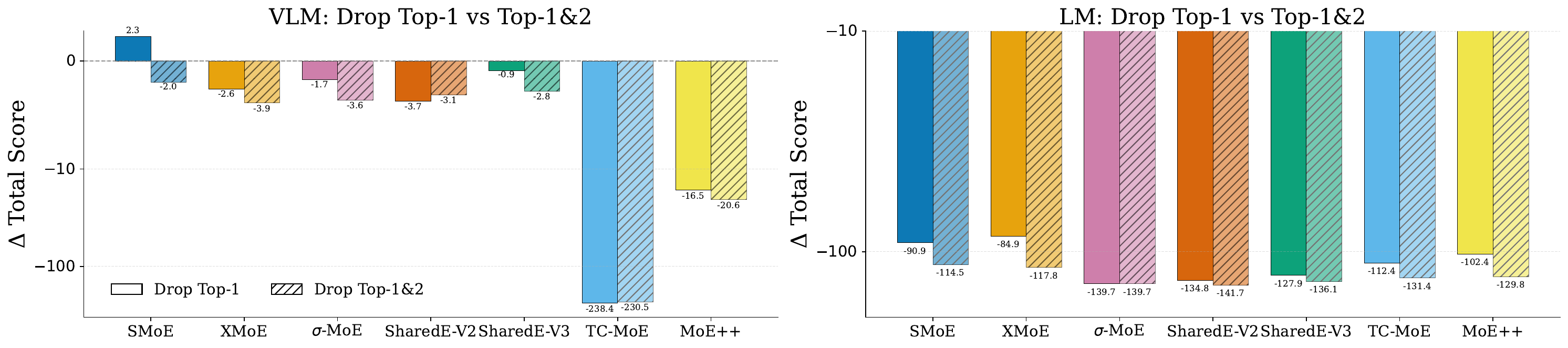}
    \caption{Performance of SMoE variants under routing perturbations, where the top-1 expert is replaced by the top-$(K{+}1)$ expert in vision–language and language modeling tasks.}
    \label{fig:droptop1}
\end{figure*}

To assess whether SMoE algorithms fully exploit their expert capacity, we conduct a diagnostic routing perturbation experiment across seven architectures. Specifically, we intentionally degrade the routing decision by replacing the top-1 expert (i.e., the expert with the highest affinity score) with the top-$(K{+}1)$ expert, and quantify the resulting performance change using $\Delta \text{Total Score}$ (see Appendix~\ref{subappx:drop1and2} for details). Here, a higher $\Delta \text{Total Score}$ indicates improved performance relative to the original routing configuration. This controlled perturbation allows us to systematically measure each model’s sensitivity to suboptimal expert assignments and to identify potential inefficiencies in expert utilization.
Figure~\ref{fig:droptop1} reports the cumulative accuracy drop across benchmarks under two settings: replacing only the top-1 expert (\textit{DropTop1}) and replacing both the top-1 and top-2 experts (\textit{DropTop1$\&$2}). As expected, most SMoE variants exhibit performance degradation, confirming that precise expert selection is critical to maintaining model quality. Overall, in the language modeling (pretraining) setting, all methods suffer large and relatively uniform performance drops, reflecting a consistent reliance on precise expert routing across architectures. In contrast, vision-language models exhibit greater divergence. Notably, SMoE shows a counterintuitive \textbf{a +2.3 gain in $\Delta \text{Total Score}$} under the \textit{DropTop1} perturbation suggesting that its original routing configuration may have failed to effectively utilize its expert capacity. This exposes a potential inefficiency in the router, where suboptimal expert choices actually yield better outcomes. Conversely, TC-MoE suffers the most severe degradation, highlighting its strong dependence on highly optimized, task-aligned routing decisions.

In summary, this experiment highlights that the impact of routing quality is fundamentally intertwined with architectural design across SMoE variants. Architectures such as TC-MoE and MoE++ exhibit pronounced performance degradation under routing perturbations, indicating that their gains rely heavily on precise and task-aligned expert selection. This sensitivity reveals a strong coupling between routing robustness and the effectiveness with which the architecture exploits its representational capacity. However, the results also demonstrate that routing robustness alone is insufficient to guarantee the highest absolute performance. Each SMoE variant is constrained by an inherent architectural capacity ceiling, beyond which improvements in routing precision yield diminishing returns. Consequently, optimal performance depends not only on selecting the “right” experts, but on whether the routing mechanism is capable of fully leveraging the architectural capacity available. These findings suggest that routing behavior should be interpreted in conjunction with architectural design, rather than as an isolated indicator of model quality.

\paragraph{c) How Does Normalized Expert-Allocation Entropy Reveal Domain Specialization Across SMoE Variants?}

\begin{figure}[htbp]
    \centering
    
    \begin{subfigure}[t]{\textwidth}
        \centering
        \includegraphics[width=\linewidth]{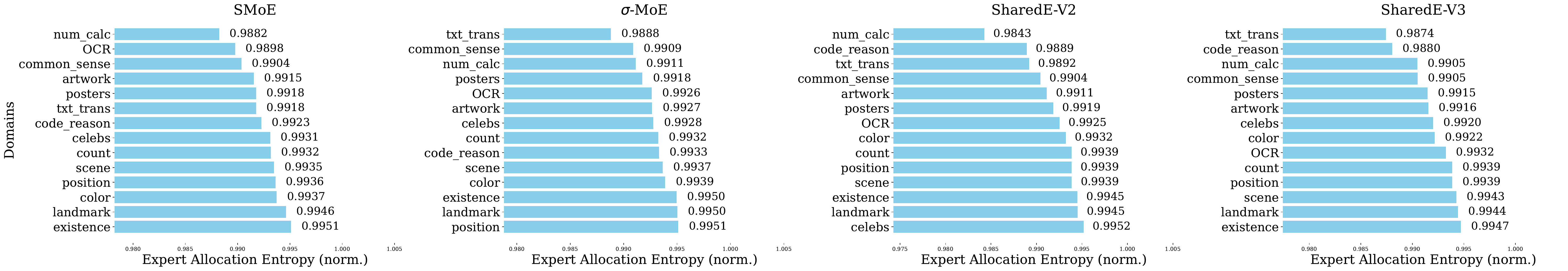}
    \end{subfigure}

    \begin{subfigure}[t]{\textwidth}
        \centering
        \includegraphics[width=\linewidth]{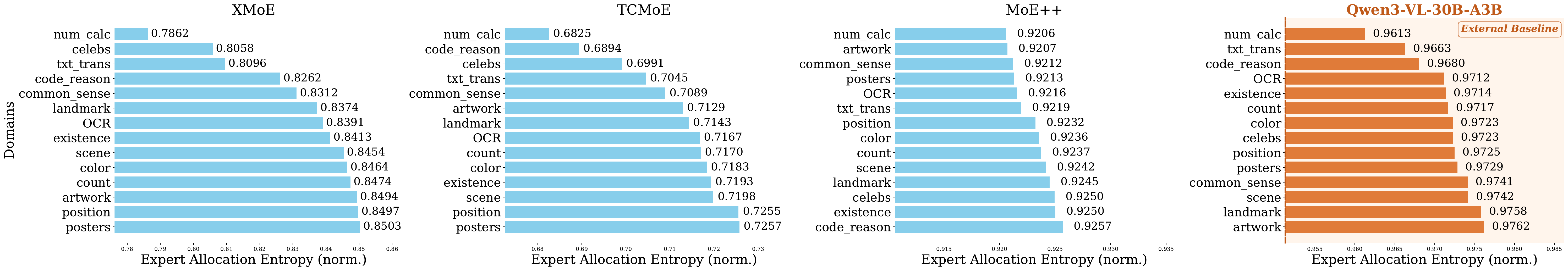}
    \end{subfigure}
    
    \caption{Normalized expert-allocation entropy across domains for seven LibMoE variants, with Qwen3-VL-30B-A3B included as a large-scale baseline.}
    \label{fig:entropy_distr_experts}
\end{figure}

In Figure~\ref{fig:entropy_distr_experts}, we analyze domain specialization through the normalized Expert Allocation Entropy (EAE) metric (see Appendix~\ref{subappx:eae}), which measures how evenly routing decisions are distributed across experts for each MME subtask. EAE provides a direct view of the specialization--coverage trade-off: lower EAE indicates that the router repeatedly relies on a smaller subset of experts, whereas higher EAE indicates more uniform expert usage. This makes the metric particularly useful for distinguishing architectures that develop strong domain-specific specialization from those that maintain broad, balanced coverage across tasks.

Across subtasks, a consistent pattern emerges: reasoning-heavy tasks such as \texttt{text\_translation}, \texttt{code\_reasoning}, and \texttt{numerical\_calculation} tend to yield lower EAE than simpler tasks. In other words, as the task becomes more compositional or algorithmic, the router shifts from broad expert sharing toward sharper expert separation. The same directional trend is also visible in Qwen3-VL-30B-A3B: even at larger scale, reasoning-oriented subtasks still reduce EAE, suggesting that domain-specialized routing is not merely a small-model artifact.

The LibMoE variants separate into three clear regimes. TC-MoE shows the lowest EAE and the strongest concentration, with normalized EAE dropping to about $0.64$ in the most concentrated cases. This indicates aggressive specialization, but it also suggests \textit{over-specialization}: much of the expert pool is used rarely, which can reduce coverage and make performance more sensitive to routing errors. This interpretation is consistent with the strong degradation of TC-MoE under routing perturbation in Figure~\ref{fig:droptop1}. XMoE occupies a more favorable middle ground. Relative to the more uniform baselines, it maintains lower entropy, but it also exhibits the largest task-wise spread in EAE, ranging from about $0.78$ to $0.85$. This wider range suggests task-adaptive specialization: XMoE concentrates routing more aggressively for difficult reasoning tasks, yet distributes computation more broadly on simpler tasks, avoiding permanent collapse onto a small expert subset.

By contrast, SMoE, $\sigma$-MoE, MoE++, SharedE-V2, and SharedE-V3 show highly stable EAE curves, with variation below one percentage point across subtasks. These routers therefore prioritize consistent expert coverage over strong task-specific separation. Such behavior may sacrifice some fine-grained specialization, but it can also improve robustness by preventing a small number of experts from dominating the routing decisions. More importantly, it suggests that any task sensitivity in these architectures may come less from \emph{which} experts are selected and more from \emph{how} the selected experts are weighted---a question we analyze next.

Overall, EAE shows that the most effective routing behavior is not simply the lowest-entropy one. Extremely low EAE can reflect useful specialization, but it can also signal brittle expert collapse; extremely flat EAE improves balance, but may under-express domain structure. The most compelling behavior is task-adaptive entropy: specialize when the input demands it, while preserving enough coverage to remain robust. This clarifies the main message of Figure~\ref{fig:entropy_distr_experts}: TC-MoE tends to over-specialize, the balanced family tends to under-specialize, and XMoE comes closest to navigating the middle ground, while Qwen3-VL-30B-A3B shows that the same trade-off remains relevant even at larger scale.

\begin{figure*}[ht]
\centering
\includegraphics[width=0.9\linewidth]{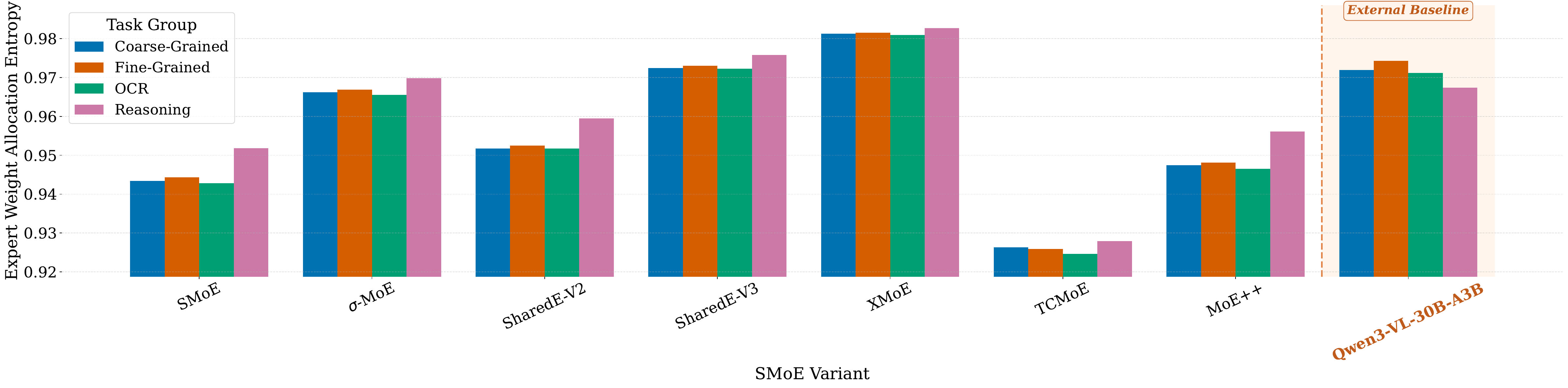}
\caption{Normalized entropy of expert weight allocation across different routing strategies. Higher values indicate more uniform sharing of router weight among the selected experts.}
\label{fig:weghts_allocation}
\end{figure*}

\paragraph{d) How Do Sparse MoE Methods Allocate Weights Across Experts and Tasks?}
Beyond analyzing how expert selection varies across subtasks, we also investigate how the router distributes the weight mass of $\vs_{\gR}$ among the selected experts, a critical factor that influences the model's output behavior.

In Figure~\ref{fig:weghts_allocation}, we report the normalized entropy of expert-weight allocation (EWA), computed as the Shannon entropy of the softmax router weights over the selected experts, normalized by the maximum entropy of a $K$-dimensional vector, and averaged over samples within each task group (Coarse-Grained, Fine-Grained, OCR, and Reasoning). Higher EWA indicates that the router distributes weight more evenly across the selected experts, whereas lower EWA indicates that it places most of the mass on only a few of them. A first notable result is that most methods achieve relatively high EWA, showing that their routers generally distribute weight broadly rather than allowing a single selected expert to dominate. In other words, once the top-$K$ experts are chosen, these methods tend to let multiple experts contribute meaningfully to the final output. This interpretation is further reinforced by the external baseline Qwen3-VL-30B-A3B, which also maintains consistently high EWA across all four task groups. At the same time, for nearly every architecture, the EWA values remain relatively stable across Coarse-Grained, Fine-Grained, OCR, and Reasoning tasks, suggesting that weight-allocation style is largely an intrinsic property of the routing mechanism rather than a strongly task-dependent effect.

The methods themselves separate into clear regimes. At the high-entropy end, XMoE shows the most uniform allocation, with $\sigma$-MoE and SharedE-V3 also exhibiting strongly balanced behavior. In these models, several selected experts tend to contribute with comparable weight, which suggests fuller use of the available expert capacity. At the opposite extreme, TC-MoE records the lowest EWA across all task groups, again with very little inter-task variation. This indicates a consistently sharper allocation pattern in which the router relies heavily on a small subset of experts regardless of task type. Such concentrated routing may encourage stronger specialization, but it also reduces expert diversity, leaves part of the model capacity underutilized, and makes the system more vulnerable when the dominant experts are not well matched to the input. SMoE, MoE++, and SharedE-V2 occupy an intermediate regime, balancing concentration and sharing more evenly.

Overall, the EWA analysis shows that expert utilization depends not only on \emph{which} experts are selected, but also on \emph{how} the router allocates weight among them. High-EWA methods behave more like collaborative routers, allowing multiple experts to participate meaningfully in each decision, whereas low-EWA methods behave more like decisive routers that place most of the computation on a narrow subset of experts. This provides a complementary lens for interpreting performance differences across SMoE variants: the central trade-off is not simply specialization versus balance, but how effectively each architecture converts its selected experts into useful computation.

\begin{figure*}[ht]
    \centering
    \includegraphics[width=\linewidth]{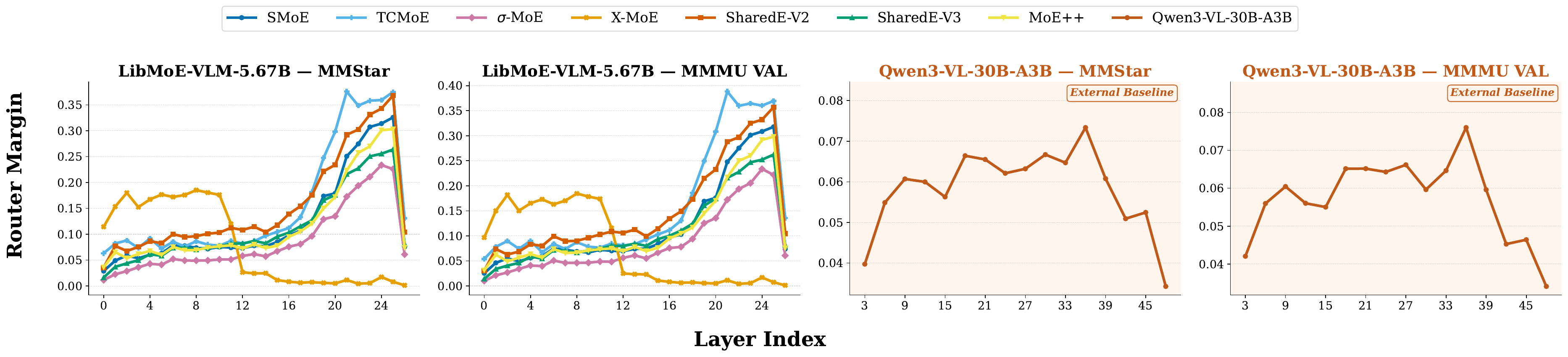}
    \caption{Layer-wise router margin in the vision--language setting, computed as the gap between the top-1 and top-2 routing scores. Left: LibMoE-VLM-5.67B evaluated on MMStar and MMMU Validation. Right: Qwen3-VL-30B evaluated on the same benchmarks.}
    \label{fig:router_margin}
\end{figure*}

\paragraph{e) Does Router Confidence Grow with Depth in MoE Networks?}

Router margin measures how strongly the router prefers the top-1 expert over the runner-up, defined as the difference between the top-1 and top-2 routing scores (see Appendix~\ref{subappx:router_margin}). Figure~\ref{fig:router_margin} reveals a consistent depth-wise pattern rather than a simple monotonic increase. For most LibMoE-VLM methods on both MMStar and MMMU Validation, margins remain modest through the early and middle layers, rise clearly in the later layers, and then drop at the final layer. Importantly, this is the same qualitative trend observed in the large-scale Qwen3-VL-30B reference model: router confidence strengthens with depth before relaxing at the last layers. This alignment is notable because Qwen3-VL-30B has substantially more layers, yet LibMoE still reproduces the same large-scale routing trajectory, suggesting that its routing dynamics capture a meaningful depth-wise behavior rather than an artifact of model size.

Across methods, TC-MoE reaches the highest late-layer margins, indicating the sharpest expert preference near the end of the network. SMoE, SharedE-V2, $\sigma$-MoE, SharedE-V3, and MoE++ all broadly follow the same late-rise-then-final-drop pattern, differing mainly in magnitude. XMoE is the main exception: it begins with relatively large margins in shallow layers, then collapses to near-zero after roughly the first half of the network, indicating increasingly weak separation between its top two experts in deeper layers. Qwen3-VL-30B provides the large-scale reference here: although its trajectory is smoother and spread over many more layers, it still exhibits the same overall increase toward later layers followed by a downturn at the end. Overall, the key observation is not merely that some LibMoE variants become more decisive with depth, but that several of them mirror the qualitative depth profile of Qwen3-VL-30B despite operating at a much smaller scale.

\paragraph{f) Do Experts Exhibit Similarity in Sparse Upcycling?}
\label{subsec:expert_similarity_sparse_upcycling}

\begin{figure*}[ht]
    \centering
    
    \begin{subfigure}[t]{0.69\linewidth}
        \centering
        \includegraphics[width=\linewidth]{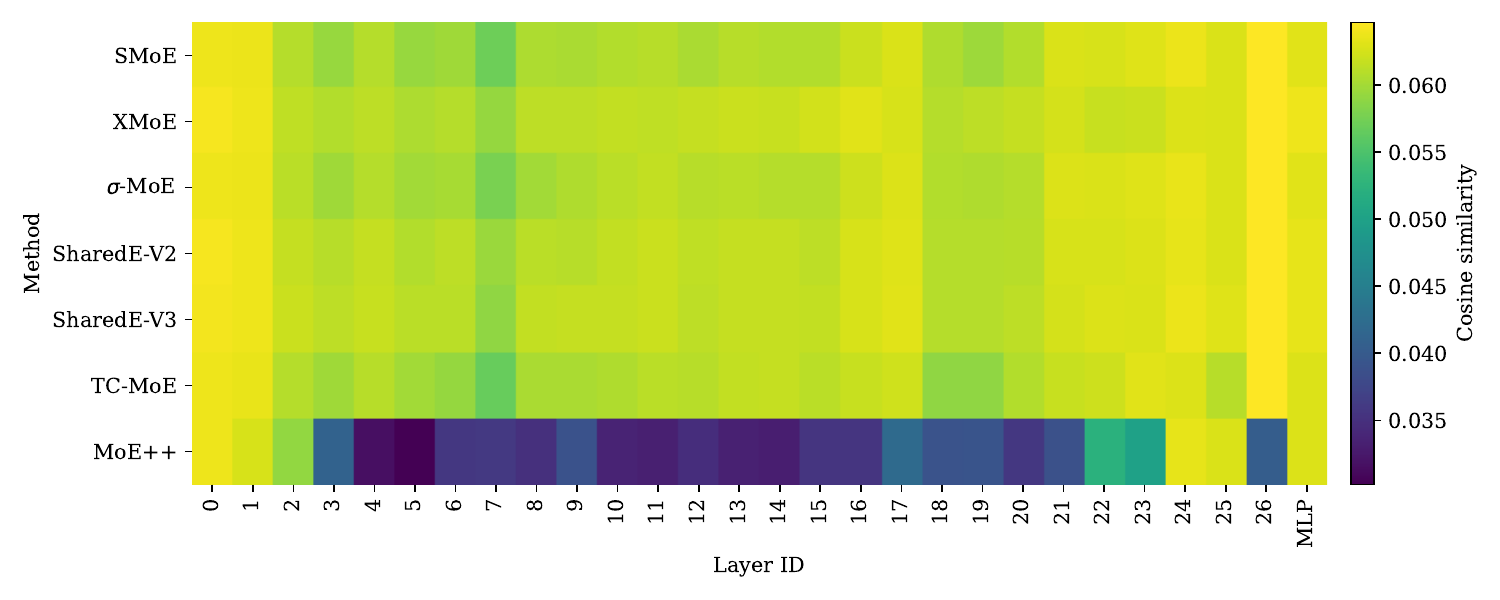}
        \caption{Expert similarity across LibMoE-VLM methods under sparse upcycling.}
        \label{fig:layer_wise_diversity}
    \end{subfigure}
    \hfill
    \begin{subfigure}[t]{0.3\linewidth}
        \centering
        \includegraphics[width=\linewidth]{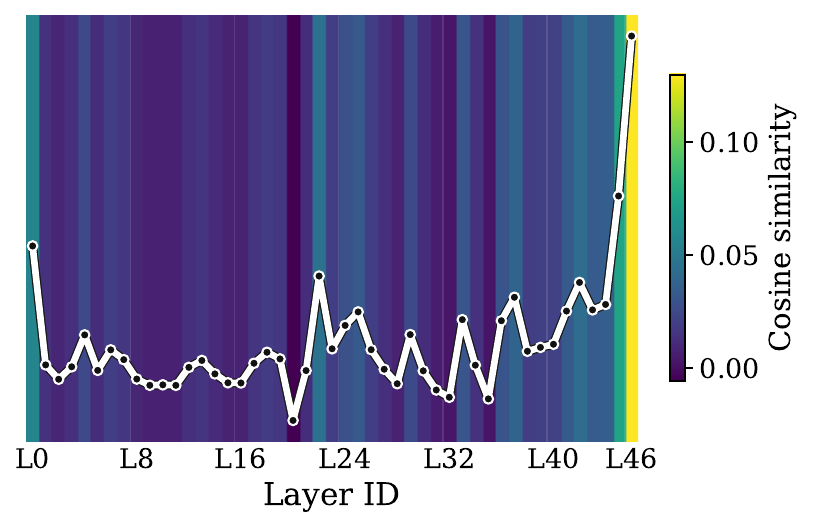}
        \caption{Expert similarity analysis for Qwen3-VL.}
        \label{fig:qwen3vl_mmstar_profile}
    \end{subfigure}
    \caption{Expert similarity under sparse upcycling. Left: layer-wise expert similarity across LibMoE-VLM methods; right: expert similarity profile for Qwen3-VL on MMStar.}
    \label{fig:expert_similarity_combined}
\end{figure*}
In sparse upcycling, experts are initialized by duplicating dense MLP layers, raising the question of whether they remain similar or progressively specialize during training. To examine this, Figure~\ref{fig:layer_wise_diversity} reports the layer-wise cosine similarity of expert output weights, where values close to zero indicate greater divergence between expert outputs, while Figure~\ref{fig:qwen3vl_mmstar_profile} provides a corresponding profile for Qwen3-VL on MMStar.

We find that most MoE variants exhibit low similarity, indicating that experts diverge and specialize over time despite identical initialization. Similarity is consistently higher at the input and output layers than in intermediate layers, suggesting that boundary layers retain more shared structure while middle layers undergo stronger specialization. This pattern is also clearly observed in Qwen3-VL, whose layer-wise profile shows the same trend of elevated similarity near the input and output layers with reduced similarity in the middle. The effect is most pronounced in MoE++, which employs both copy- and zero-experts and thus encourages greater diversity in intermediate layers. Overall, excessive expert similarity does not persist, supporting the validity of comparing MoE methods under the sparse upcycling regime.

\paragraph{g) How efficient is it to initialize a weight router network for load balancing?}

\begin{figure*}[h]
    \centering
    \includegraphics[width=0.7\textwidth]{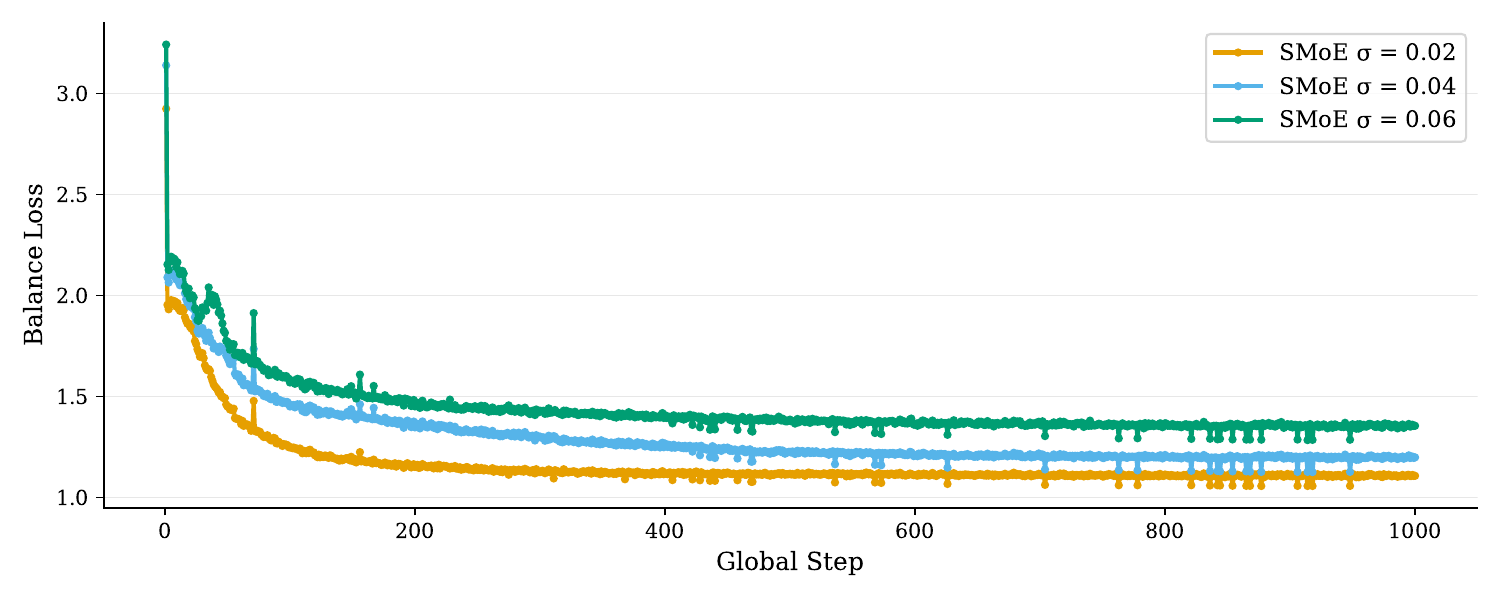}
    \caption{Impact of initialization standard deviation on balance loss dynamics in SMoE models.}
    \label{fig:init_std}
\end{figure*}
Prior works have explored various load-balancing techniques to mitigate the expert imbalance problem in MoE models~\citep{fedus2022switch, wei2024skyworkmoe, wang2024auxiliarylossfree}. However, aggressive balancing may lead to routing collapse~\citep{shazeer2017outrageously}. Recent studies, such as DeepSeekV3~\citep{liu2024deepseekv3}, argue that overly strong load-balancing loss can degrade model performance. To address this, they employed auxiliary-loss-free strategies~\citep{wang2024auxiliarylossfree} and reduced the coefficient of the standard balancing loss~\citep{fedus2022switch}. However, these approaches require introducing additional parameters. 

Interestingly, in our study, we observe that the initialization standard deviation (std) of the router network weights alone can influence balance dynamics even without modifying the loss function. As shown in Figure~\ref{fig:init_std}, we initialize the SMoE router network with different std values ($0.02$, $0.04$, and $0.06$), and train for 1000 steps using the standard balance loss~\citep{fedus2022switch}. The results clearly show that a smaller initialization std leads to better load balance under the same softmax routing configuration. This suggests that subtle changes in the initial logit distribution can enhance routing diversity, offering an alternative or complementary axis to tune alongside both the auxiliary loss coefficient and the initialization std.

%% file: contents/6_Rethinking_LibMoE/rethinking_libmoe.tex
\section{Summary of Key Results}

Across both vision--language sparse upcycling and language-model pretraining, LibMoE shows that performance differences among current SMoE algorithms remain modest once compute and data budgets are matched (Tables~\ref{tab:vlm_moe_results} and~\ref{tab:lm_result}). The multi-seed results in Appendix~\ref{appx:multiseed_lm_results} reinforce this conclusion: benchmark-level rankings may shift slightly, but no method emerges as a consistent winner across regimes. More sophisticated routing mechanisms do not reliably deliver large gains over vanilla SMoE at the scales we study, suggesting that architectural complexity alone is rarely the decisive factor. At the same time, the value of LibMoE lies well beyond benchmark ranking. Our analyses provide a broader and deeper view of MoE routing across architectures, regimes, and scales, revealing how experts stabilize, specialize, allocate weight, and respond to perturbations during training. That said, SharedE-V2 and SharedE-V3 stand out as strong practical choices because they remain competitive in quality while offering better peak memory usage, training efficiency, and inference time than most alternatives (Appendix~\ref{appx:training_time_resource_allocation}). The main contribution of this section, therefore, is not only to distill practical principles, but also to show how analytical evidence can clarify how MoE architectures should be chosen, configured, and improved.

\paragraph{Principle 1: Prioritize methods that remain strong across training regimes.}
A central practical result of LibMoE is that the same small set of methods remains competitive across both vision--language sparse upcycling and from-scratch language-model pretraining, even when their routing dynamics differ. SharedE-V2 and SharedE-V3 are the clearest examples. In vision--language sparse upcycling, they exhibit the most stable routing and among the strongest results in our higher-data VLM setting (Figures~\ref{fig:expert_change_rate} and~\ref{fig:expert_change_rate_sharedexperts}, Table~\ref{tab:vlm_moe_results}). In language-model pretraining, their routing is more volatile early in training, yet they still finish among the top-performing methods in our LM benchmark. The main takeaway, therefore, is not that the two regimes favor entirely different methods, but that strong methods can remain strong across regimes while arriving there through different optimization trajectories.

The regime-specific analyses remain important because they explain \emph{why} these cross-regime winners behave differently during training. In sparse upcycling, pretrained shared experts likely provide stable anchors that allow the router to focus on assigning residual capacity to task-specific experts. When this anchor is weakened in SharedE-Upcyc V2, the stabilizing effect diminishes, supporting this interpretation. In pretraining, by contrast, the router and shared experts must co-adapt from random initialization, creating an early optimization bottleneck rather than a fundamental weakness of the method. This makes shared-expert approaches promising targets for future work on initialization, routing stabilization, and training strategies that better unlock their eventual capacity. A similar pattern appears in XMoE: it remains competitive across settings, but its cosine-normalized routing is more stable in sparse upcycling with smaller learning rates than in pretraining, where sensitivity to scale interacts less favorably with optimization. The broader lesson is therefore twofold: methods that perform well in both regimes deserve priority as practical defaults, and evaluating both regimes is still necessary because similar end performance can mask meaningfully different routing dynamics and optimization bottlenecks.

\paragraph{Principle 2: Specialization and robustness trade off; sharper routing is not always better.}
Our entropy, weight-allocation, router-margin, and routing-perturbation analyses together place the seven algorithms on a spectrum from collaborative to highly decisive routing (Figures~\ref{fig:droptop1}, \ref{fig:entropy_distr_experts}, \ref{fig:weghts_allocation}, and~\ref{fig:router_margin}). TC-MoE lies at the decisive extreme: it concentrates routing mass on a small subset of experts, reaches the sharpest late-layer margins, and exhibits strong specialization. However, it also suffers the largest performance drops under \textit{DropTop1} and \textit{DropTop1$\&$2}, indicating that sharp routing can be brittle when the preferred experts are unavailable. At the other end, the SMoE family maintains much more uniform allocation, a pattern that is also broadly consistent with the large-scale Qwen3-VL-30B-A3B reference. Importantly, vanilla SMoE even improves under \textit{DropTop1} in the VLM setting, suggesting that ``top-1'' routing is not always the most effective use of the available experts. XMoE often occupies a useful middle ground, adapting its degree of specialization across subtasks rather than pushing toward uniformly sharp routing. The resulting lesson is that new routing mechanisms should be judged not by sharpness alone, but by whether they achieve task-adaptive specialization without sacrificing robustness.

\paragraph{Principle 3: Prefer lightweight interventions before adding architectural complexity.}
One of the most practically useful findings in LibMoE is that reducing the router-weight initialization standard deviation from $0.06$ to $0.02$ materially improves early load balancing, without changing the loss or architecture (Figure~\ref{fig:init_std}). Combined with the narrow performance gap in Tables~\ref{tab:vlm_moe_results} and~\ref{tab:lm_result}, this suggests that simple configuration choices can yield gains comparable to those of more elaborate routing changes in our setting. In other words, before proposing a new MoE architecture, it is worth exhausting cheaper levers such as initialization scale, learning-rate interaction with routing normalization, and the active-expert budget. A more complex method should therefore justify itself by delivering clear gains in accuracy, robustness, or effective capacity utilization rather than marginally refining the router in isolation.

Taken together, these three principles convert LibMoE from a benchmark suite into a decision framework. Practitioners can use them to prioritize methods that remain strong across training regimes, to avoid equating sharper routing with better routing, and to exhaust low-cost configuration improvements before adding architectural complexity. Regime-specific analyses still matter, but primarily as diagnostic tools: they help explain why a method succeeds, where its optimization bottlenecks lie, and which interventions are most likely to improve it. In this sense, LibMoE enables researchers to systematically evaluate MoE algorithms across both early-stage regimes, where from-scratch pretraining is unstable and routing is still co-adapting with the experts, and late-stage regimes, where sparse upcycling starts from a stronger pretrained foundation. This dual-regime view offers actionable insights that go beyond standard performance metrics, connecting final accuracy to routing stability, expert specialization, robustness under perturbation, and effective capacity use. For future algorithmic work, these principles also define a stronger evaluation bar: a new method should remain competitive with vanilla SMoE under matched compute across regimes, stay robust under routing perturbation, and demonstrate either more appropriate task-dependent specialization or better use of model capacity. Finally, we emphasize that these conclusions are directly supported at the accessible scales studied in LibMoE. At the same time, the strong correspondence between the behavioral conclusions we observe in LibMoE and those seen in Qwen3-VL-30B-A3B at much larger scale and trillion-token training suggests that LibMoE captures routing phenomena that remain relevant in practice. This, in turn, supports LibMoE as a reliable framework for researchers to explore and evaluate the next generation of MoE methods.

\newpage

%% file: contents/7_Conclusion/concusion.tex
\section{Conclusion}

In this work, we presented LibMoE, a unified and extensible framework that enables reproducible research on Mixture-of-Experts models across language and vision–language domains. By integrating seven recent algorithms within standardized pipelines and offering analytical tools for routing, specialization, and load balancing, LibMoE lowers the barrier for experimentation under resource constraints while providing fair and systematic comparisons. Particularly, all experiments in this study were conducted on a cluster of 4×H100 GPUs, with the longest experiment taking just under 54 hours for VLMs (of which only 35 hours were spent on MoE training) and 43 hours for language model pretraining.
Our comprehensive empirical study reveals that no single method universally dominates, with performance and stability shaped by task type, initialization, and expert design, underscoring the need for flexible and transparent evaluation. We hope that by bridging the gap between algorithmic innovation and practical deployment, LibMoE will not only accelerate MoE research but also provide the community with a principled foundation for building the next generation of scalable, efficient, and interpretable large models.

%% file: contents/A_add_exps/additional_exp.tex

\newpage
\appendix
\begin{center}
{}\textbf{\Large{Supplement to
``LibMoE: A Library for Comprehensive Research on Mixture of Experts in Large Language Models''}}
\end{center}

\definecolor{OverviewPurple}{HTML}{534AB7}

\section{Overview of Benchmarked SMoE Algorithms}
\label{appendix:moe_overview}

LibMoE benchmarks seven representative sparse Mixture-of-Experts (SMoE) algorithms that cover the main design axes explored in the literature: the router scoring function, the expert-selection rule, and structural modifications such as shared experts or zero-computation experts. To make the benchmark easier to follow, we briefly summarize the role of each method below.

\paragraph{Standard SMoE.}
We use the standard sparsely gated MoE layer introduced by \citet{shazeer2017outrageously} and later simplified for Transformer language models by Switch Transformer \citep{fedus2022switch} as our canonical baseline. In this formulation, a conventional dense feed-forward block is replaced by $N$ parallel FFN experts and a lightweight token-level router. For an input token $\vx$, the router computes expert logits with a linear projection, converts them to routing weights with a softmax, selects the top-$K$ experts in our benchmark configuration, and combines their outputs through the selected routing weights. This design is appealing because it is simple, computationally efficient, and widely adopted in practice.

\paragraph{XMoE.}
XMoE is motivated by the representation-collapse analysis of \citet{chi2022representation}, which shows that standard sparse MoE routing can push token hidden states toward expert embeddings, causing routed representations to cluster around experts and reducing representation diversity. To alleviate this issue, XMoE changes the routing score from an unnormalized dot product to a cosine-style affinity: token representations are first projected into a lower-dimensional routing space, and both projected token representations and expert embeddings are L2-normalized before computing routing scores. As a result, expert selection depends primarily on angular similarity rather than vector magnitude, which helps reduce the collapse tendency in the routing representation space. In LibMoE, XMoE provides a useful contrast to conventional dot-product-based sparse routing.

\paragraph{\texorpdfstring{$\sigma$}{sigma}-MoE.}
$\sigma$-MoE replaces softmax competition with sigmoid-based expert scoring, following the sigmoid-gating formulation studied in sparse MoE language models \citep{csordas2023approximating}. Unlike softmax routing, where selected experts compete through a normalized probability distribution, sigmoid routing scores each expert more independently before the selected expert outputs are combined. This makes $\sigma$-MoE a useful reference for isolating how the router scoring function affects routing dynamics, load balancing, and expert utilization.

\paragraph{SharedE-V2 and SharedE-V3.}
These two variants are inspired by the shared-expert design used in DeepSeekMoE and its later DeepSeek-V2/V3 instantiations \citep{dai2024deepseekmoe,liu2024deepseekv2,liu2024deepseekv3}. In this design, part of the expert capacity is assigned to shared experts that are activated for every token, while the remaining routed experts are selected conditionally by the router. This creates an always-on pathway for general-purpose computation alongside token-dependent sparse computation. In LibMoE, SharedE-V2 uses a softmax-based routed component, whereas SharedE-V3 uses a sigmoid-based affinity score with normalization over the selected experts, allowing us to separate the effect of shared experts from the effect of the routing score.

\paragraph{TC-MoE.}
TC-MoE augments the standard top-$K$ paradigm through ternary expert choice \citep{yan2025tc}. Instead of treating an expert as only selected or not selected, it constructs an expanded routing space in which experts are associated with ternary choices from ${-1,0,1}$. This gives the router a richer set of activation patterns than ordinary top-$K$ selection. The method also uses auxiliary objectives tailored to the expanded routing space, including load-balancing and reward-style terms. In our benchmark, TC-MoE represents a structurally expressive routing design aimed at reducing redundant expert usage.

\paragraph{MoE++.}
MoE++ augments the standard FFN expert pool with zero-computation experts \citep{jin2024moe++}. Instead of routing every token to a fixed number of FFN experts, it introduces zero, copy, and constant experts, which respectively discard the input, copy the input as a shortcut, or replace it with a lightweight trainable-vector pathway. This allows different tokens to use different amounts and types of computation, while gating residuals help the router incorporate pathway information from previous layers. In our benchmark, MoE++ represents a heterogeneous routing design that targets a better trade-off between computational efficiency and model expressivity.

\paragraph{Design dimensions.}
Taken together, the seven algorithms vary along three principal axes: (i) the \emph{scoring function} used by the router (softmax, sigmoid, or cosine-based normalization), (ii) the \emph{selection policy} that determines which experts are activated, and (iii) the presence of \emph{structural augmentation}, such as shared experts or zero-computation experts. Covering all three axes is important for LibMoE because it allows us to compare not only end-task performance, but also how different design choices affect routing stability, specialization, and expert utilization.

\section{Sensitivity to the Number of Active Experts}
\label{subappx:topk_sensitivity}

\begin{figure}[ht]
    \centering
    \includegraphics[width=\linewidth]{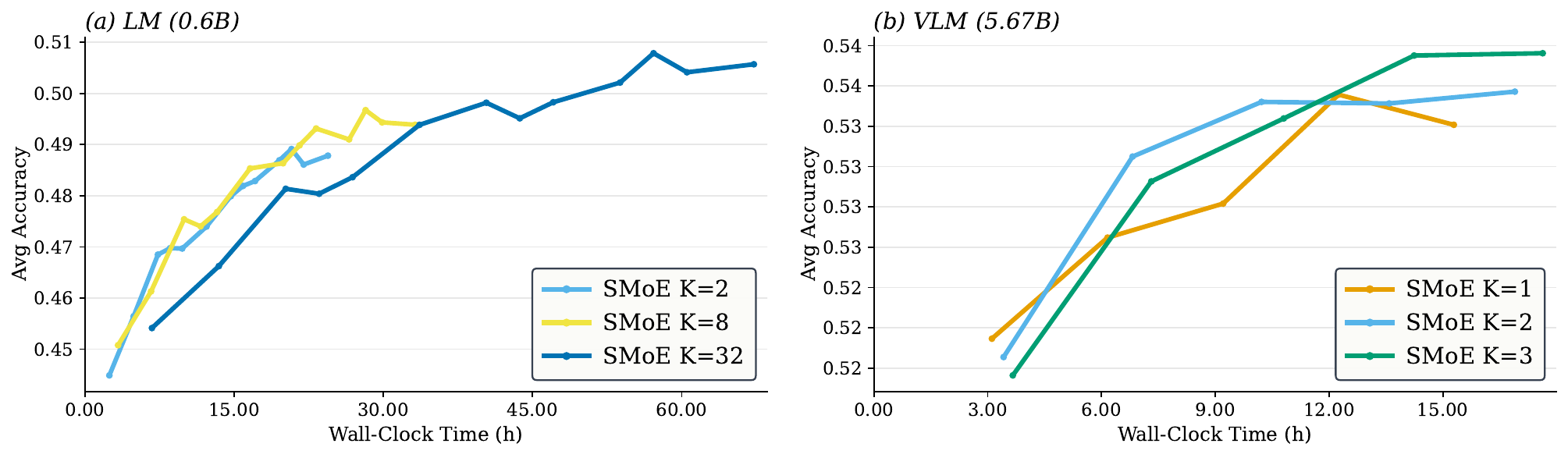}
    \caption{Average validation performance as a function of wall-clock time under different numbers of active experts $K$ for a standard SMoE model in language modeling (left) and vision-language training (right, 5.67B). For the VLM panel, training is performed on LLaVA-665K and validation performance is averaged over nine benchmarks, excluding MathVista and Hallusion Bench.}
    \label{fig:topk_overtime}
\end{figure}

In this section, we provide a supplementary sensitivity analysis on the number of active experts. Our main benchmark fixes $K$ in each setting to keep comparisons across routing algorithms computationally controlled, so Figure~\ref{fig:topk_overtime} is intended primarily to illustrate how the training dynamics change when $K$ is varied for a representative SMoE model in language modeling and vision-language training.

In the language modeling setting, we compare $K \in \{2, 8, 32\}$. Within this setup, the curves suggest a trade-off between time-to-quality and the best validation performance reached later in training. Smaller or moderate routing budgets ($K{=}2$ and $K{=}8$) improve more quickly during the earlier stages, whereas the $K{=}32$ run progresses more slowly but continues improving over a longer horizon. Among the smaller settings, $K{=}8$ appears to offer a reasonable compromise in this experiment, while the largest setting finishes highest by the end of training.

In the vision-language setting, we compare $K \in \{1, 2, 3\}$. The dependence on $K$ appears weaker here than in the language-modeling case. Moving from $K{=}1$ to $K{=}2$ improves the trajectory in our run, while increasing from $K{=}2$ to $K{=}3$ yields a smaller difference. The $K{=}3$ curve ends slightly above $K{=}2$, but the two remain close for most of training.

Overall, these plots suggest that the effect of top-$K$ is regime-dependent in our benchmark. In the language-modeling setting, smaller $K$ values appear to improve training efficiency, whereas larger $K$ may help the final validation score when longer training is allowed. In the VLM setting, the gains beyond a small active-expert budget appear limited. We view these observations as descriptive rather than prescriptive, since the preferred choice of $K$ is likely to depend on factors such as scale, data, training budget, and deployment constraints. 


\section{Multi-Seed Results for the Main Benchmark Tables}
\label{appx:multiseed_lm_results}

To complement the single-run results in the main benchmark tables, we report multi-seed mean$\pm$std results for both the LLaVA-665K VLM setting and the small-model language-modeling setting. We use three random seeds: 42, 128, and 456. This presentation provides a more complete view of performance by showing both the central tendency and the run-to-run variation of each method. Overall, the multi-seed tables are broadly consistent with the main-text picture: under matched budgets, performance differences are generally modest, and the apparent ordering depends on the benchmark, the aggregate metric, and the training regime. We therefore treat this section as descriptive evidence rather than as support for a universally superior SMoE variant.

\paragraph{Vision-language modeling (LLaVA-665K).}
Table~\ref{tab:vlm_result_multiseed_665k} provides the corresponding multi-seed results for the 665K sparse-upcycling benchmark. The overall spread is even smaller in this setting. XMoE attains the best mean on the two aggregate metrics (AVG Acc and AVG Rank), but only by a narrow margin, while the best mean on individual benchmarks is distributed across multiple methods. We therefore interpret the VLM multi-seed results conservatively: in this regime, the relative ordering of methods is benchmark-dependent, and no single variant separates cleanly from the others.

\begin{table*}[ht]
\centering
\small
\caption{Multi-seed mean$\pm$std results for the main VLM benchmark in the LLaVA-665K setting. Bold values indicate the best mean in each row; ties are boldfaced for all methods attaining the same best mean.}
\resizebox{\textwidth}{!}{
\begin{tabular}{lccccccc}
\toprule
Metric & SMoE & XMoE & $\sigma$-MoE & SharedE-V2 & SharedE-V3 & TC-MoE & MoE++ \\
\midrule
AI2D $\uparrow$ & 64.78 $\pm$ 0.73 & 65.74 $\pm$ 0.38 & 65.71 $\pm$ 0.61 & 65.53 $\pm$ 0.66 & 65.71 $\pm$ 0.20 & 65.22 $\pm$ 0.45 & \textbf{65.83 $\pm$ 0.69} \\
Text VQA $\uparrow$ & 41.57 $\pm$ 0.26 & 41.81 $\pm$ 0.20 & 41.78 $\pm$ 0.29 & \textbf{41.93 $\pm$ 0.09} & 41.78 $\pm$ 0.29 & 40.82 $\pm$ 0.41 & 40.99 $\pm$ 0.44 \\
GQA $\uparrow$ & 61.36 $\pm$ 0.28 & 61.44 $\pm$ 0.37 & 61.70 $\pm$ 0.07 & 61.55 $\pm$ 0.28 & \textbf{61.85 $\pm$ 0.51} & 61.34 $\pm$ 0.17 & 60.93 $\pm$ 0.41 \\
MM Bench $\uparrow$ & \textbf{72.45 $\pm$ 0.42} & 72.11 $\pm$ 0.35 & 71.16 $\pm$ 0.43 & 71.62 $\pm$ 0.61 & 71.99 $\pm$ 0.54 & 71.25 $\pm$ 0.30 & 71.39 $\pm$ 0.31 \\
Hallusion Bench $\uparrow$ & 41.85 $\pm$ 0.28 & 42.27 $\pm$ 0.42 & 41.71 $\pm$ 1.15 & 42.13 $\pm$ 0.58 & 41.89 $\pm$ 0.43 & \textbf{42.41 $\pm$ 0.30} & 41.99 $\pm$ 0.95 \\
Math Vista $\uparrow$ & 30.03 $\pm$ 0.50 & 30.73 $\pm$ 0.78 & 30.00 $\pm$ 0.79 & 30.77 $\pm$ 0.91 & 30.40 $\pm$ 0.20 & 30.20 $\pm$ 0.90 & \textbf{30.97 $\pm$ 0.76} \\
MMMU $\uparrow$ & 42.33 $\pm$ 0.59 & 41.45 $\pm$ 0.30 & 41.48 $\pm$ 0.36 & \textbf{42.96 $\pm$ 0.32} & 42.07 $\pm$ 0.46 & 41.85 $\pm$ 1.00 & 41.78 $\pm$ 1.13 \\
MMStar $\uparrow$ & \textbf{42.53 $\pm$ 0.30} & 40.82 $\pm$ 0.76 & 41.40 $\pm$ 1.36 & 40.92 $\pm$ 0.74 & 41.71 $\pm$ 0.04 & 40.93 $\pm$ 0.72 & 40.79 $\pm$ 0.60 \\
Pope $\uparrow$ & \textbf{86.88 $\pm$ 0.22} & 86.73 $\pm$ 0.09 & 86.66 $\pm$ 0.32 & 86.57 $\pm$ 0.22 & \textbf{86.88 $\pm$ 0.41} & 86.66 $\pm$ 0.14 & 86.52 $\pm$ 0.09 \\
MME $\uparrow$ & 60.57 $\pm$ 1.49 & \textbf{60.86 $\pm$ 0.55} & 60.71 $\pm$ 0.70 & 60.59 $\pm$ 1.03 & 60.77 $\pm$ 0.33 & 60.78 $\pm$ 1.06 & 59.63 $\pm$ 0.51 \\
MME RW $\uparrow$ & 32.41 $\pm$ 0.33 & \textbf{33.16 $\pm$ 0.54} & 32.92 $\pm$ 0.53 & 32.31 $\pm$ 0.60 & 31.86 $\pm$ 0.39 & 32.27 $\pm$ 0.99 & 31.94 $\pm$ 0.60 \\
OCR Bench $\uparrow$ & 31.93 $\pm$ 0.21 & \textbf{33.00 $\pm$ 0.35} & 32.83 $\pm$ 0.31 & 32.00 $\pm$ 0.72 & 32.40 $\pm$ 0.35 & 32.60 $\pm$ 0.46 & 31.43 $\pm$ 0.40 \\
\midrule
AVG Acc $\uparrow$ & 50.72 $\pm$ 0.17 & \textbf{50.84 $\pm$ 0.19} & 50.67 $\pm$ 0.29 & 50.74 $\pm$ 0.04 & 50.80 $\pm$ 0.13 & 50.53 $\pm$ 0.35 & 50.35 $\pm$ 0.15 \\
AVG Rank $\downarrow$ & 3.78 $\pm$ 0.34 & \textbf{3.19 $\pm$ 0.10} & 3.74 $\pm$ 0.67 & 3.93 $\pm$ 0.65 & 3.58 $\pm$ 0.78 & 4.58 $\pm$ 0.93 & 5.20 $\pm$ 0.16 \\
\bottomrule
\end{tabular}}
\label{tab:vlm_result_multiseed_665k}
\end{table*}

\paragraph{Language modeling (0.15B).}
Table~\ref{tab:lm_result_multiseed} reports the multi-seed results for the main small-model language-modeling benchmark. Mean performance remains tightly clustered across most methods in this setting. TC-MoE achieves the best mean on both aggregate metrics (AVG Acc and AVG Rank), while MoE++, SharedE-V2, and $\sigma$-MoE remain close on the aggregate means. XMoE is weaker on the aggregate metrics in this particular setup, but the best mean on individual benchmarks is still distributed across several methods. We therefore view the language-modeling results as indicating a competitive leading group rather than a decisive separation between methods.

\begin{table*}[ht]
\centering
\small
\caption{Multi-seed mean$\pm$std results for the main language-modeling benchmark in the small-model setting (0.15B). Bold values indicate the best mean in each row.}
\resizebox{\textwidth}{!}{
\begin{tabular}{lccccccc}
\toprule
Metric & SMoE & XMoE & $\sigma$-MoE & SharedE-V2 & SharedE-V3 & TC-MoE & MoE++ \\
\midrule
PPL $\downarrow$ & 13.65 $\pm$ 0.02 & 14.28 $\pm$ 0.26 & 13.63 $\pm$ 0.02 & 13.58 $\pm$ 0.08 & 13.56 $\pm$ 0.12 & 13.47 $\pm$ 0.04 & \textbf{13.46 $\pm$ 0.08} \\
LAMBADA $\uparrow$ & \textbf{25.70 $\pm$ 0.39} & 23.97 $\pm$ 0.85 & 25.56 $\pm$ 0.76 & 25.25 $\pm$ 0.98 & 25.27 $\pm$ 0.79 & 25.67 $\pm$ 0.31 & 25.69 $\pm$ 0.29 \\
BLiMP $\uparrow$ & \textbf{77.31 $\pm$ 1.15} & 75.74 $\pm$ 0.75 & 76.80 $\pm$ 0.58 & 77.13 $\pm$ 0.25 & 77.12 $\pm$ 0.27 & 76.96 $\pm$ 0.39 & 76.96 $\pm$ 0.62 \\
CBT $\uparrow$ & 84.45 $\pm$ 0.24 & 83.75 $\pm$ 0.34 & 84.39 $\pm$ 0.43 & 84.41 $\pm$ 0.09 & 84.39 $\pm$ 0.26 & \textbf{84.62 $\pm$ 0.17} & 84.60 $\pm$ 0.20 \\
HellaSwag $\uparrow$ & 29.17 $\pm$ 0.25 & 29.02 $\pm$ 0.29 & 28.99 $\pm$ 0.22 & 29.14 $\pm$ 0.21 & \textbf{29.30 $\pm$ 0.23} & 29.02 $\pm$ 0.23 & 29.15 $\pm$ 0.13 \\
PIQA $\uparrow$ & 58.14 $\pm$ 0.19 & 57.47 $\pm$ 0.72 & 58.90 $\pm$ 0.52 & 59.05 $\pm$ 0.99 & 58.71 $\pm$ 0.38 & \textbf{59.10 $\pm$ 0.08} & 58.81 $\pm$ 0.36 \\
ARC-Easy $\uparrow$ & 32.61 $\pm$ 0.12 & 32.66 $\pm$ 0.36 & 32.97 $\pm$ 0.28 & \textbf{33.67 $\pm$ 0.59} & 32.60 $\pm$ 0.18 & 33.16 $\pm$ 0.52 & 33.39 $\pm$ 0.29 \\
RACE $\uparrow$ & 30.26 $\pm$ 0.15 & 29.97 $\pm$ 0.31 & \textbf{30.98 $\pm$ 0.06} & 30.95 $\pm$ 0.28 & 30.37 $\pm$ 0.27 & 30.62 $\pm$ 0.15 & 30.30 $\pm$ 0.57 \\
SIQA $\uparrow$ & 35.09 $\pm$ 0.92 & 35.93 $\pm$ 0.49 & 35.79 $\pm$ 0.81 & 35.36 $\pm$ 0.20 & 35.88 $\pm$ 0.41 & \textbf{36.17 $\pm$ 0.43} & 35.43 $\pm$ 0.51 \\
CommonSenseQA $\uparrow$ & 24.62 $\pm$ 0.13 & 24.57 $\pm$ 0.62 & 24.87 $\pm$ 0.05 & 25.12 $\pm$ 0.71 & 24.98 $\pm$ 0.49 & \textbf{25.91 $\pm$ 0.45} & 25.33 $\pm$ 0.79 \\
\midrule
AVG Acc $\uparrow$ & 44.15 $\pm$ 0.03 & 43.67 $\pm$ 0.18 & 44.36 $\pm$ 0.20 & 44.45 $\pm$ 0.18 & 44.29 $\pm$ 0.12 & \textbf{44.58 $\pm$ 0.06} & 44.41 $\pm$ 0.18 \\
AVG Rank $\downarrow$ & 4.70 $\pm$ 0.15 & 6.02 $\pm$ 0.39 & 3.83 $\pm$ 0.50 & 3.37 $\pm$ 0.53 & 3.75 $\pm$ 0.48 & \textbf{3.08 $\pm$ 0.24} & 3.25 $\pm$ 0.49 \\
\bottomrule
\end{tabular}}
\label{tab:lm_result_multiseed}
\end{table*}

\section{Analysis of a Hybrid Dense--Sparse Model}
\label{appx:hybrid_dense_sparse}

\input{contents/A_add_exps/tabs/hybrid_dense_sparse_vlm}

To examine whether a hybrid dense--sparse layout is beneficial in our setting, we conduct an additional experiment in the 5.67B-scale VLM benchmark on LLaVA-665K, comparing a hybrid architecture against its fully sparse counterpart. Following the design of DeepSeek-V3\citep{liu2024deepseekv3}, the hybrid variant keeps the first three FFN layers dense and replaces only the remaining FFN layers with MoE layers. As shown in Table~\ref{tab:hybrid_dense_sparse_vlm}, Hybrid SharedE-V3 underperforms the fully sparse SharedE-V3, despite providing only a modest reduction in training time (16h versus 16h47m). In our benchmark, this comparison suggests that the performance lost by removing MoE layers from the bottom of the network outweighs the corresponding efficiency gain.

One plausible explanation is that early layers already play an important role in establishing token-dependent specialization. Dense FFNs apply the same MLP to every token, whereas MoE layers provide larger conditional capacity through token-wise expert selection. Replacing the lower MoE layers with dense ones may therefore weaken specialization from the start of the network and reduce the cumulative benefit of sparse routing in later layers.

Importantly, this observation is consistent with prior work \citep{Zhong2024MoExtendTN, lin2024moe}, which likewise reports that introducing experts only in the second half of the model underperforms applying experts across all layers. Taken together, these findings suggest that retaining dense early layers can reduce the benefits of sparse expert routing relative to a fully sparse allocation. Thus, while hybrid dense--MoE architectures remain an important design direction, our controlled experiment indicates that they should not be assumed to be universally superior, but instead evaluated under matched settings.

\section{Experiment Settings}
\label{appendix:exp_settings}

\subsection{Vision Language Model}
\label{appendix:exp_setting_vlm}
\begin{figure*}[ht]
    \centering
    \includegraphics[width=\linewidth]{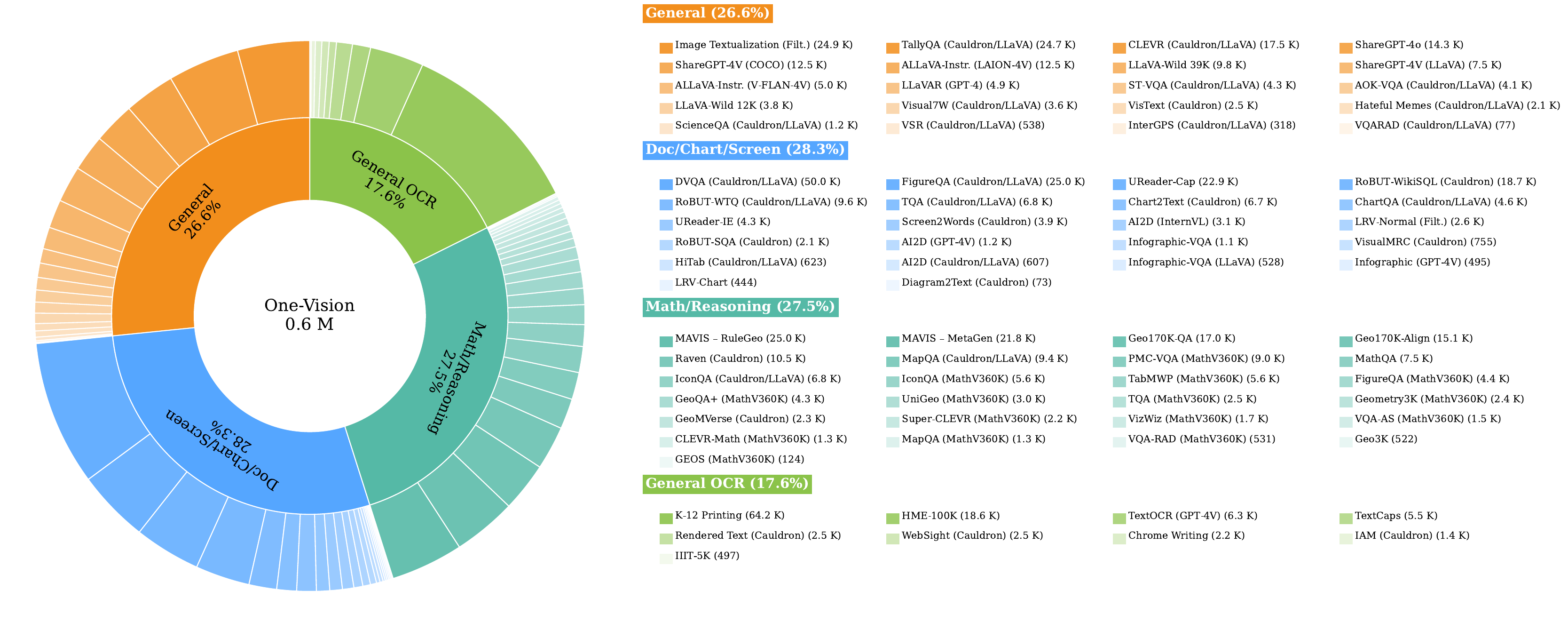}
    \caption{Visualization of dataset subsets from the OneVision dataset in LibMoE.}

    \label{fig:onevision}
\end{figure*}

\begin{figure*}[ht]
    \centering
    \includegraphics[width=\linewidth]{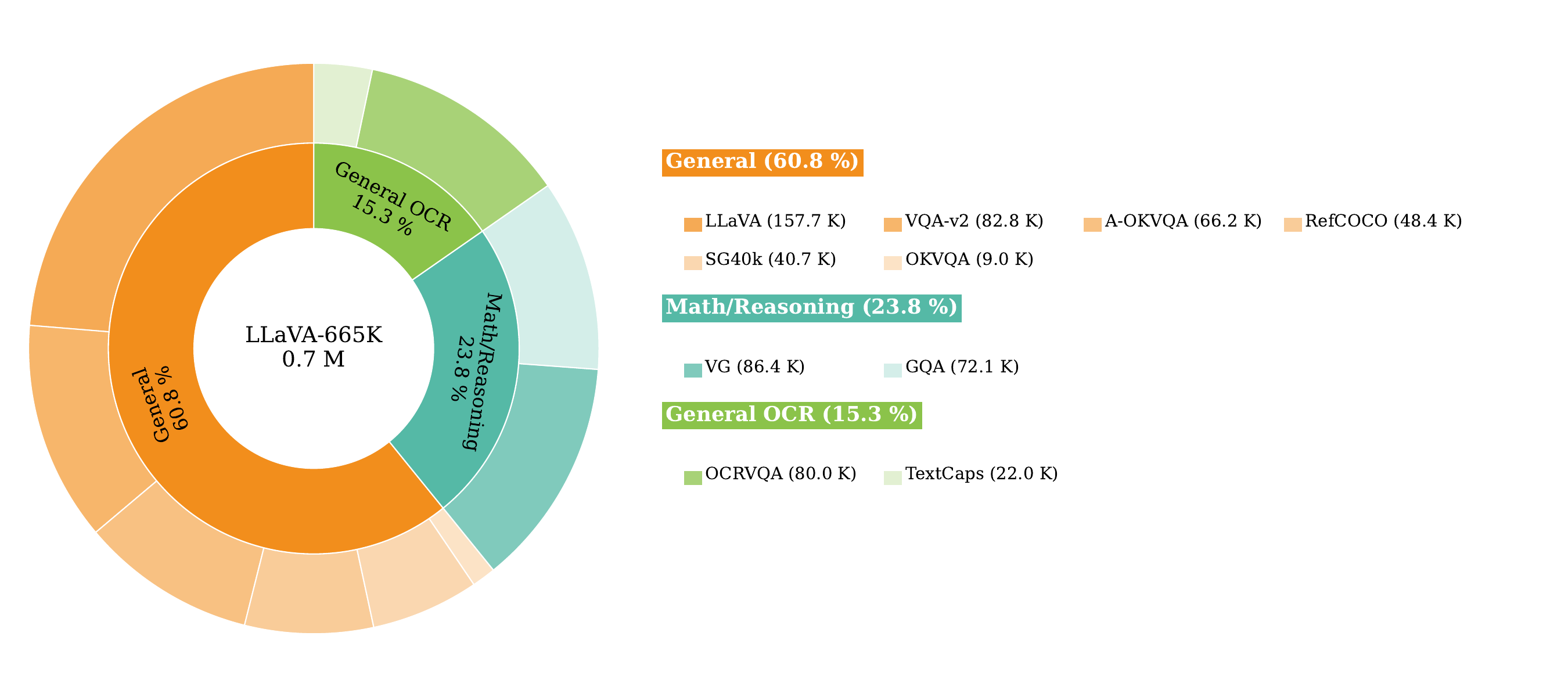}
    \caption{Visualization of dataset subsets from the LLAVA-665K dataset in LibMoE.}
    \label{fig:llava}
\end{figure*}

\paragraph{Datasets.}  
We adopt the vision-language pretraining task~\citep{Lu2019ViLBERTPT} and follow the CUMO framework~\citep{li2024cumo} to upcycle the LLaVA model~\citep{liu2023llava}, enabling systematic evaluation of various SMoE algorithms. In the initial \textbf{dense training stage}, we first initialize the MLP connector using the LLaVA-558K dataset~\citep{liu2023llava}, then jointly train all three model components (image encoder, language model, and connector) on the ALLaVA dataset~\citep{chen2024allava}. This yields a dense checkpoint that serves as the starting point for sparse upcycling.

In the subsequent SMoE training stage, we consider two training settings:
\begin{itemize}
    \item LLaVA-665K~\citep{liu2023llava}, a standard and widely used benchmark in the community;
    \item A hybrid dataset with 1M2 samples, constructed by combining LLaVA-665K with OneVision~\citep{li2024llava}, where we uniformly sample 25\% from each OneVision sub-benchmark to ensure broad domain coverage.
\end{itemize}

We visualize the data distribution across categories for both datasets in Figure~\ref{fig:llava} and Figure~\ref{fig:onevision}. All SMoE algorithms are trained using the same datasets and configurations to ensure fair benchmarking. For further details on dataset construction and training stage objectives, we refer readers to~\citet{liu2023llava,li2024cumo}.

\paragraph{Evaluation Benchmarks and Metrics} LibMoE employs a set of popular benchmarks for vision-language models, including AI2D~\citep{Kembhavi2016ADI}, Text VQA Validation~\citep{Singh2019TowardsVM}, GQA~\citep{Hudson2019GQAA}, Hallusion Benchmark~\citep{Guan2023HallusionBenchAA}, MathVista~\citep{Lu2023MathVistaEM}, MMBenchEN~\citep{Liu2023MMBenchIY}, MME~ \citep{Fu2023MMEAC}, MMMU Validation\citep{Yue2023MMMUAM}, MMStar \citep{Chen2024AreWO}, POPE \citep{Li2023EvaluatingOH}, MME-Real World\citep{zhang2024mme} and OCR Bench\citep{liu2024ocrbench}. We carefully choose these benchmarks to assess the model across several vision-language capabilities such as perception, reasoning, OCR, instruction following, and more. Beyond the performance on standard benchmarks, we analyze the algorithms holistically by evaluating their generalization throughout training, expert selection behaviors, and expert specialization, which we report in Section~\ref{sec:analysis}. For benchmarks requiring GPT-based evaluation, such as MathVista and HallusionBench, we use GPT-4o-mini, version 2024-07-18.

\paragraph{Model Architecture.}
We adopt \textsc{Phi-3.5 Mini}~\citep{abdin2024phi3tr} as the language model and \textsc{SigLIP-SO400M}~\citep{zhai2023sigmoid} as the vision encoder two widely used backbones in recent multimodal systems.  
During the VIT stage, we upcycle the dense MLP blocks in both the vision encoder and connector into sparse MoE layers, each composed of $N_E{=}6$ experts with top-$K{=}3$ routing.

Additionally, Table~\ref{tab:param_breakdown} reports the parameter breakdown across key components. The full model contains approximately 5.67 billion parameters, including a 3.82B LLM backbone, a 1.75B vision encoder equipped with MoE layers, and 99M parameters dedicated to expert MLPs. To ensure fair comparison across routing methods, all models are configured under the same total parameter budget.

\begin{table}[ht]
\centering
\small
\renewcommand{\arraystretch}{1.5}
\caption{Parameter counts for each major component. Total model size is 5.67B.}
\label{tab:param_breakdown}
\begin{tabular}{lS[table-format=1.3]}
\toprule
\textbf{Component} & \textbf{Parameters} \\
\midrule
Vision Encoder - MoE & \num{1.75} B \\
MLP Connector - MoE            & \num{0.099} B \\
LLM                 & \num{3.82} B \\
\midrule
\textbf{Total}      & \num{5.67} B \\
\bottomrule
\end{tabular}
\end{table}
\subsection{Language Model Pretrain}  \label{subappx:lm_setting}

\paragraph{Dataset.} We conduct our experiments on the widely-used SlimPajama dataset \citep{cerebras2023slimpajama}, a high-quality corpus curated from RedPajama \citep{together2023redpajama}, specifically designed for training large language models (LLMs). SlimPajama has become a standard choice for open LLM research and has been employed in the development of several influential models, including TinyLlama~\citep{zhang2024tinyllama} and BTLM~\citep{Cerebras_2025}, as well as in a range of recent empirical studies~\citep{agarwalla2024enabling, gupta2023continual}. Its adoption across diverse works underscores its value for benchmarking and advancing LLM research.

\paragraph{Tokenizer and Model Architecture.} We adopt the SentencePiece tokenizer \citep{kudo2018sentencepiece} with byte-pair encoding (BPE), which provides a balance between subword granularity and vocabulary efficiency. We generally follow Switch Transformer \citep{fedus2022switch} for our model architecture, adopting a standard Transformer backbone with sparsely activated Mixture-of-Experts (MoE) layers. Each MoE layer replaces the conventional feedforward sublayer and comprises $N$ expert networks (FFNs) and a router mechanism. Table~\ref{tab:lm_model_setting} summarizes the comprehensive set of hyperparameters and configurations for both scales and different model variants evaluated in our experiments.

\paragraph{Load Balancing Loss and Router Z-Loss.} We adopt the standard load balancing loss from Switch Transformer \citep{fedus2022switch} to penalize imbalances in routing decisions, thereby encouraging more uniform expert utilization. This auxiliary term has been shown to mitigate expert under-utilization and promote balanced load distribution, which improves both convergence and generalization in MoE models. In contrast, we do not incorporate the router z-loss in our experiments \citep{zoph2022st}. Following common practice, we set the weight of the load balancing loss to $\alpha = 0.01$. Additional experimental details and design considerations for the auxiliary load balancing loss are provided in Appendix~\ref{appx:load_balancing}.

\paragraph{Evaluation pipeline.} We evaluate our implemented models with the Perplexity score (PPL) and zero-shot performance with nine different downstream tasks: LAMBADA \citep{paperno2016lambada}, BLiMP \citep{warstadt2020blimp}, Children’s Book Test \citep{hill2015goldilocks}, HellaSwag \citep{zellers2019hellaswag}, PIQA \citep{bisk2020piqa},  ARC-Easy \citep{clark2018think}, RACE \citep{lai-etal-2017-race}, SIQA \citep{sap2019socialiqa} and CommonSenseQA \citep{talmor2018commonsenseqa}. For LAMBADA, we use the detokenized version from OpenAI, and we evaluate the top-1 accuracy of the last word (it can span multiple tokens; here we use greedy decoding). For CBT, BLiMP, and RACE, we measure the accuracy of each task and report the average accuracy of the tasks.
\newpage
\section{Hyperparameter Setting}

\label{appendix:hyper_settings}

\subsection{Vision-Language Model}
\label{appendix:hyper_settings_vlm}

\begin{table}[ht]
\centering
\small
\renewcommand{\arraystretch}{1.5}
\caption{Hyperparameter settings across the three training stages of \textsc{Phi-3.5 Mini}. MoE routing is only applied during the final VIT phase.}
\label{tab:hyperparams_vlm}
\begin{tabular}{lccc}
\toprule
\textbf{Hyperparameter}           & \textbf{PT}     & \textbf{PFT}    & \textbf{VIT}    \\
\midrule
Learning rate                    & 1e-3            & 2e-6            & 4e-6            \\
Learning rate schedule           & Cosine          & Cosine          & Cosine          \\
Batch size per GPU               & 64              & 6               & 4               \\
GPUs                             & 4$\times$H100   & 4$\times$H100   & 4$\times$H100   \\
ZeRO optimization                & ZeRO-2          & ZeRO-2          & ZeRO-3          \\
Optimizer                        & AdamW           & AdamW           & AdamW           \\
MLP parameters                   & Trained         & Trained         & Trained         \\
Vision encoder                   & Frozen          & Trained         & Trained         \\
Language model                   & Frozen          & Trained         & Trained         \\
MoE blocks                       & No              & No              & Yes             \\
Balance loss coefficient         & 0.0             & 0.0             & 0.01            \\
Z-loss coefficient               & 0.0             & 0.0             & 0.001           \\
Maximum tokens                   & 2048            & 2048            & 2048            \\
\bottomrule
\end{tabular}
\end{table}

\paragraph{Hyperparameter Settings and Training Stages.} Table~\ref{tab:hyperparams_vlm} summarizes the key hyperparameter configurations across the three sequential training stages of \textsc{Phi-3.5 Mini}: Pretraining (PT), Pre-FineTuning (PFT), and Visual Instruction Tuning (VIT). All stages are trained on 4$\times$H100 GPUs with consistent token lengths, optimizers, and learning rate schedules to ensure comparability. The Mixture-of-Experts (MoE) routing mechanism is activated only during the VIT stage, where multimodal specialization is required. Following the initialization strategy used in the official GPT-2 implementation,\footnote{\url{https://github.com/openai/gpt-2}} router parameters are sampled from a normal distribution $\mathcal{N}(0, 0.02^2)$, i.e., with standard deviation $0.02$, using a fixed random seed of 42 to ensure reproducibility. All expert parameters are trained during VIT, while only the router is initialized from scratch. This design yields a controlled and fair evaluation protocol across different sparse routing strategies.

\begin{table}[ht]
\centering
\small
\renewcommand{\arraystretch}{1.2}

\setlength{\tabcolsep}{6pt}
\caption{MoE configuration details across different methods, where $N$ denotes the total number of experts, $K$ the number of routed (active) experts per token, and $N_s$ the number of shared experts.}
\label{tab:moe_config_vlm}
\begin{tabular}{lccc}
\toprule
\textbf{MoE Method} & $\boldsymbol{N}$ & $\boldsymbol{K}$ & $\boldsymbol{N_s}$ \\
\midrule
SMoE     & 6  & 3 & 0 \\
SharedE  & 6  & 2 & 1 \\
MoE++    & 8  & 3 & 0 \\
TC-MoE   & 15 & 3 & 0 \\
\bottomrule
\label{tab:moe_config_e_vlm}
\end{tabular}
\end{table}

\paragraph{MoE Architecture Configurations.} Table ~\ref{tab:moe_config_e_vlm} summarizes the MoE configurations used across different methods, where $N$ denotes the total number of experts, $K$ the number of routed (active) experts per token, and $N_s$ the number of shared experts.
For standard routing-based methods, including SMoE, $\sigma$-MoE, and XMoE, we adopt a common configuration with $N=6$ experts and top-$K=3$ routing. For shared-expert architectures, we follow the VLM setting with one shared expert ($N_s=1$) and reduce the number of routed experts to $K=2$ to maintain comparable computational cost. For MoE++, we follow~\citet{jin2024moe++} and augment the expert set by introducing zero-computation experts, resulting in a total of $N=8$ experts while keeping $K=3$. For TC-MoE, we follow~\citet{yan2025tc}, where the effective expert pool is expanded via ternary compositions, yielding a total of $2N + K$ experts; in our configuration, this corresponds to $N=15$ with $K=3$.

\subsection{Language Modeling}  \label{subappx:lm_hyperparams}

\renewcommand{\arraystretch}{0.5}
\begin{table}[ht!]
  \centering
  \small
  \caption{Comprehensive Model Configurations for Pre-train LLM. SMoE refers to settings applied for Vanilla SMoE, $\sigma$-MoE and XMoE, whereas SharedE corresponds to configurations used for SharedE-V2 and SharedE-V3 models.}
  \label{tab:lm_model_setting}
  \setlength{\tabcolsep}{4pt}
  \renewcommand{\arraystretch}{1}

  \resizebox{\textwidth}{!}{%
    \begin{tabular}{*{12}{c}}
      \toprule
     \makecell{Scale} & \makecell{Model} & \makecell{\# params} & \makecell{\# act.\\params} & \makecell{\# trained \\ tokens} & \makecell{$d_{\mathrm{model}}$} & \makecell{H} & \makecell{$d_{\mathrm{head}}$} & \makecell{$N$} & \makecell{$K$} & \makecell{$N_s$} & \makecell{Expert \\ dim} \\
     \midrule
      
    \multirow{4}{*}{Small} & SMoE & \multirow{4}{*}{0.15B} & \multirow{4}{*}{36M} & \multirow{4}{*}{6.55B} & \multirow{4}{*}{512} & \multirow{4}{*}{12} & \multirow{4}{*}{82} & 66 & 8 & 0 & \multirow{4}{*}{128} \\
     & SharedE &  &  &  &  &  &  & 66 & 6 & 2 &  \\
     & MoE++ &  &  &  &  &  &  & 66 + 8 & 8 & 0 &  \\
     & TC-MoE &  &  &  &  &  &  & 66 * 2 + 8 & 8 & 0 & \\
    \midrule
    \multirow{4}{*}{Large} & SMoE & \multirow{4}{*}{0.68B} & \multirow{4}{*}{131M} & \multirow{4}{*}{26.2B} & \multirow{4}{*}{1024} & \multirow{4}{*}{16} & \multirow{4}{*}{128} & 66 & 8 & 0 & \multirow{4}{*}{256} \\
     & SharedE &  &  &  &  &  &  & 66 & 6 & 2 &  \\
     & MoE++ &  &  &  &  &  &  & 66 + 8 & 8 & 0 & \\
     & TC-MoE &  &  &  &  &  &  & 66 * 2 + 8 & 8 & 0 & \\
  \bottomrule   
  \end{tabular}%
  }
\end{table}
\paragraph{Hyperparameter Settings and Training Stages.}
Table~\ref{tab:lm_model_setting} summarizes the key hyperparameters, covering model dimensionality, number of attention heads, expert counts, and routing strategies. The small-scale setting processes 6.55B tokens with model dim $d_{model}=512$ and number of attention heads $H=12$, while the large-scale setting extends to 26.2B tokens with $d_{model}=1024$ and $H=16$. Each variant differs in expert number ($N$), routing capacity ($K$), and warmup strategies, with expert dimensions set to 128 for small and 256 for large models. These standardized yet diverse configurations ensure a balanced comparison across MoE algorithms while reflecting realistic large-scale training regimes.


\begin{table}[ht]
\centering
\small
\renewcommand{\arraystretch}{1.5}
\caption{\small Training hyperparameter settings for LibMoE across two model scales (0.15B and 0.68B parameters) in the language modeling task.}
\label{tab:hyperparams_lm_training}
\begin{tabular}{lcc}
\toprule
\textbf{Hyperparameter}      & \textbf{0.15B}     & \textbf{0.68B}  \\
\midrule
Learning rate & 2.5e-4 & 2.5e-4 \\
LR schedule & Cosine & Cosine \\
$N_{warmup}$ & 0 & 4000 \\
Min LR multiplier & 0.1 & 0.1 \\
Optimizer & AdamW & AdamW \\
Weight decay & 0.01 & 0.01 \\
Gradient clip ($\kappa$) & 0.1 & 0.25 \\
Dropout & No & No \\
Batch size (per device) & 16 & 16 \\
Total batch size & 64 & 64 \\
Sequence length & 1024 & 1024 \\
Training steps & 100k & 400k \\
Validation ratio & 0.5\% & 0.5\% \\
Precision & AMP (fp16) & AMP (fp16) \\
GPUs & 4 × H100 & 4 × H100 \\
\bottomrule
\end{tabular}
\end{table}

Table~\ref{tab:hyperparams_lm_training} details the training hyperparameters, where both scales are optimized with AdamW using a cosine learning rate schedule, gradient clipping, and mixed-precision training on 4×H100 GPUs. While the small-scale models are trained for 100k steps on 6.55B tokens, large-scale models extend to 400k steps on 26.2B tokens, with warm-up and stronger gradient clipping applied to ensure stability. Together, these settings provide a consistent yet scalable foundation for benchmarking diverse MoE algorithms.

\paragraph{MoE Architecture Configurations.}
Table~\ref{tab:lm_model_setting} also specifies the MoE architecture configurations used in the language modeling pre-training setting, where $N$ denotes the total number of experts, $K$ the number of routed (active) experts per token, and $N_s$ the number of shared experts. For standard routing-based methods (SMoE, $\sigma$-MoE, and XMoE), we adopt a common configuration with $N=66$ experts and top-$K=8$ routing. For shared-expert variants (SharedE-V2/V3), we set $N_s=2$ shared experts and reduce routed capacity to $K=6$ to maintain comparable compute. For MoE++, following~\citet{jin2024moe++}, we augment the expert pool by adding zero-computation experts, resulting in $N=66+8$ while keeping $K=8$. For TC-MoE, following~\citet{yan2025tc}, the effective expert pool is expanded via ternary compositions, yielding $2N + K$ experts; under our configuration, this corresponds to $N=66\times 2 + 8$ with $K=8$. Across both model scales, we keep the expert dimension fixed (128 for 0.15B and 256 for 0.68B) to ensure a controlled comparison across MoE designs.

\section{Comparison Between Dense and SMoE Models}

\begin{figure}[ht]
    \centering
    \includegraphics[width=\linewidth]{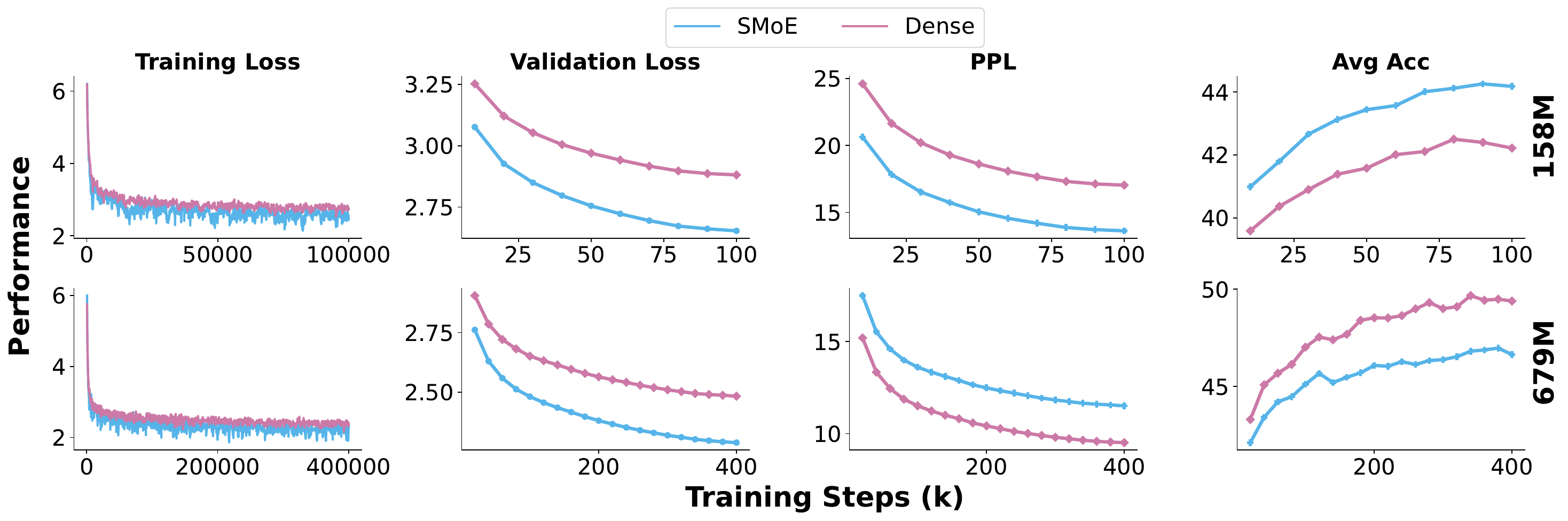}
    \caption{Training benchmark curves comparing Dense and SMoE models during language model pre-training (0.67B parameters).}
    \label{fig:dense_and_sparse}
\end{figure}

\input{contents/A_add_exps/tabs/dense_versue_smoe}

Figure~\ref{fig:dense_and_sparse} and Table \ref{tab:dense_vs_smoe_lm_result} presents a comparative analysis between dense baselines and SMoE models in the language model pre-training setting. Across all reported metrics, SMoE models consistently achieve stronger performance during the early stages of training, indicating faster optimization and more efficient utilization of model capacity compared to dense counterparts.

\section{Additional Analysis}    \label{appx:expert_coactivation}  

\subsection{Router Saturation}

\textbf{Router Saturation} - first introduced in OLMoE \citep{muennighoff2024olmoe} - quantifies the proportion of overlapping activated experts between an intermediate checkpoint at training step $t$ and the final checkpoint. This metric serves as an indicator of the router’s convergence dynamics throughout training. Higher router saturation values indicate greater alignment in expert selection, signifying that the router’s decisions are becoming increasingly consistent with its final checkpoint. Consequently, router saturation provides insight into the convergence of expert assignments of routing strategies during training. The formal definition and formula are defined in the Appendix~\ref{subappx:router_saturation}.

\begin{figure}[t!]
    \centering
    \includegraphics[width=\linewidth]{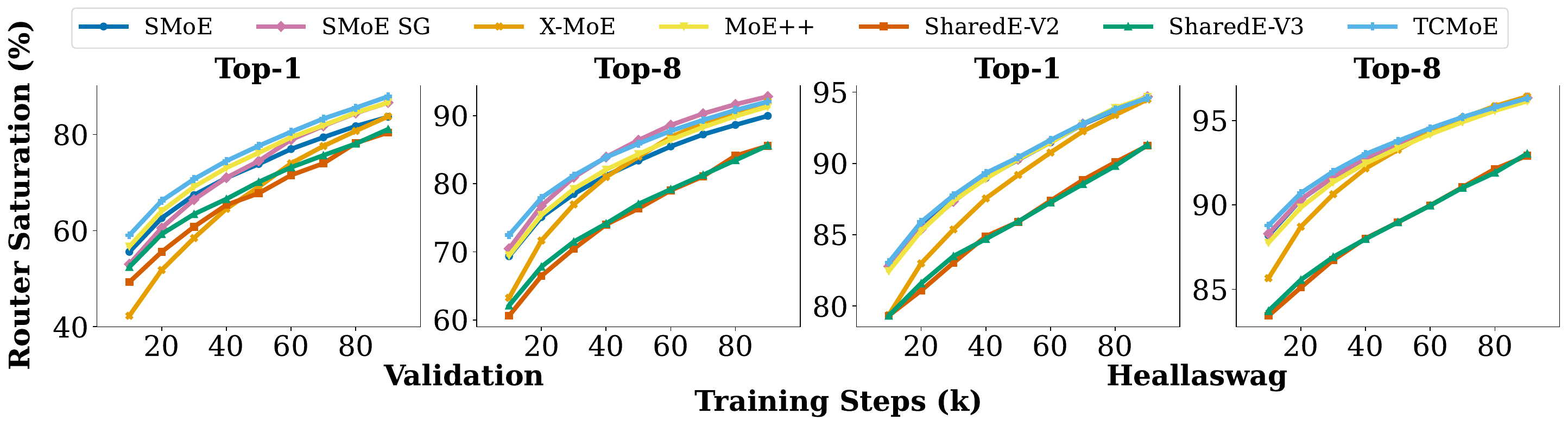}
    \caption{Router Saturation across methods during training on the language modeling task. We present both Top-1 and Top-8 routing results for small (0.15B) and large (0.68B) models, illustrating the progression of router convergence across model scales and expert selection strategies.}
    \label{fig:router_saturation}
\end{figure}

As shown in Figure~\ref{fig:router_saturation}, router saturation for all evaluated methods rises sharply during training, with most surpassing 60\%, and even reaching over 85\% on the HellaSwag benchmark, within the first 10\% of training under top-8 selection. In contrast, top-1 selection exhibits slightly slower convergence, indicating a more gradual stabilization of expert assignments. Notably, XMoE stands out as an outlier, converging more slowly than other variants—a trend consistent with its relatively lower performance reported in Table~\ref{tab:lm_result}. Overall, this early stabilization behavior aligns with prior work \citep{muennighoff2024olmoe, xue2024openmoe, nguyen2025deepseekmoe, kang2025flame}, suggesting a general tendency for MoE routers to converge rapidly toward stable expert assignments. Such early convergence is also consistent with the learning strategy advocated by Stable MoE \citep{dai2022stablemoe}, which aims to mitigate fluctuations in expert allocation during training.

\subsection{Cooperation or Competition}

\input{contents/A_add_exps/tabs/temperature}

To characterize expert behavior when contributing to the creation of outcomes, we analyze the performance change that occurs when we adjust the router's temperature. Under the parameterization below, a low router temperature makes the expert weights more peaky, indicating stronger competition between experts. In contrast, a high temperature makes expert weights more uniform, indicating more cooperative behavior. For sigmoid gating, the same trend holds: decreasing $\tau$ pushes gate activations closer to $\{0, 1\}$, whereas increasing $\tau$ moves them closer to $0.5$. To be more specific, if the logit of expert $k$ is $s_k$, our router is adjusted as follows:

\begin{align}
g(s_k) = 
\begin{cases}
\frac{\exp(s_k / \tau)}{\sum_{j=1}^{N_E} \exp(s_j / \tau)}, & \textit{if softmax router} \\
\sigma(s_k / \tau), & \textit{if sigmoid router}
\end{cases}
\end{align}

Table~\ref{tab:cooperation_or_competition} summarizes the effects of varying the router’s temperature parameter on model performance for both small (0.15B) and large (0.68B) model sizes across two representative language understanding tasks. We observe that deviating from the original temperature used during training generally results in performance degradation across both small and large model scales. In most settings, applying a low temperature ($\tau=0.1$), which induces sharper and more competitive routing, leads to the larger performance drop, although a few cases---such as XMoE and MoE++ on the small model---remain comparatively robust. Increasing the temperature to $\tau=10.0$, which encourages flatter and more cooperative routing, also often hurts performance, but is typically less damaging than over-sharpening the router. Interestingly, XMoE demonstrates greater robustness to temperature changes compared to other variants.

Overall, these findings suggest that most existing MoE architectures operate best near the training temperature and, when perturbed, are generally more resilient to enhanced cooperation (higher temperature) than to heightened competition (lower temperature). We leave a deeper investigation of the underlying mechanisms driving this cooperation-competition trade-off to future work.

\subsection{Expert Co-Activation Over Time}
\label{subappx:expert_coactivation_additional}

\begin{figure*}[ht!]
    \centering
    \includegraphics[width=.9\linewidth]{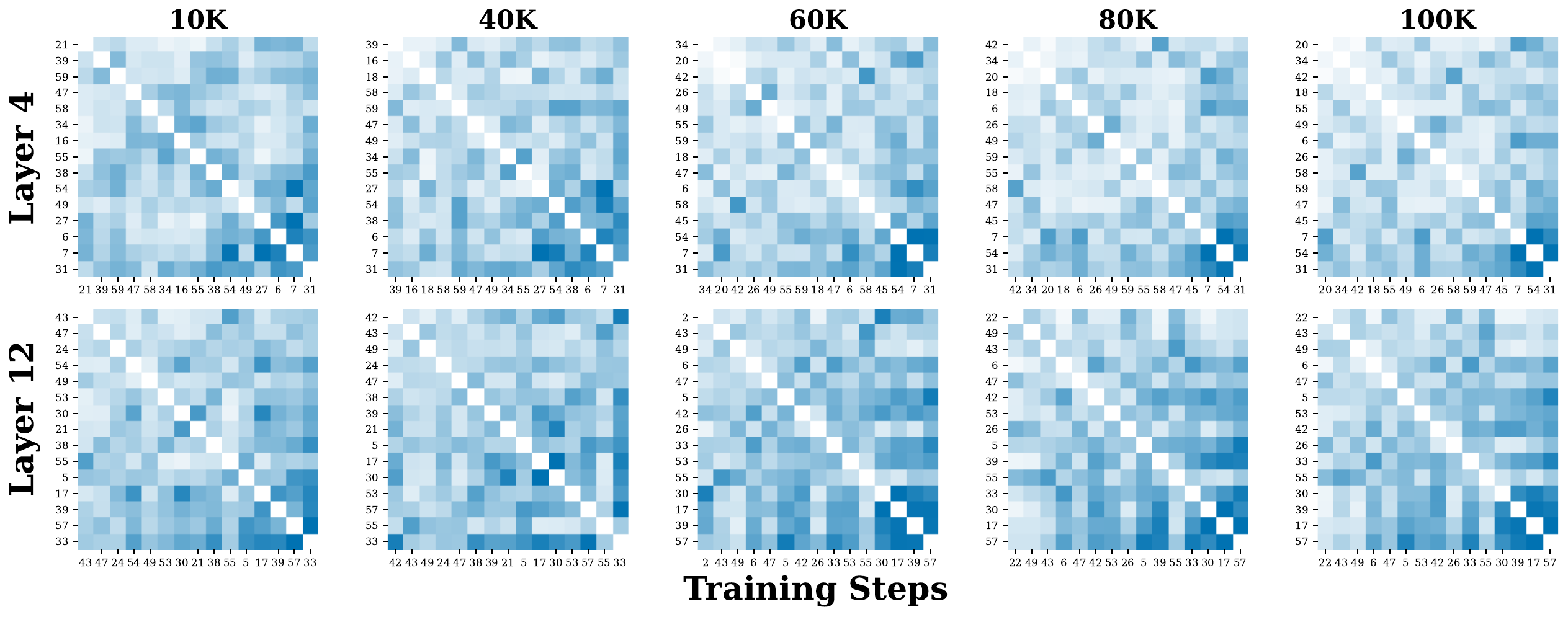}
    \caption{Expert co-activation (ECA) matrices for a Vanilla SMoE (0.15B) on the language modeling task, shown at Layers 4 and 12 across training checkpoints (10K–100K steps) on the validation set.}
    \label{fig:expert_coactivation_smoe}
\end{figure*}

To further examine expert interaction dynamics, we analyze the evolution of expert co-activation (ECA) patterns throughout training. Following \citet{muennighoff2024olmoe}, ECA measures how frequently two experts are activated together, normalized by the total number of activations of one expert (see Appendix~\ref{subappx:expert_coactivation} for the formal definition). Higher ECA values indicate persistent co-utilization between expert pairs.

Figure~\ref{fig:expert_coactivation_smoe} shows the ECA matrices for Layer~4 and Layer~12 of a small (0.15B) Vanilla SMoE model at multiple training checkpoints. Across training, the identity of the most strongly co-activated expert pairs remains largely unchanged, with only modest fluctuations in co-activation strength. This consistency suggests that expert collaboration structures emerge early in training and remain stable thereafter, indicating limited reorganization of expert interactions during later optimization.

\begin{figure*}[ht!]
    \centering
    \includegraphics[width=1\linewidth]{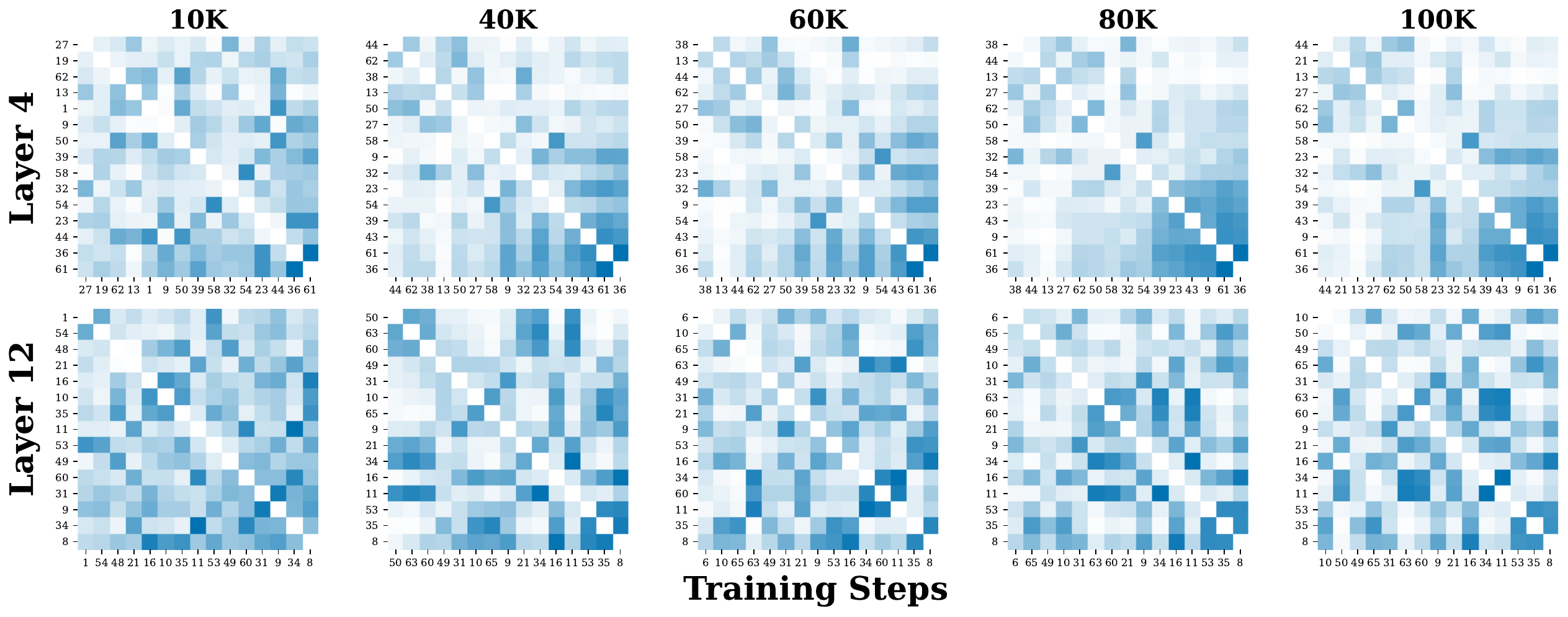}
    \caption{Expert Co-Activation across training XMoE model in a language modeling task.}
    \label{fig:expert_coactivation_xmoe}
\end{figure*}

\begin{figure*}[ht!]
    \centering
    \includegraphics[width=1\linewidth]{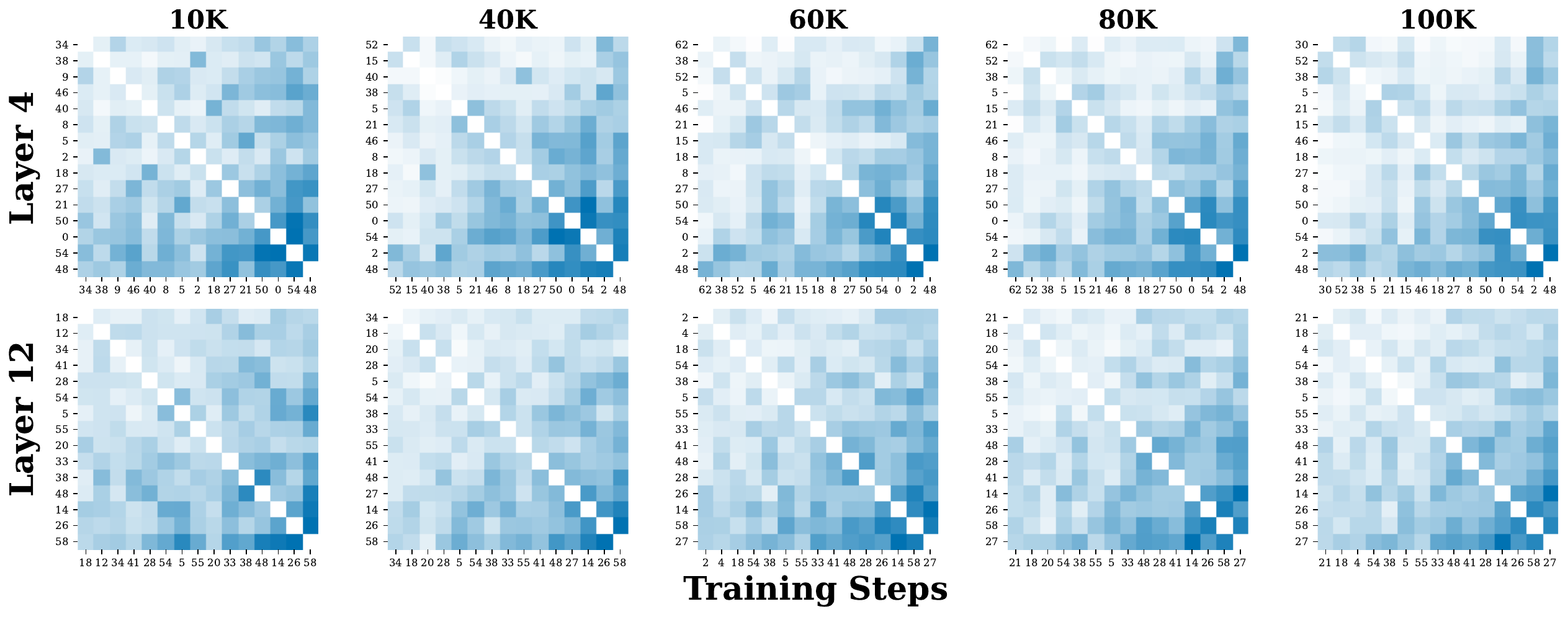}
    \caption{Expert Co-Activation across training $\sigma$-MoE model in a language modeling task.}
    \label{fig:expert_coactivation_smoe_sigmoid}
\end{figure*}

\begin{figure*}[ht!]
    \centering
    \includegraphics[width=1\linewidth]{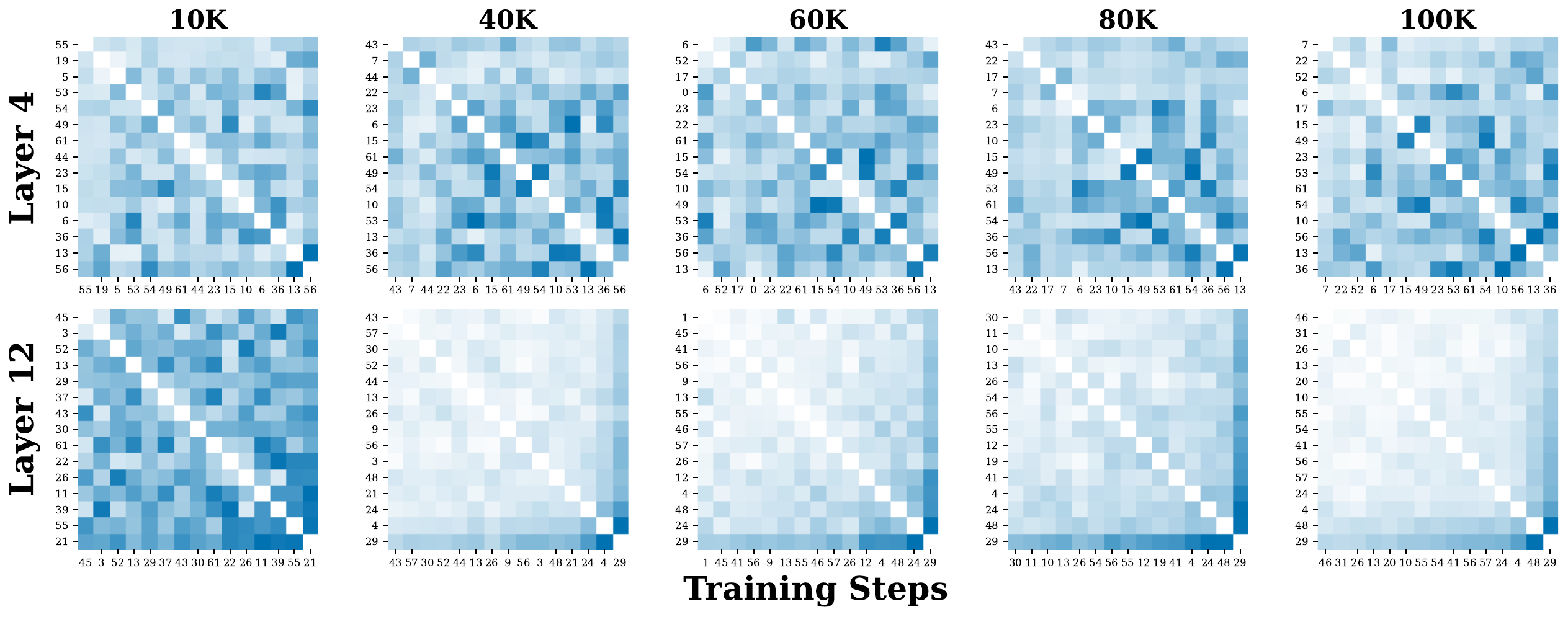}
    \caption{Expert Co-Activation across training SharedE-V2 model in a language modeling task.}
    \label{fig:expert_coactivation_deepseekv2}
\end{figure*}

\begin{figure*}[ht!]
    \centering
    \includegraphics[width=1\linewidth]{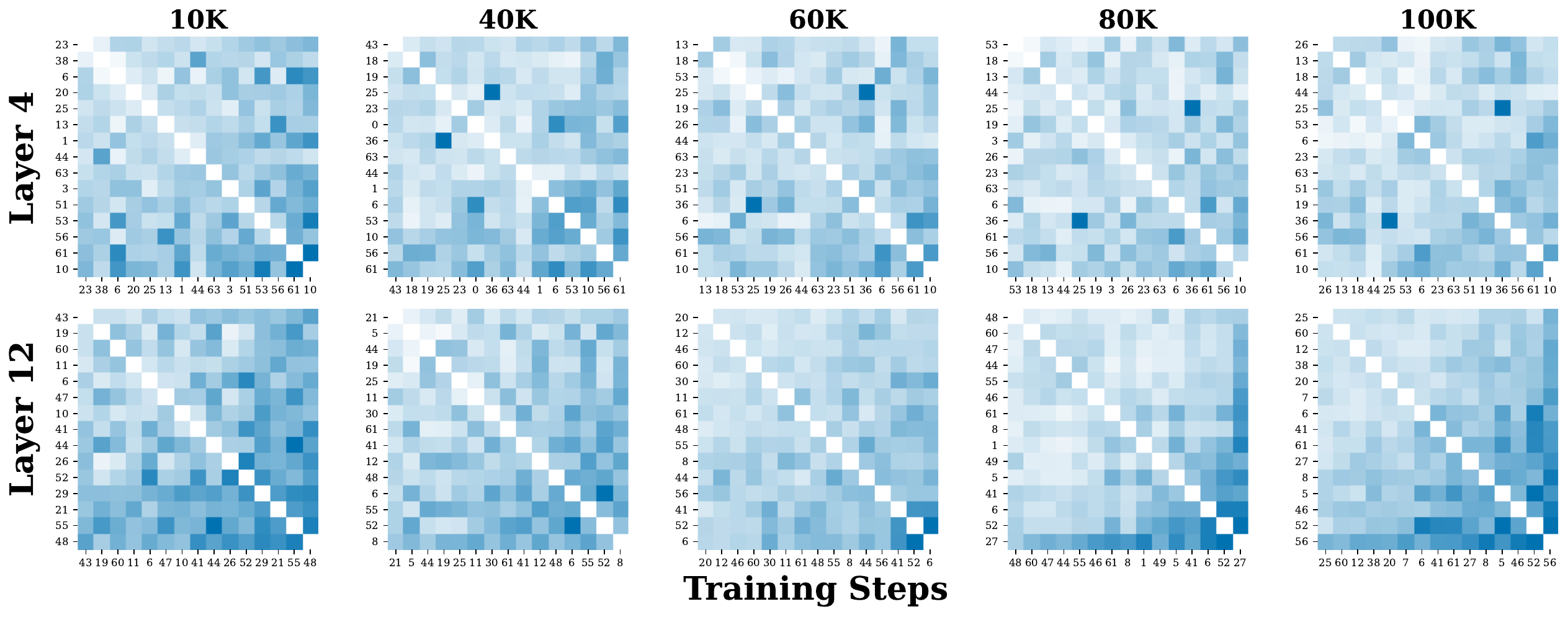}
    \caption{Expert Co-Activation across training SharedE-V3 model in a language modeling task.}
    \label{fig:expert_coactivation_deepseekv3}
\end{figure*}

\begin{figure*}[ht!]
    \centering
    \includegraphics[width=1\linewidth]{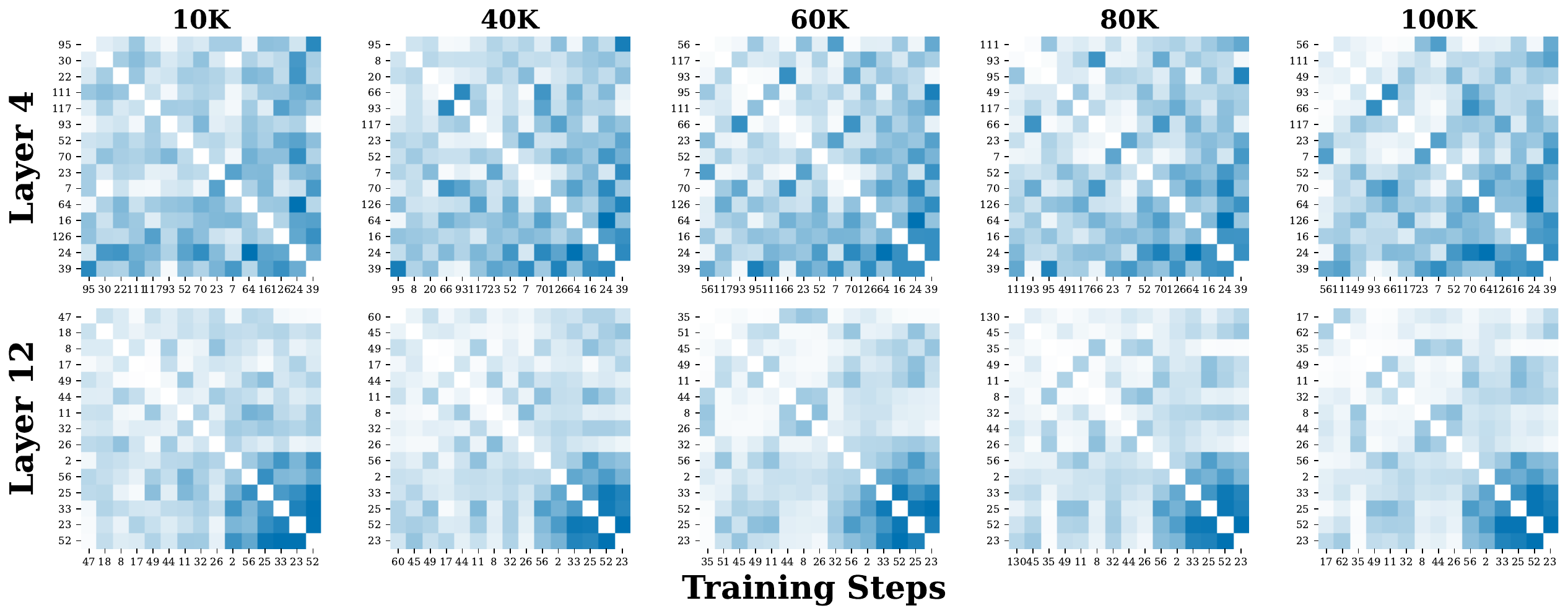}
    \caption{Expert Co-Activation across training TC-MoE model in a language modeling task.}
    \label{fig:expert_coactivation_tcmoe}
\end{figure*}

\begin{figure*}[ht!]
    \centering
    \includegraphics[width=1\linewidth]{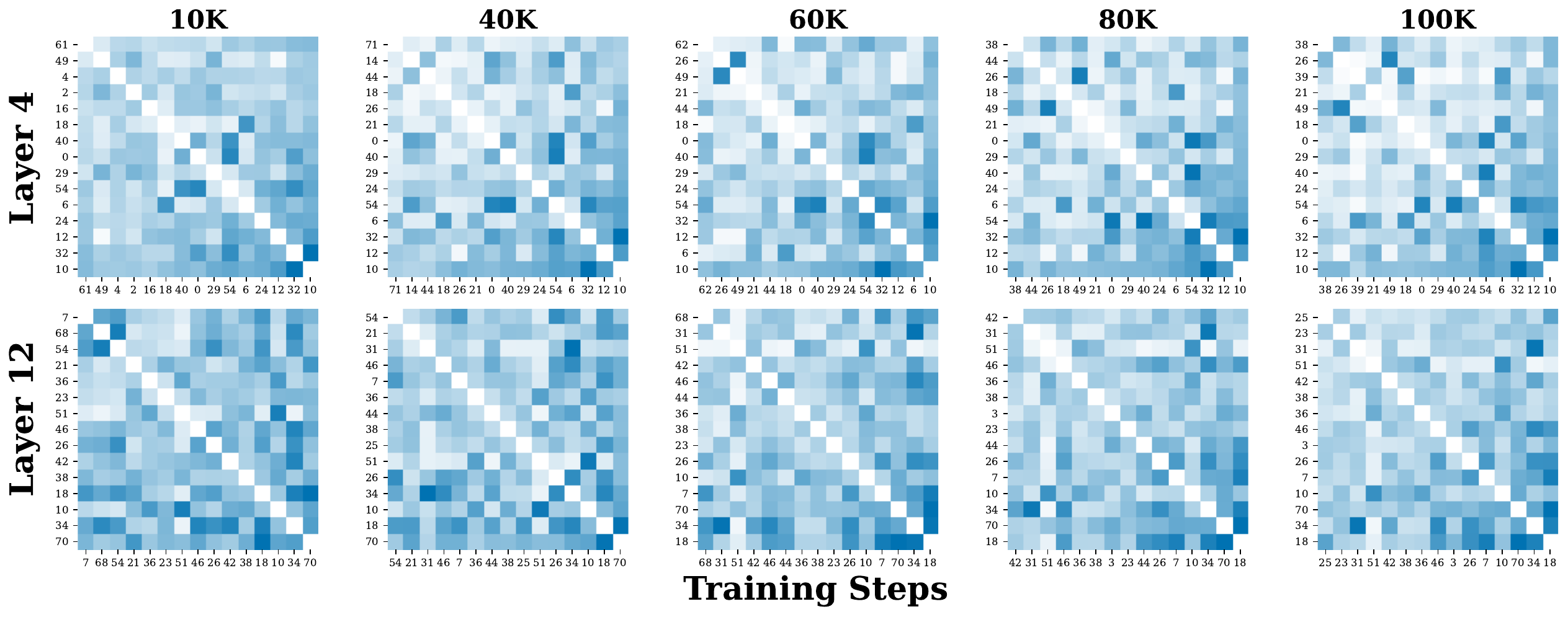}
    \caption{Expert Co-Activation across training MoE++ model in a language modeling task.}
    \label{fig:expert_coactivation_moe_plus_plus}
\end{figure*}

We observe similar stability in expert co-activation patterns across other SMoE variants, including XMoE, SMoE-Sigmoid, DeepSeek-V2, DeepSeek-V3, TCMoE, and MoE++. Corresponding ECA visualizations are provided in Figure~\ref{fig:expert_coactivation_xmoe}, Figure~\ref{fig:expert_coactivation_smoe_sigmoid}, Figure~\ref{fig:expert_coactivation_deepseekv2}, Figure~\ref{fig:expert_coactivation_moe_plus_plus}, Figure~\ref{fig:expert_coactivation_tcmoe}, and Figure~\ref{fig:expert_coactivation_deepseekv3}. Collectively, these results indicate that, despite differences in routing mechanisms and architectural refinements, expert co-activation relationships in SMoE-style models are remarkably stable over training.

\subsection{Load Balancing Loss and Router Z-loss Experiment} \label{appx:load_balancing}

\begin{figure}[ht]
    \centering
    \includegraphics[width=\linewidth]{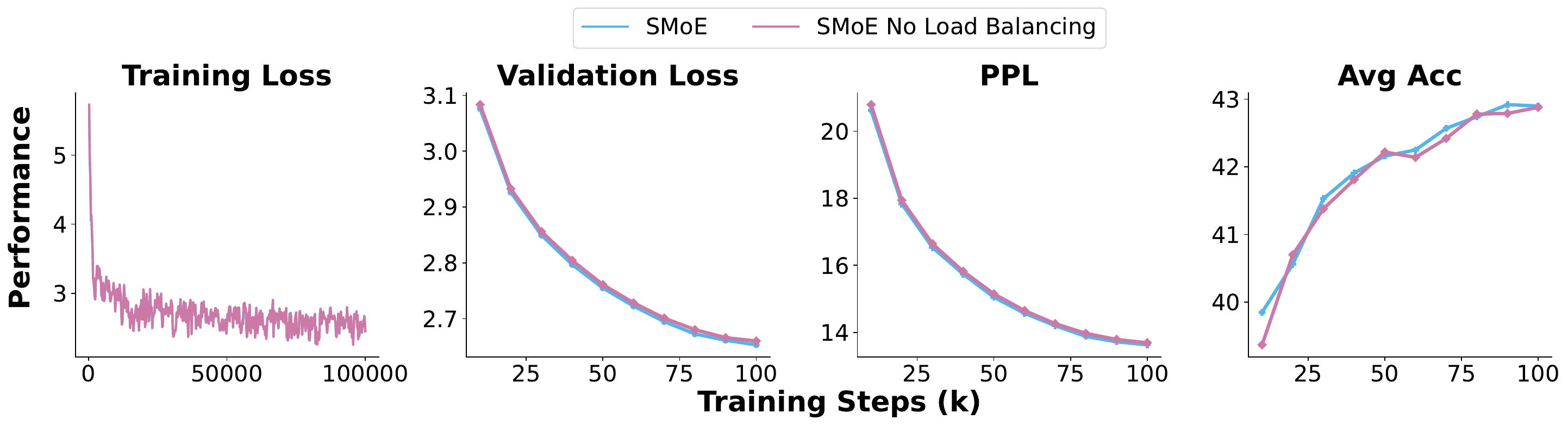}
    \caption{Benchmark curves during training in language modeling tasks for models with 0.15B parameters with and without load balancing loss.}
    \label{fig:load_balancing_curve}
\end{figure}

\begin{figure}[t!]
    \centering
    \includegraphics[width=\linewidth]{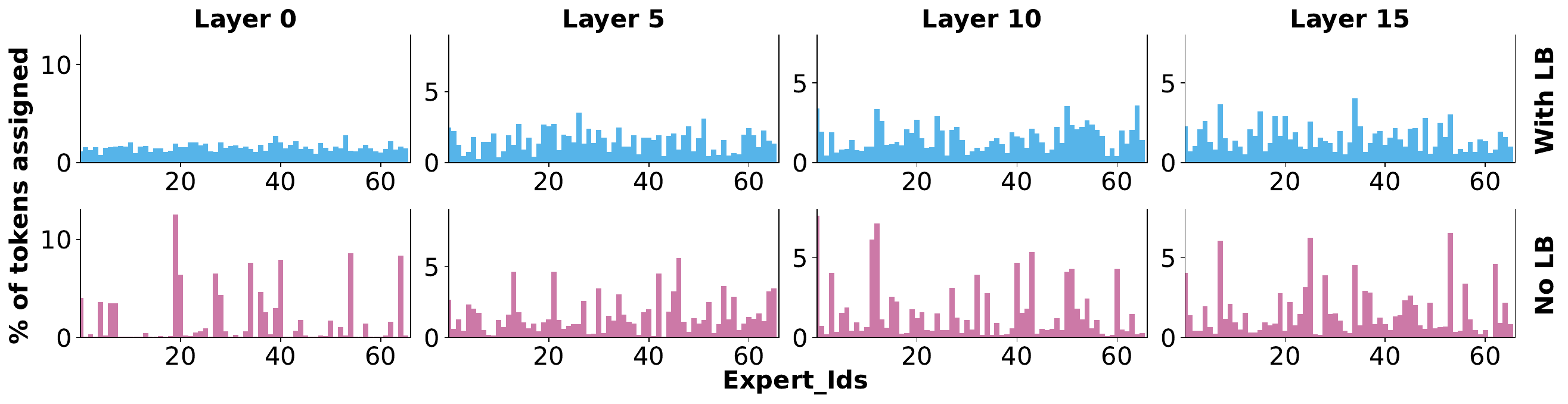}
    \caption{Expert selection ratio of SMoE model with and without load balancing loss in language modeling tasks (0.15B parameters model size).}
    \label{fig:load_balancing_dis}
\end{figure}

\renewcommand{\arraystretch}{1.02}
\begin{table}[ht!]
\centering
\small
\caption{\small Performance comparisons in the impact of load balancing loss and router z-loss of small model (0.15B parameters) in language modeling task.}
\resizebox{\textwidth}{!}{
\begin{tabular}{c cccccccccc >{\columncolor{customyellow!80}}c}
\toprule
\makecell{\textbf{MoE Method}} & \makecell{\textbf{PPL} $\downarrow$} & \makecell{\textbf{LAM}\\ \textbf{BADA}} & \makecell{\textbf{BLiMP}} & \makecell{\textbf{CBT}} & \makecell{\textbf{Hella}\\ \textbf{Swag}} & \makecell{\textbf{PIQA}} & \makecell{\textbf{ARC-}\\ \textbf{Challenge}} & \makecell{\textbf{RACE}} & \makecell{\textbf{SIQA}} & \makecell{\textbf{Common}\\ \textbf{SenseQA}} & \makecell{\textbf{Average}} \\
\midrule
\textbf{SMoE (0.01 lb)} & 13.63 & 25.27\% & 77.71\% & 84.18\% & 29.43\% & 57.94\% & 21.20\% & 30.11\% & 35.62\% & 24.65\% & \textbf{42.90\%} \\
\textbf{SMoE (no-lb)} & 13.69 & 24.90\% & 76.81\% & 84.13\% & 29.38\% & 57.51\% & 21.63\% & 30.26\% & 35.67\% & 25.63\% & 42.88\% \\
\textbf{SMoE (0.01 lb + 0.001 z-loss)} & 13.62 & 25.49\% & 77.55\% & 84.23\% & 29.11\% & 58.71\% & 21.89\% & 29.63\% & 35.21\% & 23.91\% & 42.86\%   \\
 \bottomrule
\end{tabular}}

\label{tab:aux_loss_result}
\end{table}

Figure~\ref{fig:load_balancing_curve} and Figure~\ref{fig:load_balancing_dis} illustrate the effect of the load balancing loss on both training dynamics and expert assignment in Mixture-of-Experts models. The results demonstrate that incorporating load balancing loss contributes to more stable training and improved overall performance. Specifically, Figure~\ref{fig:load_balancing_dis} highlights how the load balancing loss promotes a more uniform distribution of tokens across experts, mitigating the risk of expert under-utilization.

Beyond load balancing, \citep{zoph2022st} introduced the router z-loss as an additional regularization strategy to further stabilize MoE training. To assess the practical impact of these auxiliary losses, we report in Table~\ref{tab:aux_loss_result} a comparison of model performance under different loss configurations for language-model pretraining. Our results indicate that applying the router z-loss does not yield a performance improvement in this setting; in fact, it slightly degrades accuracy relative to using only the load balancing loss. Therefore, we opt not to include the router z-loss in our final language-model experimental setup.

\section{Definitions and Formulations of Analysis Metrics} \label{appx:analysis}

\subsection{Total Score Change under Routing Perturbations: DropTop1 vs.\ DropTop1\&2}
\label{subappx:drop1and2}
Let $\mathcal{D}=\{D_i\}_{i=1}^{|\mathcal{D}|}$ denote the set of evaluation benchmarks.
For a given MoE model, let $M_i \in \mathbb{R}$ be its original evaluation score (e.g., accuracy; higher is better) on benchmark $D_i$.
After applying a routing perturbation $\pi \in \{\textit{DropTop1},\,\textit{DropTop1$\&$2}\}$ at inference time, we obtain the perturbed score $\hat{M}_i^{\,\pi} \in \mathbb{R}$.

We define the aggregate performance change (reported as $\Delta$ Total Score in Figure \ref{fig:droptop1}) as:
\begin{equation}
\Delta \text{TotalScore}^{\,\pi}
= \sum_{i=1}^{|\mathcal{D}|}\left(\hat{M}_i^{\,\pi} - M_i\right).
\end{equation}
Negative values indicate that the routing perturbation leads to a net performance degradation relative to the original routing configuration, whereas positive values indicate that the perturbed routing achieves higher aggregate performance than the original model. In the latter case, the improvement suggests that the original routing decision may have been suboptimal, and that alternative expert assignments can better exploit the model’s representational capacity.

\subsection{Expert Entropy Allocation} \label{subappx:eae}

We investigate the behaviors of the expert selection mechanism by exploring how often each expert is selected in the MME benchmark. 
To this end, we analyze the frequency of the expert selection across different subtasks to gain insights into the specialization behavior of each expert. Given an MoE algorithm with $N$ experts and $L$ layers, the selection frequency of each expert $i$ at a given layer $l$ is denoted as $\text{freq}_i^{(l)}$ ($i = 1, 2, \ldots, N$ and $l = 1, 2, \ldots, L$). Note that this selection frequency is counted across all samples in the benchmark, in this case we choose to be MME.
Then, the entropy $H^{(l)}$ at each layer $l$ is calculated by integrating the probability of selecting expert $i$ into Shannon's entropy formula as follows:

\begin{align}
EAE^{(l)} = \frac{- \sum_{i=1}^{N} \left( \frac{\text{freq}_i^{(l)}}{\sum_{j=1}^{N} \text{freq}_j^{(l)}} \right) \log_2 \left( \frac{\text{freq}_i^{(l)}}{\sum_{j=1}^{N} \text{freq}_j^{(l)}} \right)}{\log_2(N)},
\end{align}

where:
\begin{itemize}
    \item $\text{freq}_i^{(l)}$: The number of times expert $i$ is selected at layer $l$.
    \item $N$: The total number of experts in the MoE algorithms.
    \item $\sum_{j=1}^{N} \text{freq}_j^{(l)}$: The total number of expert selections at layer $l$.
    \item $EAE^{(l)}$: The entropy value at layer $l$, measuring the uncertainty or diversity in expert selections.
\end{itemize}

With $EAE$, we can measure the frequency of expert selections across all layers $H_{EAE}$. Importantly, high $EAE$ Indicates a balanced expert utilization across all layers, where the model tends to distribute selections evenly among all experts. In contrast, low $EAE$ suggests a concentrated usage of a few experts in most layers, indicating specialization or a preference for certain experts.

\subsection{Router Margin}  \label{subappx:router_margin}

Router Margin is a metric that quantifies the dominance of the highest-scoring expert in a Mixture-of-Experts (MoE) routing decision. It is defined as the difference between the top-1 and top-2 gating scores:

\begin{align*}
    \text{Router Margin}(l) = \frac{1}{N} \sum_{i=1}^{N} \bigl(top1(x_i) - top2(x_i)\bigr),
\end{align*}

Where:
\begin{itemize}
    \item $\text{RouterMargin}(l)$: Router margin at layer $l$
    \item $N$: The total number of tokens in the dataset.
    \item $x_i$: $i$-th input token
    \item $top1(x_i)$: top-1 routing score of input $x_i$
    \item $top2(x_i)$: top-2 routing score of input $x_i$
\end{itemize}

Router Margin provides a quantitative measure of how decisively the router selects the top expert relative to alternatives. A larger margin indicates that the router strongly favors the top-1 expert, reflecting more confident and specialized routing, whereas a smaller margin implies greater ambiguity and potential overlap among experts. This metric is thus a valuable diagnostic tool for understanding the dynamics of expert dominance and the evolution of routing confidence during training.

\subsection{Router Saturation}  \label{subappx:router_saturation}

In formal terms, router saturation is the proportion of expert activations at some intermediary checkpoint at time $t$ that matches the expert IDs activated at some final checkpoint $T$ over the same dataset:

\begin{align}
    \text{RouterSaturation}(t) = \frac{1}{N} \sum_{i=1}^{N} \frac{\left| \mathcal{E}_i^{(t)} \cap \mathcal{E}_i^{(T)} \right|}{k},
\end{align}

Where:
\begin{itemize}
    \item $N$: The total number of tokens in the dataset.
    \item $k$: The number of top-k experts activated per input token.
    \item $\mathcal{E}_i^{(t)}$: The set of $k$ experts activated for the $i$-th token at the $t$-th checkpoint.
    \item $\mathcal{E}_i^{(T)}$: The set of $k$ experts activated for the $i$-th token at the final checkpoint $T$.
    \item $\left|\mathcal{E}_i^{(t)} \cap \mathcal{E}_i^{(T)}\right|$: The number of common experts activated for the $i$-th token between the $t$-th and final checkpoints $T$.
\end{itemize}

Router saturation provides a quantitative measure of how early the routing decisions converge during training. A saturation value of 100\% indicates that the router at an intermediate checkpoint routes to the same set of experts as at the final checkpoint. High saturation values at early checkpoints reflect early convergence in expert selection, indicating that the router has rapidly settled into a stable assignment pattern. In contrast, low saturation values suggest ongoing exploration or adaptation in expert allocations, signaling that the routing mechanism is still undergoing significant adjustments.

\subsection{Experts Co-activation}  \label{subappx:expert_coactivation}

We define expert co-activation as the proportion of times two specific experts, $E_i$ and $E_j$, are simultaneously activated out of the total number of activations of one of those experts:
\begin{equation}
\text{Expert co-activation}(E_i, E_j) = \frac{N_{E_i, E_j}}{N_{E_i}},
\label{eq:expert_coactivation}
\end{equation}
where:
\begin{itemize}
    \item $E_i$: The first expert.
    \item $E_j$: The second expert.
    \item $N_{E_i, E_j}$: The number of times experts $E_i$ and $E_j$ are activated together.
    \item $N_{E_i}$: The total number of times expert $E_i$ is activated.
\end{itemize}

A co-activation of 100\% indicates that if $E_i$ is activated, $E_j$ is also always activated. A value of 0\% indicates that the experts never co-occur. If multiple expert pairs have high co-activation, it may suggest that these experts could be merged, benefiting less from keeping them separate. In a distributed setup, we could place highly co-activated experts on the same device to reduce communication costs during model inference.

\section{Training Time and Resource Allocation}
\label{appx:training_time_resource_allocation}
\subsection{Training Time and Resource Usage}
Table~\ref{tab:time_and_resource} reports the training time and GPU resource allocation across all experimental configurations. In the VLM setting, \textsc{SharedE-V2} and \textsc{SharedE-V3} are consistently the most training-efficient methods, completing OneVision training roughly two hours faster than the competing approaches. This advantage stems from their shared-expert design, which reduces per-token routing computation. The router therefore scores only the non-shared experts, i.e., $N$ minus the number of shared experts, lowering routing overhead and improving end-to-end efficiency. Importantly, this runtime reduction does not come at the cost of model quality. As shown in Table~\ref{tab:vlm_moe_results}, both variants remain highly competitive and, in several cases, achieve the best overall performance on OneVision. For language-model pretraining, however, we use the CVMM Triton kernel~\cite{csordas2024moeut}, which substantially accelerates sparse MoE computation. As a result, runtime differences in this setting are influenced more by kernel-level optimization than by architectural choices such as \textsc{SharedE-V2} or \textsc{SharedE-V3}. We therefore view training-time comparisons as most informative in the VLM setting, where the observed differences more directly reflect methodological design choices.

\begin{table}[htbp]
  \centering
  \caption{Training Time and GPU Resource Allocation across all Experimental Settings.}
  \label{tab:time_and_resource}
  \setlength{\tabcolsep}{4pt}
  \renewcommand{\arraystretch}{1.2}

  \resizebox{.8\textwidth}{!}{%
    \begin{tabular}{*{5}{c}}
      \toprule
        \multicolumn{3}{c}{\textbf{Model}} & \makecell{\textbf{Training Time} \\ (hours)} & \makecell{\textbf{Resource}\\ \textbf{Allocation}} \\
      \midrule
        \multirow{6}{*}{\makecell{VLM}} & \multicolumn{2}{c}{Pre-Training} & 2h35m & 4xH100 \\
      \cmidrule{2-5}
         & \multicolumn{2}{c}{Pre-FineTuning} & 16h & 4xH100 \\
      \cmidrule{2-5}
         & \multirow{7}{*}{\makecell{Visual Instruction\\ Tuning \\ LLAVA-665K}} & SMoE & 17h19m & 4xH100 \\
         &  & XMoE & 17h59m & 4xH100 \\
         &  & $\sigma$-MoE & 17h32m & 4xH100 \\
         &  & SharedE-V2 & 16h29m & 4xH100 \\
         &  & SharedE-V3 & 16h47m &  4xH100 \\
         &  & TC-MoE & 18h01m & 4xH100 \\
         &  & MoE++ & 18h50m & 4xH100 \\
      \cmidrule{2-5}
         & \multirow{7}{*}{\makecell{Visual Instruction\\ Tuning \\ OneVision / 1M2 samples}} & SMoE & 33h01m & 4xH100 \\
         &  & XMoE & 34h27m & 4xH100 \\
         &  & $\sigma$-MoE & 33h33m & 4xH100 \\
         &  & SharedE-V2 & 31h19m & 4xH100 \\
         &  & SharedE-V3 & 31h39m & 4xH100 \\
         &  & TC-MoE & 33h47m & 4xH100 \\
         &  & MoE++ & 35h40m & 4xH100 \\

      \midrule
        \multirow{16}{*}{\makecell{Language \\ Modeling}} & \multirow{8}{*}{0.15B parametes} & Dense & 5h47m & 4xH100 \\
         &  & SMoE & 6h15m & 4xH100 \\
         &  & XMoE & 6h21m & 4xH100 \\
         &  & $\sigma$-MoE & 6h12m & 4xH100 \\
         &  & SharedE-V2 & 6h25m & 4xH100 \\
         &  & SharedE-V3 & 6h23m & 4xH100 \\
         &  & TC-MoE & 6h10m & 4xH100 \\
         &  & MoE++ & 6h17m & 4xH100 \\
      \cmidrule{2-5}
         & \multirow{8}{*}{0.68B parametes} & Dense & 41h30m & 4xH100 \\
         &  & SMoE & 42h11m & 4xH100 \\
         &  & XMoE & 42h54m & 4xH100 \\
         &  & $\sigma$-MoE & 41h57m & 4xH100 \\
         &  & SharedE-V2 & 43h01m & 4xH100 \\
         &  & SharedE-V3 & 43h09m & 4xH100 \\
         &  & TC-MoE & 41h38m & 4xH100 \\
         &  & MoE++ & 42h22m & 4xH100 \\
      \bottomrule
      \end{tabular}%
   }
\end{table}

\subsection{Peak GPU Memory Usage and Per-Sample Inference Latency}
Table~\ref{tab:peak_training_memory} reports the peak GPU memory usage observed during training for each SMoE method, together with the average per-sample inference latency, in the 5.67B VLM setting. These measurements provide a more complete practical comparison by showing how each routing design affects both memory footprint and inference efficiency under matched settings.

\begin{table}[htbp]
  \centering
  \caption{Peak GPU memory usage during training and per-sample inference latency across SMoE methods in the 5.67B VLM setting.}
  \label{tab:peak_training_memory}
  \renewcommand{\arraystretch}{1.15}
  \begin{tabular}{lcc}
    \toprule
    \textbf{Method} & \textbf{Peak GPU Memory Usage} & \makecell{\textbf{Inference Latency} \\ \textbf{(s/sample)}} \\
    \midrule
    SMoE & 44.73 GB & 0.187 \\
    XMoE & 44.76 GB & 0.204 \\
    $\sigma$-MoE & 44.79 GB & 0.192 \\
    SharedE-V2 & 43.52 GB & 0.182 \\
    SharedE-V3 & 43.51 GB & 0.186 \\
    TC-MoE & 44.76 GB & 0.201 \\
    MoE++ & 42.76 GB & 0.183 \\
    \bottomrule
  \end{tabular}
\end{table}

Two practical patterns are worth highlighting. First, the shared-expert variants occupy a particularly favorable efficiency region. Relative to the standard SMoE baseline, \textsc{SharedE-V2} and \textsc{SharedE-V3} reduce peak training memory by about 1.2~GB while also delivering equal or lower inference latency, and \textsc{SharedE-V2} achieves the lowest latency overall (0.182~s/sample). This is consistent with the architectural intuition discussed above: because some capacity is moved to always-on shared experts, the router only scores the remaining routed experts, which lowers routing overhead in the VLM setting.

Second, Table~\ref{tab:peak_training_memory} shows that methods with broadly similar benchmark quality can nevertheless differ meaningfully in system cost. In this setting, inference latency ranges from 0.182 to 0.204~s/sample, even though the main benchmark tables show only modest quality differences across methods. The contrast is especially clear for \textsc{XMoE} and \textsc{TC-MoE}, which are among the slowest methods at inference while remaining close to the rest of the benchmark in peak memory usage. By contrast, \textsc{MoE++} achieves the lowest peak memory usage, indicating that different architectural modifications can shift different parts of the quality--efficiency trade-off. Taken together, these results reinforce our main practical claim: SMoE methods should be compared not only by end-task quality, but by the full quality--efficiency frontier they induce.

\section{Training Benchmark Curves} \label{appx:training_curve}

\begin{figure}[ht!]
    \centering
    \includegraphics[width=\linewidth]{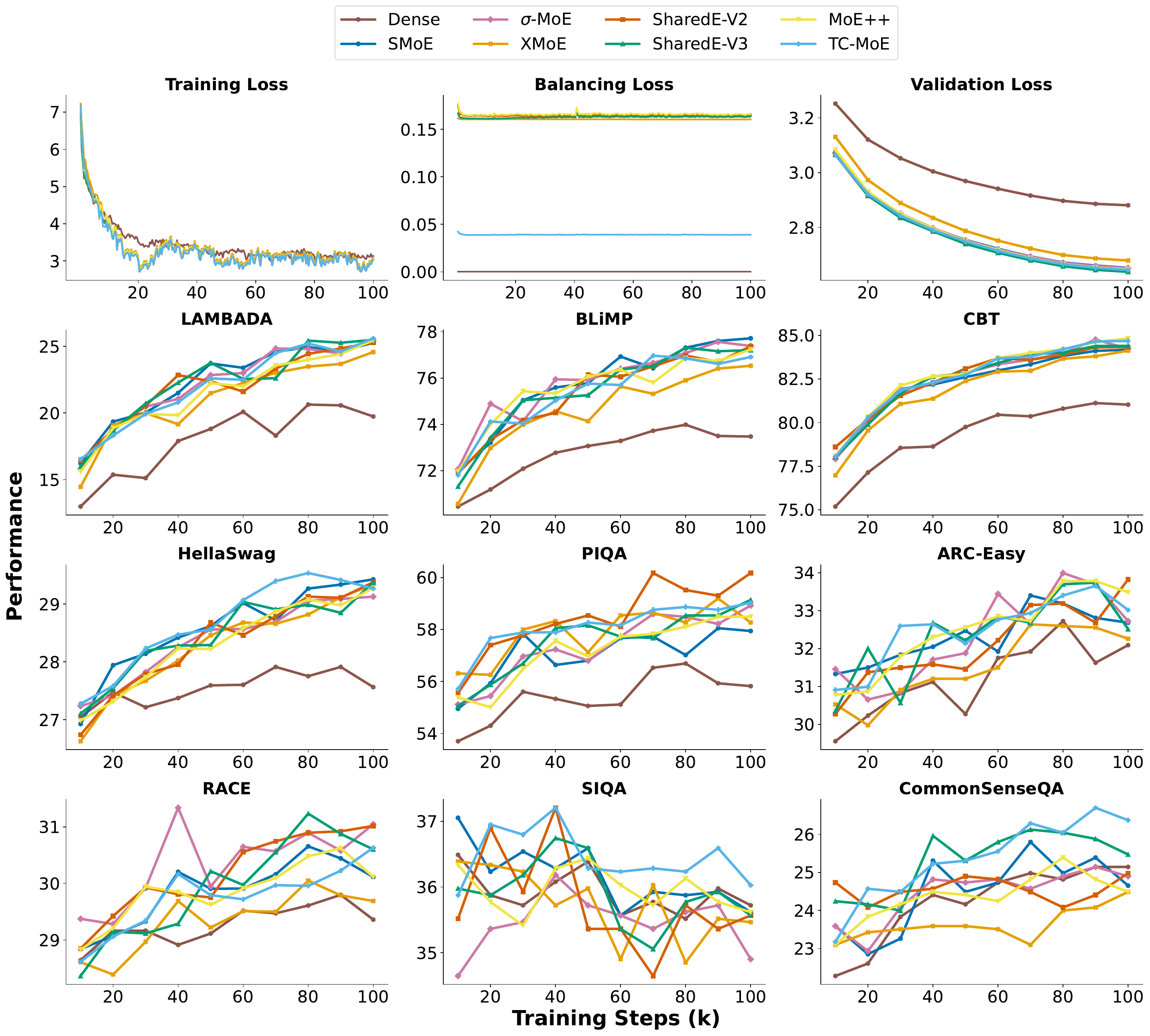}
    \caption{Benchmark curves during training in language modeling tasks for models with 0.15B parameters.}
    \label{fig:lm_benchmark_curve_0.15B}
\end{figure}

\begin{figure}[ht!]
    \centering
    \includegraphics[width=\linewidth]{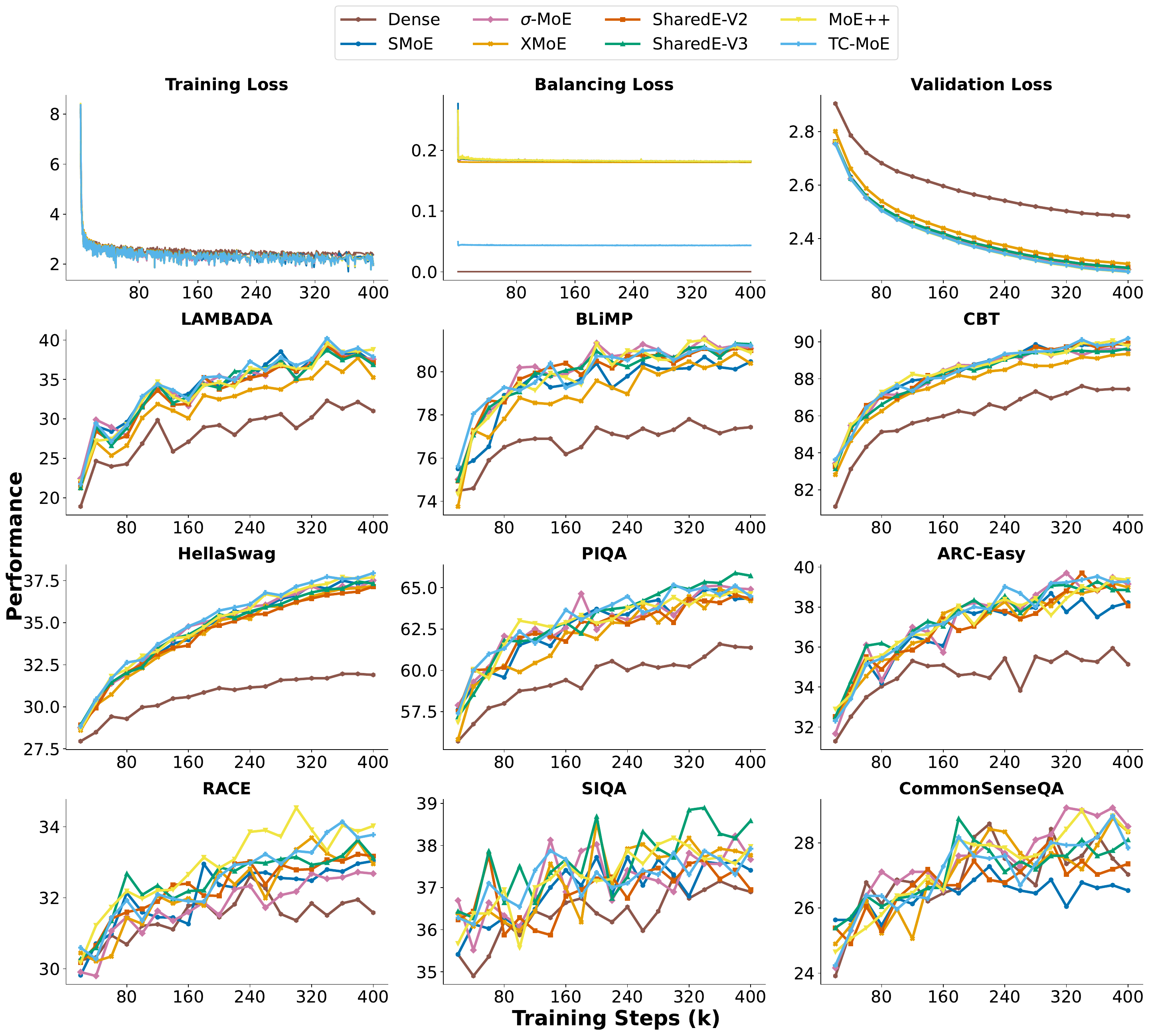}
    \caption{Benchmark curves during training in language modeling tasks for models with 0.68B parameters.}
    \label{fig:lm_benchmark_curve_0.68B}
\end{figure}

\newpage
\newpage

%% file: contents/A_add_exps/tabs/hybrid_dense_sparse_vlm.tex
\begin{table}[ht]
\centering
\caption{Comparison for the \textsc{SharedE-V3} hybrid sparse-upcycling setting on LLaVA-665K. Hybrid \textsc{SharedE-V3} denotes the dense precursor, and \textsc{SharedE-V3} denotes the sparsely upcycled counterpart. Bold values mark the better score in each column.}
\scriptsize
\renewcommand{\arraystretch}{1.02}
\begin{adjustbox}{width=\textwidth,center}
\begin{tabular}{lccccccccccccc}
\toprule
\textbf{\begin{tabular}[c]{@{}c@{}}Model \\ Variant\end{tabular}} &
\textbf{AI2D} &
\textbf{\begin{tabular}[c]{@{}c@{}}Text \\ VQA\end{tabular}} &
\textbf{GQA} &
\textbf{\begin{tabular}[c]{@{}c@{}}MM \\ Bench\end{tabular}} &
\textbf{\begin{tabular}[c]{@{}c@{}}Hallusion \\ Bench\end{tabular}} &
\textbf{\begin{tabular}[c]{@{}c@{}}Math \\ Vista\end{tabular}} &
\textbf{MMMU} &
\textbf{MMStar} &
\textbf{Pope} &
\textbf{MME} &
\textbf{\begin{tabular}[c]{@{}c@{}}MME \\ RW\end{tabular}} &
\textbf{\begin{tabular}[c]{@{}c@{}}OCR \\ Bench\end{tabular}} &
\textbf{\begin{tabular}[c]{@{}c@{}}AVG \\ Acc\end{tabular}}$\uparrow$ \\
\midrule
Hybrid \textsc{SharedE-V3} & \textbf{66.22} & 41.44 & \textbf{61.73} & 71.22 & \textbf{42.17} & \textbf{30.80} & 40.78 & 40.91 & 86.78 & 59.08 & \textbf{32.00} & 32.10 & 50.43 \\
\textsc{SharedE-V3} & 65.58 & \textbf{42.06} & 61.26 & \textbf{72.42} & 41.43 & 30.60 & \textbf{42.44} & \textbf{41.75} & \textbf{86.81} & \textbf{60.93} & 31.47 & \textbf{32.60} & \textbf{50.86} \\
\bottomrule
\end{tabular}
\end{adjustbox}
\label{tab:hybrid_dense_sparse_vlm}
\end{table}

%% file: contents/A_add_exps/tabs/dense_versue_smoe.tex
\begin{table}[H]
\centering
\small
\renewcommand{\arraystretch}{1.05}

\caption{Performance comparison between dense and SMoE models for language model pre-training, evaluated on a small-scale model (0.15B parameters) and a large-scale model (0.68B parameters). PPL denotes perplexity; lower values indicate better performance.}

\resizebox{\textwidth}{!}{
\begin{tabular}{c l c ccccccccc c}
\toprule
 & \textbf{MoE Method}
 & \textbf{PPL} $\downarrow$
 & \makecell{\textbf{LAMBADA}}
 & \textbf{BLiMP}
 & \textbf{CBT}
 & \makecell{\textbf{Hella}\\\textbf{Swag}}
 & \textbf{PIQA}
 & \makecell{\textbf{ARC-}\\\textbf{Easy}}
 & \textbf{RACE}
 & \textbf{SIQA}
 & \makecell{\textbf{Common}\\\textbf{SenseQA}}
 & \textbf{AVG} $\uparrow$ \\
\midrule
\multirow{2}{*}{\makecell{Small Model\\(0.15B)}}
 & Dense (36M)
 & 17.04
 & 19.74
 & 73.48
 & 81.03
 & 27.56
 & 55.82
 & 32.09
 & 29.36
 & 35.72
 & 25.14
 & 42.22 \\
 & SMoE
 & 13.63
 & 25.27
 & 77.71
 & 84.18
 & 29.43
 & 57.94
 & 32.68
 & 30.11
 & 35.62
 & 24.65
 & 44.18 \\
\midrule
\multirow{2}{*}{\makecell{Large Model\\(0.68B)}}
 & Dense (131M)
 & 11.51
 & 31.00
 & 77.43
 & 87.45
 & 31.90
 & 61.37
 & 35.14
 & 31.58
 & 36.90
 & 27.03
 & 46.64 \\
 & SMoE
 & 9.51
 & 37.13
 & 80.47
 & 89.83
 & 37.49
 & 64.36
 & 38.22
 & 33.03
 & 37.41
 & 26.54
 & 49.39 \\
\bottomrule
\end{tabular}
}

\label{tab:dense_vs_smoe_lm_result}
\end{table}

%% file: contents/A_add_exps/tabs/temperature.tex
\renewcommand{\arraystretch}{1.02}
\begin{table}[t!]
\centering
\small

\caption{\small Performance variation under different router temperatures across MoE algorithms. We evaluate the impact of router temperature ($\tau$) on expert cooperation and competition in both small (0.15B) and large (0.68B) models, using two language understanding tasks (BLiMP and HellaSwag). Under this parameterization, lower $\tau$ sharpens routing and increases competition, whereas higher $\tau$ smooths routing and promotes cooperation.}

\resizebox{\textwidth}{!}{
\begin{tabular}{ccc lllllll c}
\toprule
 & \textbf{$\tau$} & \textbf{Task} & \textbf{SMoE} & \textbf{SMoE SG} & \textbf{XMoE} & \textbf{SharedE-V2} & \textbf{SharedE-V3} & \textbf{TC-MoE} & \textbf{MoE++} & \textbf{AVG $\Delta$} \\
 \midrule
\multirow{6}{*}{\makecell{Small Model\\(0.15B)}} & \multirow{2}{*}{1.0} & BLIMP & 77.71\% & 76.75\% & 76.53\% & 77.37\% & 77.20\% & 76.91\% & 77.23\% \\
 &  & HellaSwag & 29.43\% & 29.15\% & 29.34\% & 29.38\% & 29.38\% & 29.27\% & 29.28\% \\
 \cmidrule{2-11}
 & \multirow{2}{*}{10.0} & BLIMP & 69.64\%\textcolor{red}{\scriptsize$\downarrow$8.07} & 68.52\%\textcolor{red}{\scriptsize$\downarrow$8.24} & 76.23\%\textcolor{red}{\scriptsize$\downarrow$0.30} & 56.87\%\textcolor{red}{\scriptsize$\downarrow$20.5} & 55.41\%\textcolor{red}{\scriptsize$\downarrow$21.8} & 63.73\%\textcolor{red}{\scriptsize$\downarrow$13.2} & 67.98\%\textcolor{red}{\scriptsize$\downarrow$9.25} & \textcolor{red}{$\downarrow$11.6\%} \\
 &  & HellaSwag & 29.55\%\textcolor{green}{\scriptsize$\uparrow$0.12} & 29.41\%\textcolor{green}{\scriptsize$\uparrow$0.26} & 29.31\%\textcolor{red}{\scriptsize$\downarrow$0.03} & 28.75\%\textcolor{red}{\scriptsize$\downarrow$0.63} & 28.79\%\textcolor{red}{\scriptsize$\downarrow$0.59} & 29.42\%\textcolor{green}{\scriptsize$\uparrow$0.15} & 29.40\%\textcolor{green}{\scriptsize$\uparrow$0.12} & \textcolor{red}{$\downarrow$0.09\%} \\
 \cmidrule{2-11}
 & \multirow{2}{*}{0.1} & BLIMP & 63.89\%\textcolor{red}{\scriptsize$\downarrow$13.8} & 51.85\%\textcolor{red}{\scriptsize$\downarrow$24.9} & 76.47\%\textcolor{red}{\scriptsize$\downarrow$0.06} & 70.30\%\textcolor{red}{\scriptsize$\downarrow$7.07} & 69.65\%\textcolor{red}{\scriptsize$\downarrow$7.56} & 63.57\%\textcolor{red}{\scriptsize$\downarrow$13.3} & 64.14\%\textcolor{red}{\scriptsize$\downarrow$13.1} & \textcolor{red}{$\downarrow$11.4\%} \\
 &  & HellaSwag & 27.82\%\textcolor{red}{\scriptsize$\downarrow$1.60} & 25.03\%\textcolor{red}{\scriptsize$\downarrow$4.11} & 28.69\%\textcolor{red}{\scriptsize$\downarrow$0.65} & 28.91\%\textcolor{red}{\scriptsize$\downarrow$0.47} & 29.22\%\textcolor{red}{\scriptsize$\downarrow$0.16} & 27.41\%\textcolor{red}{\scriptsize$\downarrow$1.85} & 27.66\%\textcolor{red}{\scriptsize$\downarrow$1.61} & \textcolor{red}{$\downarrow$1.49\%} \\
 \midrule
\multirow{6}{*}{\makecell{Large Model\\(0.68B)}} & \multirow{2}{*}{1.0} & BLIMP & 80.47\% & 81.08\% & 80.38\% & 80.98\% & 81.28\% & 81.21\% & 80.88\% \\
 &  & HellaSwag & 37.49\% & 37.52\% & 37.19\% & 37.14\% & 37.32\% & 37.95\% & 37.70\% \\
 \cmidrule{2-11}
 & \multirow{2}{*}{10.0} & BLIMP & 73.96\%\textcolor{red}{\scriptsize$\downarrow$6.51} & 76.56\%\textcolor{red}{\scriptsize$\downarrow$4.52} & 80.52\%\textcolor{green}{\scriptsize$\uparrow$0.13} & 75.05\%\textcolor{red}{\scriptsize$\downarrow$5.93} & 76.45\%\textcolor{red}{\scriptsize$\downarrow$4.83} & 66.34\%\textcolor{red}{\scriptsize$\downarrow$14.9} & 73.37\%\textcolor{red}{\scriptsize$\downarrow$7.51} & \textcolor{red}{$\downarrow$6.29\%} \\
 &  & HellaSwag & 34.37\%\textcolor{red}{\scriptsize$\downarrow$3.12} & 35.30\%\textcolor{red}{\scriptsize$\downarrow$2.22} & 36.72\%\textcolor{red}{\scriptsize$\downarrow$0.48} & 33.44\%\textcolor{red}{\scriptsize$\downarrow$3.70} & 34.92\%\textcolor{red}{\scriptsize$\downarrow$2.40} & 32.12\%\textcolor{red}{\scriptsize$\downarrow$5.84} & 33.98\%\textcolor{red}{\scriptsize$\downarrow$3.72} & \textcolor{red}{$\downarrow$3.07\%} \\
 \cmidrule{2-11}
 & \multirow{2}{*}{0.1} & BLIMP & 67.94\%\textcolor{red}{\scriptsize$\downarrow$12.5} & 71.24\%\textcolor{red}{\scriptsize$\downarrow$9.84} & 79.13\%\textcolor{red}{\scriptsize$\downarrow$1.25} & 71.69\%\textcolor{red}{\scriptsize$\downarrow$9.29} & 74.78\%\textcolor{red}{\scriptsize$\downarrow$6.50} & 66.33\%\textcolor{red}{\scriptsize$\downarrow$14.9} & 68.03\%\textcolor{red}{\scriptsize$\downarrow$12.9} & \textcolor{red}{$\downarrow$9.59\%} \\
 &  & HellaSwag & 32.57\%\textcolor{red}{\scriptsize$\downarrow$4.92} & 35.42\%\textcolor{red}{\scriptsize$\downarrow$2.10} & 36.44\%\textcolor{red}{\scriptsize$\downarrow$0.76} & 33.72\%\textcolor{red}{\scriptsize$\downarrow$3.42} & 35.18\%\textcolor{red}{\scriptsize$\downarrow$2.14} & 32.52\%\textcolor{red}{\scriptsize$\downarrow$5.43} & 33.17\%\textcolor{red}{\scriptsize$\downarrow$4.53} & \textcolor{red}{$\downarrow$3.33\%} \\
\bottomrule
\end{tabular}}
\label{tab:cooperation_or_competition}
\end{table}

%% file: main.bib
@inproceedings{Zhong2024MoExtendTN,
  title={MoExtend: Tuning New Experts for Modality and Task Extension},
  author={Shan Zhong and Shanghua Gao and Zhongzhan Huang and Wushao Wen and Marinka Zitnik and Pan Zhou},
  booktitle={Annual Meeting of the Association for Computational Linguistics},
  year={2024},
  url={https://api.semanticscholar.org/CorpusID:271744833}
}

@inproceedings{Yang2025Qwen3TR,
  title={Qwen3 Technical Report},
  author={An Yang and Anfeng Li and Baosong Yang and Beichen Zhang and Binyuan Hui and Bo Zheng and Bowen Yu and Chang Gao and Chengen Huang and Chenxu Lv and Chujie Zheng and Dayiheng Liu and Fan Zhou and Fei Huang and Feng Hu and Hao Ge and Haoran Wei and Huan Lin and Jialong Tang and Jian Yang and Jianhong Tu and Jianwei Zhang and Jianxin Yang and Jiaxin Yang and Jingren Zhou and Jingren Zhou and Junyan Lin and Kai Dang and Keqin Bao and Ke‐Pei Yang and Le Yu and Li-Chun Deng and Mei Li and Min Xue and Mingze Li and Pei Zhang and Peng Wang and Qin Zhu and Rui Men and Ruize Gao and Shi-Qiang Liu and Shuang Luo and Tianhao Li and Tianyi Tang and Wenbiao Yin and Xingzhang Ren and Xinyu Wang and Xinyu Zhang and Xuancheng Ren and Yang Fan and Yang Su and Yi-Chao Zhang and Yinger Zhang and Yu Wan and Yuqiong Liu and Zekun Wang and Zeyu Cui and Zhenru Zhang and Zhipeng Zhou and Zihan Qiu},
  year={2025},
  url={https://api.semanticscholar.org/CorpusID:278602855}
}

@article{rybakakov2024stability,
  title={Methods of improving LLM training stability},
  author={Oleg Rybakov and Mike Chrzanowski and Peter Dykas and Jinze Xue and Ben Lanir},
  journal={ArXiv},
  year={2024},
  volume={abs/2410.16682},
  url={https://api.semanticscholar.org/CorpusID:273507181}
}

@article{xiong2020_layernorm_transformer,
  title={On Layer Normalization in the Transformer Architecture},
  author={Ruibin Xiong and Yunchang Yang and Di He and Kai Zheng and Shuxin Zheng and Chen Xing and Huishuai Zhang and Yanyan Lan and Liwei Wang and Tie-Yan Liu},
  journal={ArXiv},
  year={2020},
  volume={abs/2002.04745},
  url={https://api.semanticscholar.org/CorpusID:211082816}
}

@article{agarwala_temperature,
  title={Temperature check: theory and practice for training models with softmax-cross-entropy losses},
  author={Atish Agarwala and Jeffrey Pennington and Yann Dauphin and Samuel S. Schoenholz},
  journal={ArXiv},
  year={2020},
  volume={abs/2010.07344},
  url={https://api.semanticscholar.org/CorpusID:222380046}
}

@inproceedings{openai2025gptoss120bgptoss20bmodel,
  title={gpt-oss-120b\&gpt-oss-20b Model Card},
  author={OpenAI Sandhini Agarwal and Lama Ahmad and Jason Ai and Sam Altman and Andy Applebaum and Edwin Arbus and Rahul K. Arora and Yu Bai and Bowen Baker and Hai-Biao Bao and Boaz Barak and Ally Bennett and Tyler Bertao and N. Archer Brett and Eugene Brevdo and Greg Brockman and S{\'e}bastien Bubeck and Cheng Chang and Kai Chen and Mark Chen and Enoch Cheung and Aidan Clark and Dan Cook and Marat Dukhan and C. Dvorak and K Fives and Vlad Fomenko and T. Garipov and Kristian Georgiev and Mia Glaese and Tarun Gogineni and A. B. Goucher and Lukas Gross and Katia Gil Guzman and John Hallman and Jackie Hehir and Johannes Heidecke and Alec Helyar and Haitang Hu and Romain Huet and Jacob Huh and Saachi Jain and Zach Johnson and Chris Koch and Irina Kofman and Dominika Kundel and Jason Kwon and Volodymyr Kyrylov and Elaine Ya Le and Guillaume Leclerc and James Lennon and Scott Lessans and Mario Lezcano-Casado and Yuanzhi Li and Zhuohan Li and Ji Lin and Jordan Liss and Lily Liu and Jiancheng Liu and Kevin Lu and Chris Lu and Zoran N. Martinovic and Lindsay McCallum and Josh McGrath and Scott McKinney and Aidan McLaughlin and Song Mei and Steve Mostovoy and Tong Mu and Gideon Myles and Alexander Neitz and Alex Nichol and Jakub W. Pachocki and Alex Paino and Dana Palmie and Ashley Pantuliano and Giambattista Parascandolo and Jongsoo Park and Leher Pathak and Carolina Paz and Ludovic Peran and Dmitry Pimenov and Michelle Pokrass and Elizabeth Proehl and Huida Qiu and Gaby Raila and Filippo Raso and Hongyu Ren and K. Richardson and David Robinson and Bob Rotsted and Hadi Salman and Suvansh Sanjeev and Max Schwarzer and Daniel Sculley and Harshit S. Sikchi and Kendal Simon and Karan Singhal and Yang Song and Dane Stuckey and Zhiqing Sun and Phil Tillet and Sam Toizer and Foivos Tsimpourlas and Nikhil Vyas and Eric Wallace and Xin Wang and Miles Wang and Olivia Watkins and Kevin Weil and Amy E. Wendling and Kevin Whinnery and Cedric Whitney and Hannah Wong and Lin Yang and Yu Yang and Michihiro Yasunaga and Kristen Ying and Wojciech Zaremba and Wenting Zhan and Cyril Zhang and Brian Hu Zhang and Eddie Zhang and Shengjia Zhao},
  year={2025},
  url={https://api.semanticscholar.org/CorpusID:280671456}
}

@misc{wang2024auxiliarylossfree,
    title={Auxiliary-Loss-Free Load Balancing Strategy for Mixture-of-Experts},
    author={Lean Wang and Huazuo Gao and Chenggang Zhao and Xu Sun and Damai Dai},
    year={2024},
    eprint={2408.15664},
    archivePrefix={arXiv},
    primaryClass={cs.LG}
}

@article{wang2024remoe,
  title={ReMoE: Fully Differentiable Mixture-of-Experts with ReLU Routing},
  author={Ziteng Wang and Jianfei Chen and Jun Zhu},
  journal={ArXiv},
  year={2024},
  volume={abs/2412.14711},
  url={https://api.semanticscholar.org/CorpusID:274859355}
}

@misc{wei2024skyworkmoe,
    title={Skywork-MoE: A Deep Dive into Training Techniques for Mixture-of-Experts Language Models},
    author={Tianwen Wei and Bo Zhu and Liang Zhao and Cheng Cheng and Biye Li and Weiwei Lü and Peng Cheng and Jianhao Zhang and Xiaoyu Zhang and Liang Zeng and Xiaokun Wang and Yutuan Ma and Rui Hu and Shuicheng Yan and Han Fang and Yahui Zhou},
    year={2024},
    eprint={2406.06563},
    archivePrefix={arXiv},
    primaryClass={cs.CL}
}

@article{liu2024ocrbench,
  title={Ocrbench: on the hidden mystery of ocr in large multimodal models},
  author={Liu, Yuliang and Li, Zhang and Huang, Mingxin and Yang, Biao and Yu, Wenwen and Li, Chunyuan and Yin, Xu-Cheng and Liu, Cheng-Lin and Jin, Lianwen and Bai, Xiang},
  journal={Science China Information Sciences},
  volume={67},
  number={12},
  pages={220102},
  year={2024},
  publisher={Springer}
}

@article{zhang2024mme,
  title={Mme-realworld: Could your multimodal llm challenge high-resolution real-world scenarios that are difficult for humans?},
  author={Zhang, Yi-Fan and Zhang, Huanyu and Tian, Haochen and Fu, Chaoyou and Zhang, Shuangqing and Wu, Junfei and Li, Feng and Wang, Kun and Wen, Qingsong and Zhang, Zhang and others},
  journal={arXiv preprint arXiv:2408.13257},
  year={2024}
}

@article{li2024llava,
  title={Llava-onevision: Easy visual task transfer},
  author={Li, Bo and Zhang, Yuanhan and Guo, Dong and Zhang, Renrui and Li, Feng and Zhang, Hao and Zhang, Kaichen and Zhang, Peiyuan and Li, Yanwei and Liu, Ziwei and others},
  journal={arXiv preprint arXiv:2408.03326},
  year={2024}
}

@misc{openai2023gpt4,
    title={GPT-4 Technical Report},
    author={OpenAI and Josh Achiam and Steven Adler and Sandhini Agarwal and Lama Ahmad and Ilge Akkaya and Florencia Leoni Aleman and Diogo Almeida and Janko Altenschmidt and Sam Altman and Shyamal Anadkat and Red Avila and Igor Babuschkin and Suchir Balaji and Valerie Balcom and Paul Baltescu and Haiming Bao and Mohammad Bavarian and Jeff Belgum and Irwan Bello and Jake Berdine and Gabriel Bernadett-Shapiro and Christopher Berner and Lenny Bogdonoff and Oleg Boiko and Madelaine Boyd and Anna-Luisa Brakman and Greg Brockman and Tim Brooks and Miles Brundage and Kevin Button and Trevor Cai and Rosie Campbell and Andrew Cann and Brittany Carey and Chelsea Carlson and Rory Carmichael and Brooke Chan and Che Chang and Fotis Chantzis and Derek Chen and Sully Chen and Ruby Chen and Jason Chen and Mark Chen and Ben Chess and Chester Cho and Casey Chu and Hyung Won Chung and Dave Cummings and Jeremiah Currier and Yunxing Dai and Cory Decareaux and Thomas Degry and Noah Deutsch and Damien Deville and Arka Dhar and David Dohan and Steve Dowling and Sheila Dunning and Adrien Ecoffet and Atty Eleti and Tyna Eloundou and David Farhi and Liam Fedus and Niko Felix and Simón Posada Fishman and Juston Forte and Isabella Fulford and Leo Gao and Elie Georges and Christian Gibson and Vik Goel and Tarun Gogineni and Gabriel Goh and Rapha Gontijo-Lopes and Jonathan Gordon and Morgan Grafstein and Scott Gray and Ryan Greene and Joshua Gross and Shixiang Shane Gu and Yufei Guo and Chris Hallacy and Jesse Han and Jeff Harris and Yuchen He and Mike Heaton and Johannes Heidecke and Chris Hesse and Alan Hickey and Wade Hickey and Peter Hoeschele and Brandon Houghton and Kenny Hsu and Shengli Hu and Xin Hu and Joost Huizinga and Shantanu Jain and Shawn Jain and Joanne Jang and Angela Jiang and Roger Jiang and Haozhun Jin and Denny Jin and Shino Jomoto and Billie Jonn and Heewoo Jun and Tomer Kaftan and Łukasz Kaiser and Ali Kamali and Ingmar Kanitscheider and Nitish Shirish Keskar and Tabarak Khan and Logan Kilpatrick and Jong Wook Kim and Christina Kim and Yongjik Kim and Jan Hendrik Kirchner and Jamie Kiros and Matt Knight and Daniel Kokotajlo and Łukasz Kondraciuk and Andrew Kondrich and Aris Konstantinidis and Kyle Kosic and Gretchen Krueger and Vishal Kuo and Michael Lampe and Ikai Lan and Teddy Lee and Jan Leike and Jade Leung and Daniel Levy and Chak Ming Li and Rachel Lim and Molly Lin and Stephanie Lin and Mateusz Litwin and Theresa Lopez and Ryan Lowe and Patricia Lue and Anna Makanju and Kim Malfacini and Sam Manning and Todor Markov and Yaniv Markovski and Bianca Martin and Katie Mayer and Andrew Mayne and Bob McGrew and Scott Mayer McKinney and Christine McLeavey and Paul McMillan and Jake McNeil and David Medina and Aalok Mehta and Jacob Menick and Luke Metz and Andrey Mishchenko and Pamela Mishkin and Vinnie Monaco and Evan Morikawa and Daniel Mossing and Tong Mu and Mira Murati and Oleg Murk and David Mély and Ashvin Nair and Reiichiro Nakano and Rajeev Nayak and Arvind Neelakantan and Richard Ngo and Hyeonwoo Noh and Long Ouyang and Cullen O'Keefe and Jakub Pachocki and Alex Paino and Joe Palermo and Ashley Pantuliano and Giambattista Parascandolo and Joel Parish and Emy Parparita and Alex Passos and Mikhail Pavlov and Andrew Peng and Adam Perelman and Filipe de Avila Belbute Peres and Michael Petrov and Henrique Ponde de Oliveira Pinto and Michael and Pokorny and Michelle Pokrass and Vitchyr H. Pong and Tolly Powell and Alethea Power and Boris Power and Elizabeth Proehl and Raul Puri and Alec Radford and Jack Rae and Aditya Ramesh and Cameron Raymond and Francis Real and Kendra Rimbach and Carl Ross and Bob Rotsted and Henri Roussez and Nick Ryder and Mario Saltarelli and Ted Sanders and Shibani Santurkar and Girish Sastry and Heather Schmidt and David Schnurr and John Schulman and Daniel Selsam and Kyla Sheppard and Toki Sherbakov and Jessica Shieh and Sarah Shoker and Pranav Shyam and Szymon Sidor and Eric Sigler and Maddie Simens and Jordan Sitkin and Katarina Slama and Ian Sohl and Benjamin Sokolowsky and Yang Song and Natalie Staudacher and Felipe Petroski Such and Natalie Summers and Ilya Sutskever and Jie Tang and Nikolas Tezak and Madeleine B. Thompson and Phil Tillet and Amin Tootoonchian and Elizabeth Tseng and Preston Tuggle and Nick Turley and Jerry Tworek and Juan Felipe Cerón Uribe and Andrea Vallone and Arun Vijayvergiya and Chelsea Voss and Carroll Wainwright and Justin Jay Wang and Alvin Wang and Ben Wang and Jonathan Ward and Jason Wei and CJ Weinmann and Akila Welihinda and Peter Welinder and Jiayi Weng and Lilian Weng and Matt Wiethoff and Dave Willner and Clemens Winter and Samuel Wolrich and Hannah Wong and Lauren Workman and Sherwin Wu and Jeff Wu and Michael Wu and Kai Xiao and Tao Xu and Sarah Yoo and Kevin Yu and Qiming Yuan and Wojciech Zaremba and Rowan Zellers and Chong Zhang and Marvin Zhang and Shengjia Zhao and Tianhao Zheng and Juntang Zhuang and William Zhuk and Barret Zoph},
    year={2023},
    eprint={2303.08774},
    archivePrefix={arXiv},
    primaryClass={cs.CL}
}

@misc{nguyen2024least,
      title={On Least Square Estimation in Softmax Gating Mixture of Experts}, 
      author={Huy Nguyen and Nhat Ho and Alessandro Rinaldo},
      year={2024},
      eprint={2402.02952},
      archivePrefix={arXiv},
      primaryClass={stat.ML},
      url={https://arxiv.org/abs/2402.02952}, 
}

@article{han2024fusemoe,
  title={Fusemoe: Mixture-of-experts transformers for fleximodal fusion},
  author={Han, Xing and Nguyen, Huy and Harris, Carl and Ho, Nhat and Saria, Suchi},
  journal={arXiv preprint arXiv:2402.03226},
  year={2024}
}

@article{abdin2024phi3tr,
  title={Phi-3 Technical Report: A Highly Capable Language Model Locally on Your Phone},
  author={Marah Abdin and Sam Ade Jacobs and Ammar Ahmad Awan and Jyoti Aneja and Ahmed Awadallah and Hany Hassan Awadalla and Nguyen Bach and Amit Bahree and Arash Bakhtiari and Harkirat Singh Behl and Alon Benhaim and Misha Bilenko and Johan Bjorck and S{\'e}bastien Bubeck and Martin Cai and Caio C'esar Teodoro Mendes and Weizhu Chen and Vishrav Chaudhary and Parul Chopra and Allison Del Giorno and Gustavo de Rosa and Matthew Dixon and Ronen Eldan and Victor Fragoso and Dan Iter and Abhishek Goswami and Suriya Gunasekar and Emman Haider and Junheng Hao and Russell J. Hewett and Jamie Huynh and Mojan Javaheripi and Xin Jin and Piero Kauffmann and Nikos Karampatziakis and Dongwoo Kim and Young Jin Kim and Mahoud Khademi and Lev Kurilenko and James R. Lee and Yin Tat Lee and Yuanzhi Li and Chen Liang and Weishung Liu and Eric Lin and Zeqi Lin and Piyush Madan and Arindam Mitra and Hardik Modi and Anh Hong Nguyen and Brandon Norick and Barun Patra and Daniel Perez-Becker and Thomas Portet and Reid Pryzant and Heyang Qin and Marko Radmilac and Liliang Ren and Corby Rosset and Sambudha Roy and Olli Saarikivi and Amin Saied and Adil Salim and Michael Santacroce and Shital Shah and Ning Shang and Hiteshi Sharma and Xianmin Song and Olatunji Ruwase and Praneetha Vaddamanu and Xin Wang and Rachel Ward and Guanhua Wang and Philipp Andre Witte and Michael Wyatt and Can Xu and Jiahang Xu and Sonal Yadav and Fan Yang and Ziyi Yang and Donghan Yu and Cheng-yuan Zhang and Cyril Zhang and Jianwen Zhang and Li Lyna Zhang and Yi Zhang and Yunan Zhang and Xiren Zhou and Yifan Yang},
  journal={ArXiv},
  year={2024},
  volume={abs/2404.14219},
  url={https://api.semanticscholar.org/CorpusID:269293048}
}

@inproceedings{Lu2019ViLBERTPT,
  title={ViLBERT: Pretraining Task-Agnostic Visiolinguistic Representations for Vision-and-Language Tasks},
  author={Jiasen Lu and Dhruv Batra and Devi Parikh and Stefan Lee},
  booktitle={Neural Information Processing Systems},
  year={2019},
  url={https://api.semanticscholar.org/CorpusID:199453025}
}

@misc{zhang2024lmmsevalrealitycheckevaluation,
      title={LMMs-Eval: Reality Check on the Evaluation of Large Multimodal Models}, 
      author={Kaichen Zhang and Bo Li and Peiyuan Zhang and Fanyi Pu and Joshua Adrian Cahyono and Kairui Hu and Shuai Liu and Yuanhan Zhang and Jingkang Yang and Chunyuan Li and Ziwei Liu},
      year={2024},
      eprint={2407.12772},
      archivePrefix={arXiv},
      primaryClass={cs.CL},
      url={https://arxiv.org/abs/2407.12772}, 
}

@article{Lian2024UniversalCE,
  title={Universal Checkpointing: Efficient and Flexible Checkpointing for Large Scale Distributed Training},
  author={Xinyu Lian and Sam Ade Jacobs and Lev Kurilenko and Masahiro Tanaka and Stas Bekman and Olatunji Ruwase and Minjia Zhang},
  journal={ArXiv},
  year={2024},
  volume={abs/2406.18820},
  url={https://api.semanticscholar.org/CorpusID:270764954}
}

@article{yang2024qwen2tr,
  title={Qwen2 Technical Report},
  author={An Yang and Baosong Yang and Binyuan Hui and Bo Zheng and Bowen Yu and Chang Zhou and Chengpeng Li and Chengyuan Li and Dayiheng Liu and Fei Huang and Guanting Dong and Haoran Wei and Huan Lin and Jialong Tang and Jialin Wang and Jian Yang and Jianhong Tu and Jianwei Zhang and Jianxin Ma and Jin Xu and Jingren Zhou and Jinze Bai and Jinzheng He and Junyang Lin and Kai Dang and Keming Lu and Ke-Yang Chen and Kexin Yang and Mei Li and Min Xue and Na Ni and Pei Zhang and Peng Wang and Ru Peng and Rui Men and Ruize Gao and Runji Lin and Shijie Wang and Shuai Bai and Sinan Tan and Tianhang Zhu and Tianhao Li and Tianyu Liu and Wenbin Ge and Xiaodong Deng and Xiaohuan Zhou and Xingzhang Ren and Xinyu Zhang and Xipin Wei and Xuancheng Ren and Yang Fan and Yang Yao and Yichang Zhang and Yunyang Wan and Yunfei Chu and Zeyu Cui and Zhenru Zhang and Zhi-Wei Fan},
  journal={ArXiv},
  year={2024},
  volume={abs/2407.10671},
  url={https://api.semanticscholar.org/CorpusID:271212307}
}

@article{jacobs1991adaptive,
  title={Adaptive mixtures of local experts},
  author={Jacobs, Robert A and Jordan, Michael I and Nowlan, Steven J and Hinton, Geoffrey E},
  journal={Neural computation},
  volume={3},
  number={1},
  pages={79--87},
  year={1991},
  publisher={MIT Press}
}

@article{shazeer2017outrageously,
  title={Outrageously large neural networks: The sparsely-gated mixture-of-experts layer},
  author={Shazeer, Noam and Mirhoseini, Azalia and Maziarz, Krzysztof and Davis, Andy and Le, Quoc and Hinton, Geoffrey and Dean, Jeff},
  journal={arXiv preprint arXiv:1701.06538},
  year={2017}
}

@article{fedus2022switch,
  title={Switch transformers: Scaling to trillion parameter models with simple and efficient sparsity},
  author={Fedus, William and Zoph, Barret and Shazeer, Noam},
  journal={Journal of Machine Learning Research},
  volume={23},
  number={120},
  pages={1--39},
  year={2022}
}

@article{lepikhin2020gshard,
  title={Gshard: Scaling giant models with conditional computation and automatic sharding},
  author={Lepikhin, Dmitry and Lee, HyoukJoong and Xu, Yuanzhong and Chen, Dehao and Firat, Orhan and Huang, Yanping and Krikun, Maxim and Shazeer, Noam and Chen, Zhifeng},
  journal={arXiv preprint arXiv:2006.16668},
  year={2020}
}

@article{komatsuzaki2022sparse,
  title={Sparse upcycling: Training mixture-of-experts from dense checkpoints},
  author={Komatsuzaki, Aran and Puigcerver, Joan and Lee-Thorp, James and Ruiz, Carlos Riquelme and Mustafa, Basil and Ainslie, Joshua and Tay, Yi and Dehghani, Mostafa and Houlsby, Neil},
  journal={arXiv preprint arXiv:2212.05055},
  year={2022}
}

@inproceedings{liu2023llava,
    author      = {Liu, Haotian and Li, Chunyuan and Wu, Qingyang and Lee, Yong Jae},
    title       = {Visual Instruction Tuning},
    booktitle   = {NeurIPS},
    year        = {2023}
}

@article{hwang2023tutel,
  title={Tutel: Adaptive mixture-of-experts at scale},
  author={Hwang, Changho and Cui, Wei and Xiong, Yifan and Yang, Ziyue and Liu, Ze and Hu, Han and Wang, Zilong and Salas, Rafael and Jose, Jithin and Ram, Prabhat and others},
  journal={Proceedings of Machine Learning and Systems},
  volume={5},
  pages={269--287},
  year={2023}
}

@article{he2021fastmoe,
  title={Fastmoe: A fast mixture-of-expert training system},
  author={He, Jiaao and Qiu, Jiezhong and Zeng, Aohan and Yang, Zhilin and Zhai, Jidong and Tang, Jie},
  journal={arXiv preprint arXiv:2103.13262},
  year={2021}
}

@article{xue2024openmoe,
  title={Openmoe: An early effort on open mixture-of-experts language models},
  author={Xue, Fuzhao and Zheng, Zian and Fu, Yao and Ni, Jinjie and Zheng, Zangwei and Zhou, Wangchunshu and You, Yang},
  journal={arXiv preprint arXiv:2402.01739},
  year={2024}
}

@article{roller2021hash,
  title={Hash layers for large sparse models},
  author={Roller, Stephen and Sukhbaatar, Sainbayar and Weston, Jason and others},
  journal={Advances in Neural Information Processing Systems},
  volume={34},
  pages={17555--17566},
  year={2021}
}

@article{zoph2022st,
  title={St-moe: Designing stable and transferable sparse expert models},
  author={Zoph, Barret and Bello, Irwan and Kumar, Sameer and Du, Nan and Huang, Yanping and Dean, Jeff and Shazeer, Noam and Fedus, William},
  journal={arXiv preprint arXiv:2202.08906},
  year={2022}
}

@article{dai2022stablemoe,
  title={Stablemoe: Stable routing strategy for mixture of experts},
  author={Dai, Damai and Dong, Li and Ma, Shuming and Zheng, Bo and Sui, Zhifang and Chang, Baobao and Wei, Furu},
  journal={arXiv preprint arXiv:2204.08396},
  year={2022}
}

@article{dai2024deepseekmoe,
  title={Deepseekmoe: Towards ultimate expert specialization in mixture-of-experts language models},
  author={Dai, Damai and Deng, Chengqi and Zhao, Chenggang and Xu, RX and Gao, Huazuo and Chen, Deli and Li, Jiashi and Zeng, Wangding and Yu, Xingkai and Wu, Y and others},
  journal={arXiv preprint arXiv:2401.06066},
  year={2024}
}

@article{li2024cumo,
  title={Cumo: Scaling multimodal llm with co-upcycled mixture-of-experts},
  author={Li, Jiachen and Wang, Xinyao and Zhu, Sijie and Kuo, Chia-Wen and Xu, Lu and Chen, Fan and Jain, Jitesh and Shi, Humphrey and Wen, Longyin},
  journal={arXiv preprint arXiv:2405.05949},
  year={2024}
}

@article{Kembhavi2016ADI,
  title={A Diagram is Worth a Dozen Images},
  author={Aniruddha Kembhavi and Michael Salvato and Eric Kolve and Minjoon Seo and Hannaneh Hajishirzi and Ali Farhadi},
  journal={ArXiv},
  year={2016},
  volume={abs/1603.07396},
  url={https://api.semanticscholar.org/CorpusID:2682274}
}

@article{Singh2019TowardsVM,
  title={Towards VQA Models That Can Read},
  author={Amanpreet Singh and Vivek Natarajan and Meet Shah and Yu Jiang and Xinlei Chen and Dhruv Batra and Devi Parikh and Marcus Rohrbach},
  journal={2019 IEEE/CVF Conference on Computer Vision and Pattern Recognition (CVPR)},
  year={2019},
  pages={8309-8318},
  url={https://api.semanticscholar.org/CorpusID:85553602}
}

@misc{Hudson2019GQAA,
      title={GQA: A New Dataset for Real-World Visual Reasoning and Compositional Question Answering}, 
      author={Drew A. Hudson and Christopher D. Manning},
      year={2019},
      eprint={1902.09506},
      archivePrefix={arXiv},
      primaryClass={cs.CL},
      url={https://arxiv.org/abs/1902.09506}, 
}

@inproceedings{Guan2023HallusionBenchAA,
  title={HallusionBench: An Advanced Diagnostic Suite for Entangled Language Hallucination and Visual Illusion in Large Vision-Language Models},
  author={Tianrui Guan and Fuxiao Liu and Xiyang Wu and Ruiqi Xian and Zongxia Li and Xiaoyu Liu and Xiyang Wu and Xijun Wang and Lichang Chen and Furong Huang and Yaser Yacoob and Dinesh Manocha and Tianyi Zhou},
  booktitle={arXiv preprint},
  year={2023},
  url={https://api.semanticscholar.org/CorpusID:265499116}
}

@inproceedings{Lu2023MathVistaEM,
  title={MathVista: Evaluating Mathematical Reasoning of Foundation Models in Visual Contexts},
  author={Pan Lu and Hritik Bansal and Tony Xia and Jiacheng Liu and Chun-yue Li and Hannaneh Hajishirzi and Hao Cheng and Kai-Wei Chang and Michel Galley and Jianfeng Gao},
  booktitle={International Conference on Learning Representations},
  year={2023},
  url={https://api.semanticscholar.org/CorpusID:264491155}
}

@article{Liu2023MMBenchIY,
  title={MMBench: Is Your Multi-modal Model an All-around Player?},
  author={Yuanzhan Liu and Haodong Duan and Yuanhan Zhang and Bo Li and Songyang Zhang and Wangbo Zhao and Yike Yuan and Jiaqi Wang and Conghui He and Ziwei Liu and Kai Chen and Dahua Lin},
  journal={ArXiv},
  year={2023},
  volume={abs/2307.06281},
  url={https://api.semanticscholar.org/CorpusID:259837088}
}

@article{Fu2023MMEAC,
  title={MME: A Comprehensive Evaluation Benchmark for Multimodal Large Language Models},
  author={Chaoyou Fu and Peixian Chen and Yunhang Shen and Yulei Qin and Mengdan Zhang and Xu Lin and Zhenyu Qiu and Wei Lin and Jinrui Yang and Xiawu Zheng and Ke Li and Xing Sun and Rongrong Ji},
  journal={ArXiv},
  year={2023},
  volume={abs/2306.13394},
  url={https://api.semanticscholar.org/CorpusID:259243928}
}

@article{Yue2023MMMUAM,
  title={MMMU: A Massive Multi-discipline Multimodal Understanding and Reasoning Benchmark for Expert AGI},
  author={Xiang Yue and Yuansheng Ni and Kai Zhang and Tianyu Zheng and Ruoqi Liu and Ge Zhang and Samuel Stevens and Dongfu Jiang and Weiming Ren and Yuxuan Sun and Cong Wei and Botao Yu and Ruibin Yuan and Renliang Sun and Ming Yin and Boyuan Zheng and Zhenzhu Yang and Yibo Liu and Wenhao Huang and Huan Sun and Yu Su and Wenhu Chen},
  journal={ArXiv},
  year={2023},
  volume={abs/2311.16502},
  url={https://api.semanticscholar.org/CorpusID:265466525}
}

@article{Chen2024AreWO,
  title={Are We on the Right Way for Evaluating Large Vision-Language Models?},
  author={Lin Chen and Jinsong Li and Xiao-wen Dong and Pan Zhang and Yuhang Zang and Zehui Chen and Haodong Duan and Jiaqi Wang and Yu Qiao and Dahua Lin and Feng Zhao},
  journal={ArXiv},
  year={2024},
  volume={abs/2403.20330},
  url={https://api.semanticscholar.org/CorpusID:268793433}
}

@inproceedings{Li2023EvaluatingOH,
  title={Evaluating Object Hallucination in Large Vision-Language Models},
  author={Yifan Li and Yifan Du and Kun Zhou and Jinpeng Wang and Wayne Xin Zhao and Ji-rong Wen},
  booktitle={Conference on Empirical Methods in Natural Language Processing},
  year={2023},
  url={https://api.semanticscholar.org/CorpusID:258740697}
}

@article{zuo2021taming,
  title={Taming sparsely activated transformer with stochastic experts},
  author={Zuo, Simiao and Liu, Xiaodong and Jiao, Jian and Kim, Young Jin and Hassan, Hany and Zhang, Ruofei and Zhao, Tuo and Gao, Jianfeng},
  journal={arXiv preprint arXiv:2110.04260},
  year={2021}
}

@article{chi2022representation,
  title={On the representation collapse of sparse mixture of experts},
  author={Chi, Zewen and Dong, Li and Huang, Shaohan and Dai, Damai and Ma, Shuming and Patra, Barun and Singhal, Saksham and Bajaj, Payal and Song, Xia and Mao, Xian-Ling and others},
  journal={Advances in Neural Information Processing Systems},
  volume={35},
  pages={34600--34613},
  year={2022}
}

@inproceedings{zhai2023sigmoid,
  title={Sigmoid loss for language image pre-training},
  author={Zhai, Xiaohua and Mustafa, Basil and Kolesnikov, Alexander and Beyer, Lucas},
  booktitle={Proceedings of the IEEE/CVF International Conference on Computer Vision},
  pages={11975--11986},
  year={2023}
}

@article{chen2024allava,
  title={Allava: Harnessing gpt4v-synthesized data for a lite vision-language model},
  author={Chen, Guiming Hardy and Chen, Shunian and Zhang, Ruifei and Chen, Junying and Wu, Xiangbo and Zhang, Zhiyi and Chen, Zhihong and Li, Jianquan and Wan, Xiang and Wang, Benyou},
  journal={arXiv preprint arXiv:2402.11684},
  year={2024}
}

@article{do2023hyperrouter,
  title={HyperRouter: Towards efficient training and inference of sparse mixture of experts},
  author={Do, Giang and Le, Khiem and Pham, Quang and Nguyen, Trungtin and Doan, Thanh-Nam and Nguyen, Bint T and Liu, Chenghao and Ramasamy, Savitha and Li, Xiaoli and Hoi, Steven},
  journal={arXiv preprint arXiv:2312.07035},
  year={2023}
}

@misc{Cerebras_2025,
      title={BTLM-3B-8K: 7B Parameter Performance in a 3B Parameter Model}, 
      author={Nolan Dey and Daria Soboleva and Faisal Al-Khateeb and Bowen Yang and Ribhu Pathria and Hemant Khachane and Shaheer Muhammad and Zhiming and Chen and Robert Myers and Jacob Robert Steeves and Natalia Vassilieva and Marvin Tom and Joel Hestness},
      year={2023},
      eprint={2309.11568},
      archivePrefix={arXiv},
      primaryClass={cs.AI},
      url={https://arxiv.org/abs/2309.11568}, 
}

@article{gupta2023continual,
  title={Continual pre-training of large language models: How to (re) warm your model?},
  author={Gupta, Kshitij and Th{\'e}rien, Benjamin and Ibrahim, Adam and Richter, Mats L and Anthony, Quentin and Belilovsky, Eugene and Rish, Irina and Lesort, Timoth{\'e}e},
  journal={arXiv preprint arXiv:2308.04014},
  year={2023}
}

@article{agarwalla2024enabling,
  title={Enabling high-sparsity foundational llama models with efficient pretraining and deployment},
  author={Agarwalla, Abhinav and Gupta, Abhay and Marques, Alexandre and Pandit, Shubhra and Goin, Michael and Kurtic, Eldar and Leong, Kevin and Nguyen, Tuan and Salem, Mahmoud and Alistarh, Dan and others},
  journal={arXiv preprint arXiv:2405.03594},
  year={2024}
}

@article{zhang2024tinyllama,
  title={Tinyllama: An open-source small language model},
  author={Zhang, Peiyuan and Zeng, Guangtao and Wang, Tianduo and Lu, Wei},
  journal={arXiv preprint arXiv:2401.02385},
  year={2024}
}

@misc{cerebras2023slimpajama,
    author = {Soboleva, Daria and Al-Khateeb, Faisal and Myers, Robert and Steeves, Jacob R and Hestness, Joel and Dey, Nolan},
    title = {{SlimPajama: A 627B token cleaned and deduplicated version of RedPajama}},
    month = {June},
    year = 2023,
    howpublished = {\url{https://cerebras.ai/blog/slimpajama-a-627b-token-cleaned-and-deduplicated-version-of-redpajama}},
    url = {https://huggingface.co/datasets/cerebras/SlimPajama-627B},
}

@misc{together2023redpajama,
  author = {Together Computer},
  title = {RedPajama: An Open Source Recipe to Reproduce LLaMA training dataset},
  month = {April},
  year = 2023,
  url = {https://github.com/togethercomputer/RedPajama-Data}
}

@article{kudo2018sentencepiece,
  title={SentencePiece: A simple and language independent subword tokenizer and detokenizer for neural text processing},
  author={Kudo, Taku and Richardson, John},
  journal={arXiv preprint arXiv:1808.06226},
  year={2018}
}

@article{paperno2016lambada,
  title={The LAMBADA dataset: Word prediction requiring a broad discourse context},
  author={Paperno, Denis and Kruszewski, Germ{\'a}n and Lazaridou, Angeliki and Pham, Quan Ngoc and Bernardi, Raffaella and Pezzelle, Sandro and Baroni, Marco and Boleda, Gemma and Fern{\'a}ndez, Raquel},
  journal={arXiv preprint arXiv:1606.06031},
  year={2016}
}

@article{warstadt2020blimp,
  title={BLiMP: The benchmark of linguistic minimal pairs for English},
  author={Warstadt, Alex and Parrish, Alicia and Liu, Haokun and Mohananey, Anhad and Peng, Wei and Wang, Sheng-Fu and Bowman, Samuel R},
  journal={Transactions of the Association for Computational Linguistics},
  volume={8},
  pages={377--392},
  year={2020},
  publisher={MIT Press One Rogers Street, Cambridge, MA 02142-1209, USA journals-info~…}
}

@article{hill2015goldilocks,
  title={The goldilocks principle: Reading children's books with explicit memory representations},
  author={Hill, Felix and Bordes, Antoine and Chopra, Sumit and Weston, Jason},
  journal={arXiv preprint arXiv:1511.02301},
  year={2015}
}

@article{zellers2019hellaswag,
  title={Hellaswag: Can a machine really finish your sentence?},
  author={Zellers, Rowan and Holtzman, Ari and Bisk, Yonatan and Farhadi, Ali and Choi, Yejin},
  journal={arXiv preprint arXiv:1905.07830},
  year={2019}
}

@inproceedings{bisk2020piqa,
  title={Piqa: Reasoning about physical commonsense in natural language},
  author={Bisk, Yonatan and Zellers, Rowan and Gao, Jianfeng and Choi, Yejin and others},
  booktitle={Proceedings of the AAAI conference on artificial intelligence},
  volume={34},
  number={05},
  pages={7432--7439},
  year={2020}
}

@article{clark2018think,
  title={Think you have solved question answering? try arc, the ai2 reasoning challenge},
  author={Clark, Peter and Cowhey, Isaac and Etzioni, Oren and Khot, Tushar and Sabharwal, Ashish and Schoenick, Carissa and Tafjord, Oyvind},
  journal={arXiv preprint arXiv:1803.05457},
  year={2018}
}

@inproceedings{lai-etal-2017-race,
    title = "{RACE}: Large-scale {R}e{A}ding Comprehension Dataset From Examinations",
    author = "Lai, Guokun  and
      Xie, Qizhe  and
      Liu, Hanxiao  and
      Yang, Yiming  and
      Hovy, Eduard",
    booktitle = "Proceedings of the 2017 Conference on Empirical Methods in Natural Language Processing",
    month = sep,
    year = "2017",
    address = "Copenhagen, Denmark",
    publisher = "Association for Computational Linguistics",
    url = "https://aclanthology.org/D17-1082",
    doi = "10.18653/v1/D17-1082",
    pages = "785--794",
}

@article{sap2019socialiqa,
  title={Socialiqa: Commonsense reasoning about social interactions},
  author={Sap, Maarten and Rashkin, Hannah and Chen, Derek and LeBras, Ronan and Choi, Yejin},
  journal={arXiv preprint arXiv:1904.09728},
  year={2019}
}

@article{talmor2018commonsenseqa,
  title={Commonsenseqa: A question answering challenge targeting commonsense knowledge},
  author={Talmor, Alon and Herzig, Jonathan and Lourie, Nicholas and Berant, Jonathan},
  journal={arXiv preprint arXiv:1811.00937},
  year={2018}
}

@inproceedings{yan2025tc,
  title={TC-MoE: Augmenting Mixture of Experts with Ternary Expert Choice},
  author={Yan, Shen and Bin, Xingyan and Zhang, Sijun and Wang, Yisen and Lin, Zhouchen},
  booktitle={The Thirteenth International Conference on Learning Representations},
  year={2025}
}

@article{jin2024moe++,
  title={Moe++: Accelerating mixture-of-experts methods with zero-computation experts},
  author={Jin, Peng and Zhu, Bo and Yuan, Li and Yan, Shuicheng},
  journal={arXiv preprint arXiv:2410.07348},
  year={2024}
}

@article{liu2024deepseekv3,
  title={Deepseek-v3 technical report},
  author={Liu, Aixin and Feng, Bei and Xue, Bing and Wang, Bingxuan and Wu, Bochao and Lu, Chengda and Zhao, Chenggang and Deng, Chengqi and Zhang, Chenyu and Ruan, Chong and others},
  journal={arXiv preprint arXiv:2412.19437},
  year={2024}
}

@article{liu2024deepseekv2,
  title={Deepseek-v2: A strong, economical, and efficient mixture-of-experts language model},
  author={Liu, Aixin and Feng, Bei and Wang, Bin and Wang, Bingxuan and Liu, Bo and Zhao, Chenggang and Dengr, Chengqi and Ruan, Chong and Dai, Damai and Guo, Daya and others},
  journal={arXiv preprint arXiv:2405.04434},
  year={2024}
}

@article{muennighoff2024olmoe,
  title={Olmoe: Open mixture-of-experts language models},
  author={Muennighoff, Niklas and Soldaini, Luca and Groeneveld, Dirk and Lo, Kyle and Morrison, Jacob and Min, Sewon and Shi, Weijia and Walsh, Pete and Tafjord, Oyvind and Lambert, Nathan and others},
  journal={arXiv preprint arXiv:2409.02060},
  year={2024}
}

@article{kang2025flame,
  title={FLAME-MoE: A Transparent End-to-End Research Platform for Mixture-of-Experts Language Models},
  author={Kang, Hao and Yu, Zichun and Xiong, Chenyan},
  journal={arXiv preprint arXiv:2505.20225},
  year={2025}
}

@article{nguyen2025deepseekmoe,
  title={On DeepSeekMoE: Statistical Benefits of Shared Experts and Normalized Sigmoid Gating},
  author={Nguyen, Huy and Doan, Thong T and Pham, Quang and Bui, Nghi DQ and Ho, Nhat and Rinaldo, Alessandro},
  journal={arXiv preprint arXiv:2505.10860},
  year={2025}
}

@article{raffel2020exploring,
  title={Exploring the limits of transfer learning with a unified text-to-text transformer},
  author={Raffel, Colin and Shazeer, Noam and Roberts, Adam and Lee, Katherine and Narang, Sharan and Matena, Michael and Zhou, Yanqi and Li, Wei and Liu, Peter J},
  journal={Journal of machine learning research},
  volume={21},
  number={140},
  pages={1--67},
  year={2020}
}

@article{jiang2024mixtral,
  title={Mixtral of experts},
  author={Jiang, Albert Q and Sablayrolles, Alexandre and Roux, Antoine and Mensch, Arthur and Savary, Blanche and Bamford, Chris and Chaplot, Devendra Singh and Casas, Diego de las and Hanna, Emma Bou and Bressand, Florian and others},
  journal={arXiv preprint arXiv:2401.04088},
  year={2024}
}

@article{comanici2025gemini,
  title={Gemini 2.5: Pushing the frontier with advanced reasoning, multimodality, long context, and next generation agentic capabilities},
  author={Comanici, Gheorghe and Bieber, Eric and Schaekermann, Mike and Pasupat, Ice and Sachdeva, Noveen and Dhillon, Inderjit and Blistein, Marcel and Ram, Ori and Zhang, Dan and Rosen, Evan and others},
  journal={arXiv preprint arXiv:2507.06261},
  year={2025}
}

@article{han2024vimoe,
  title={Vimoe: An empirical study of designing vision mixture-of-experts},
  author={Han, Xumeng and Wei, Longhui and Dou, Zhiyang and Wang, Zipeng and Qiang, Chenhui and He, Xin and Sun, Yingfei and Han, Zhenjun and Tian, Qi},
  journal={arXiv preprint arXiv:2410.15732},
  year={2024}
}

@inproceedings{fan2022m3vit,
 author = {liang, hanxue and Fan, Zhiwen and Sarkar, Rishov and Jiang, Ziyu and Chen, Tianlong and Zou, Kai and Cheng, Yu and Hao, Cong and Wang, Zhangyang},
 booktitle = {Advances in Neural Information Processing Systems},
 editor = {S. Koyejo and S. Mohamed and A. Agarwal and D. Belgrave and K. Cho and A. Oh},
 pages = {28441--28457},
 publisher = {Curran Associates, Inc.},
 title = {{M\textsuperscript{3}ViT}: Mixture-of-Experts Vision Transformer for Efficient Multi-task Learning with Model-Accelerator Co-design},
 url = {https://proceedings.neurips.cc/paper_files/paper/2022/file/b653f34d576d1790481e3797cb740214-Paper-Conference.pdf},
 volume = {35},
 year = {2022}
}

@article{riquelme2021scaling,
  title={Scaling vision with sparse mixture of experts},
  author={Riquelme, Carlos and Puigcerver, Joan and Mustafa, Basil and Neumann, Maxim and Jenatton, Rodolphe and Susano Pinto, Andr{\'e} and Keysers, Daniel and Houlsby, Neil},
  journal={Advances in Neural Information Processing Systems},
  volume={34},
  pages={8583--8595},
  year={2021}
}

@article{lin2024moe,
  title={MoE-LLaVA: Mixture of Experts for Large Vision-Language Models},
  author={Bin Lin and Zhenyu Tang and Yang Ye and Jiaxi Cui and Bin Zhu and Peng Jin and Jinfa Huang and Junwu Zhang and Munan Ning and Li Yuan},
  journal={ArXiv},
  year={2024},
  volume={abs/2401.15947},
  url={https://api.semanticscholar.org/CorpusID:267311517}
}

@article{yun2024flex,
  title={Flex-moe: Modeling arbitrary modality combination via the flexible mixture-of-experts},
  author={Yun, Sukwon and Choi, Inyoung and Peng, Jie and Wu, Yangfan and Bao, Jingxuan and Zhang, Qiyiwen and Xin, Jiayi and Long, Qi and Chen, Tianlong},
  journal={Advances in Neural Information Processing Systems},
  volume={37},
  pages={98782--98805},
  year={2024}
}

@article{csordas2023approximating,
  title={Approximating two-layer feedforward networks for efficient transformers},
  author={Csord{\'a}s, R{\'o}bert and Irie, Kazuki and Schmidhuber, J{\"u}rgen},
  journal={arXiv preprint arXiv:2310.10837},
  year={2023}
}

@article{jordan1994hierarchical,
  title={Hierarchical mixtures of experts and the EM algorithm},
  author={Jordan, Michael I and Jacobs, Robert A},
  journal={Neural computation},
  volume={6},
  number={2},
  pages={181--214},
  year={1994},
  publisher={MIT Press}
}

@article{chen1999improved,
  title={Improved learning algorithms for mixture of experts in multiclass classification},
  author={Chen, Ke and Xu, Lei and Chi, Huisheng},
  journal={Neural networks},
  volume={12},
  number={9},
  pages={1229--1252},
  year={1999},
  publisher={Elsevier}
}

@article{gales2006product,
  title={Product of Gaussians for speech recognition},
  author={Gales, Mark JF and Airey, SS},
  journal={Computer Speech \& Language},
  volume={20},
  number={1},
  pages={22--40},
  year={2006},
  publisher={Elsevier}
}

@article{eigen2013learning,
  title={Learning factored representations in a deep mixture of experts},
  author={Eigen, David and Ranzato, Marc'Aurelio and Sutskever, Ilya},
  journal={arXiv preprint arXiv:1312.4314},
  year={2013}
}

@article{zhong2024lory,
  title={Lory: Fully differentiable mixture-of-experts for autoregressive language model pre-training},
  author={Zhong, Zexuan and Xia, Mengzhou and Chen, Danqi and Lewis, Mike},
  journal={arXiv preprint arXiv:2405.03133},
  year={2024}
}

@article{muqeeth2023soft,
  title={Soft merging of experts with adaptive routing},
  author={Muqeeth, Mohammed and Liu, Haokun and Raffel, Colin},
  journal={arXiv preprint arXiv:2306.03745},
  year={2023}
}

@article{panda2025dense,
  title={Dense Backpropagation Improves Training for Sparse Mixture-of-Experts},
  author={Ashwinee Panda and Vatsal Baherwani and Zain Sarwar and Benjamin Th{\'e}rien and Supriyo Chakraborty and Tom Goldstein},
  journal={ArXiv},
  year={2025},
  volume={abs/2504.12463},
  url={https://api.semanticscholar.org/CorpusID:277856963}
}

@article{yang2024solving,
  title={Solving token gradient conflict in mixture-of-experts for large vision-language model},
  author={Yang, Longrong and Shen, Dong and Cai, Chaoxiang and Yang, Fan and Gao, Tingting and Zhang, Di and Li, Xi},
  journal={arXiv preprint arXiv:2406.19905},
  year={2024}
}

@article{nielsen2025tight,
  title={Tight clusters make specialized experts},
  author={Nielsen, Stefan K and Teo, Rachel SY and Abdullaev, Laziz U and Nguyen, Tan M},
  journal={arXiv preprint arXiv:2502.15315},
  year={2025}
}

@misc{xai2024grok1,
  title={Open Release of Grok-1},
  author={{xAI}},
  howpublished={\url{https://x.ai/news/grok-os}},
  note={314B-parameter Mixture-of-Experts model; open weights and architecture},
  year={2024}
}

@misc{meta2025llama4,
  title={The Llama 4 herd: The beginning of a new era of natively multimodal AI innovation},
  author={{Meta AI}},
  howpublished={\url{https://ai.meta.com/blog/llama-4-multimodal-intelligence/}},
  year={2025},
  note={Official blog post describing Llama~4’s MoE design}
}

@article{he2024upcycling,
  title={Upcycling Large Language Models into Mixture of Experts},
  author={He, Ethan and Khattar, Abhinav and Prenger, Ryan and Korthikanti, Vijay and Yan, Zijie and Liu, Tong and Fan, Shiqing and Aithal, Ashwath and Shoeybi, Mohammad and Catanzaro, Bryan},
  journal={arXiv preprint arXiv:2410.07524},
  year={2024}
}

@article{wu2024yuanm32,
  title={Yuan 2.0-M32: Mixture of Experts with Attention Router},
  author={Wu, Shaohua and Luo, Jiangang and Chen, Xi and Li, Lingjun and Zhao, Xudong and Yu, Tong and Wang, Chao and Wang, Yue and Wang, Fei and Qiao, Weixu and He, Houbo and Zhang, Zeru and Sun, Zeyu and Mao, Junxiong and Shen, Chong},
  journal={arXiv preprint arXiv:2405.17976},
  year={2024}
}

@article{park2024monet,
  title={Monet: Mixture of monosemantic experts for transformers},
  author={Park, Jungwoo and Ahn, Young Jin and Kim, Kee-Eung and Kang, Jaewoo},
  journal={arXiv preprint arXiv:2412.04139},
  year={2024}
}

@article{nguyen2025competesmoe,
  title={CompeteSMoE--Statistically Guaranteed Mixture of Experts Training via Competition},
  author={Nguyen, Nam V and Nguyen, Huy and Pham, Quang and Nguyen, Van and Ramasamy, Savitha and Ho, Nhat},
  journal={arXiv preprint arXiv:2505.13380},
  year={2025}
}

@article{he2024mixture,
  title={Mixture of a million experts},
  author={He, Xu Owen},
  journal={arXiv preprint arXiv:2407.04153},
  year={2024}
}

@inproceedings{rajbhandari2022deepspeed,
  title={Deepspeed-moe: Advancing mixture-of-experts inference and training to power next-generation ai scale},
  author={Rajbhandari, Samyam and Li, Conglong and Yao, Zhewei and Zhang, Minjia and Aminabadi, Reza Yazdani and Awan, Ammar Ahmad and Rasley, Jeff and He, Yuxiong},
  booktitle={International conference on machine learning},
  pages={18332--18346},
  year={2022},
  organization={PMLR}
}

@misc{qwen_moe,
    title = {Qwen1.5-MoE: Matching 7B Model Performance with 1/3 Activated Parameters"},
    url = {https://qwenlm.github.io/blog/qwen-moe/},
    author = {Qwen Team},
    month = {February},
    year = {2024}
}

@article{nakamura2025drop,
  title={Drop-Upcycling: Training sparse mixture of experts with partial re-initialization},
  author={Nakamura, Taishi and Akiba, Takuya and Fujii, Kazuki and Oda, Yusuke and Yokota, Rio and Suzuki, Jun},
  journal={arXiv preprint arXiv:2502.19261},
  year={2025}
}

@article{hui2024upcycling,
  title={Upcycling Instruction Tuning from Dense to Mixture-of-Experts via Parameter Merging},
  author={Hui, Tingfeng and Zhang, Zhenyu and Wang, Shuohuan and Sun, Yu and Wu, Hua and Su, Sen},
  journal={arXiv preprint arXiv:2410.01610},
  year={2024}
}

@article{chen2025automatic,
  title={Automatic Expert Discovery in LLM Upcycling via Sparse Interpolated Mixture-of-Experts},
  author={Chen, Shengzhuang and Wei, Ying and Schwarz, Jonathan Richard},
  journal={arXiv preprint arXiv:2506.12597},
  year={2025}
}

@article{shen2023scaling,
  title={Scaling vision-language models with sparse mixture of experts},
  author={Shen, Sheng and Yao, Zhewei and Li, Chunyuan and Darrell, Trevor and Keutzer, Kurt and He, Yuxiong},
  journal={arXiv preprint arXiv:2303.07226},
  year={2023}
}

@article{wu2024deepseek,
  title={Deepseek-vl2: Mixture-of-experts vision-language models for advanced multimodal understanding},
  author={Wu, Zhiyu and Chen, Xiaokang and Pan, Zizheng and Liu, Xingchao and Liu, Wen and Dai, Damai and Gao, Huazuo and Ma, Yiyang and Wu, Chengyue and Wang, Bingxuan and others},
  journal={arXiv preprint arXiv:2412.10302},
  year={2024}
}

@misc{dbrx_2024,
  title={DBRX: A Fine-Grained Mixture-of-Experts Open LLM},
  author={{Databricks}},
  year={2024},
  howpublished={Hugging Face model card and technical blog},
  note={132B total, 36B active, 16 experts choose 4, 12T tokens, 32K context},
  url={https://huggingface.co/databricks/dbrx-instruct}
}

@article{lin2024moma,
  title={Moma: Efficient early-fusion pre-training with mixture of modality-aware experts},
  author={Lin, Xi Victoria and Shrivastava, Akshat and Luo, Liang and Iyer, Srinivasan and Lewis, Mike and Ghosh, Gargi and Zettlemoyer, Luke and Aghajanyan, Armen},
  journal={arXiv preprint arXiv:2407.21770},
  year={2024}
}

@article{nguyen2024sigmoid,
  title={Sigmoid gating is more sample efficient than softmax gating in mixture of experts},
  author={Nguyen, Huy and Ho, Nhat and Rinaldo, Alessandro},
  journal={Advances in Neural Information Processing Systems},
  volume={37},
  pages={118357--118388},
  year={2024}
}

@article{yan2024understanding,
  title={Understanding expert structures on minimax parameter estimation in contaminated mixture of experts},
  author={Yan, Fanqi and Nguyen, Huy and Le, Dung and Akbarian, Pedram and Ho, Nhat},
  journal={arXiv preprint arXiv:2410.12258},
  year={2024}
}

@article{teo2024momentumsmoe,
  title={MomentumSMoe: Integrating momentum into sparse mixture of experts},
  author={Teo, Rachel S and Nguyen, Tan M},
  journal={Advances in Neural Information Processing Systems},
  volume={37},
  pages={28965--29000},
  year={2024}
}

@article{csordas2024moeut,
  title={Moeut: Mixture-of-experts universal transformers},
  author={Csord{\'a}s, R{\'o}bert and Irie, Kazuki and Schmidhuber, J{\"u}rgen and Potts, Christopher and Manning, Christopher D},
  journal={Advances in Neural Information Processing Systems},
  volume={37},
  pages={28589--28614},
  year={2024}
}

@article{teo2025molex,
  title={MoLEx: Mixture of Layer Experts for Finetuning with Sparse Upcycling},
  author={Teo, Rachel SY and Nguyen, Tan M},
  journal={arXiv preprint arXiv:2503.11144},
  year={2025}
}

@article{nguyen2025camex,
  title={Camex: Curvature-aware merging of experts},
  author={Nguyen, Dung V and Nguyen, Minh H and Nguyen, Luc Q and Teo, Rachel SY and Nguyen, Tan M and Tran, Linh Duy},
  journal={arXiv preprint arXiv:2502.18821},
  year={2025}
}

@article{fan2024_moe_design_choices,
  title={Towards an empirical understanding of MoE design choices},
  author={Fan, Dongyang and Messmer, Bettina and Jaggi, Martin},
  journal={arXiv preprint arXiv:2402.13089},
  year={2024}
}

@article{zhao2024_sparse_moe_generalization,
  title={Generalization Error Analysis for Sparse Mixture-of-Experts: A Preliminary Study},
  author={Zhao, Jinze and Wang, Peihao and Wang, Zhangyang},
  journal={arXiv preprint arXiv:2403.17404},
  year={2024}
}

@article{zhou2022_expert_choice_routing,
  title={Mixture-of-Experts with Expert Choice Routing},
  author={Zhou, Yanqi and Lei, Tao and Liu, Hanxiao and Du, Nan and Huang, Yanping and Zhao, Vincent and Dai, Andrew and Chen, Zhifeng and Le, Quoc and Laudon, James},
  journal={arXiv preprint arXiv:2202.09368},
  year={2022}
}

@article{hazimeh2021_dselectk,
  title={DSelect-k: Differentiable Selection in the Mixture of Experts with Applications to Multi-Task Learning},
  author={Hazimeh, Hussein and Zhao, Zhe and Chowdhery, Aakanksha and Sathiamoorthy, Maheswaran and Chen, Yihua and Mazumder, Rahul and Hong, Lichan and Chi, Ed H.},
  journal={arXiv preprint arXiv:2106.03760},
  year={2021}
}

@article{lewis2021_base_layers,
  title={BASE Layers: Simplifying Training of Large, Sparse Models},
  author={Lewis, Mike and Bhosale, Shruti and Dettmers, Tim and Goyal, Naman and Zettlemoyer, Luke},
  journal={arXiv preprint arXiv:2103.16716},
  year={2021}
}

@article{zeng2024_adamoe,
  title={AdaMoE: Token-Adaptive Routing with Null Experts for Mixture-of-Experts Language Models},
  author={Zeng, Zihao and Miao, Yibo and Gao, Hongcheng and Zhang, Hao and Deng, Zhijie},
  journal={arXiv preprint arXiv:2406.13233},
  year={2024}
}

@article{antoine2024_pos_router_moe,
  title={Part-Of-Speech Sensitivity of Routers in Mixture of Experts Models},
  author={Antoine, Elie and B{\'e}chet, Fr{\'e}d{\'e}ric and Langlais, Philippe},
  journal={arXiv preprint arXiv:2412.16971},
  year={2024}
}

@article{gale2022_megablocks,
  title={MegaBlocks: Efficient Sparse Training with Mixture-of-Experts},
  author={Gale, Trevor and Narayanan, Deepak and Young, Cliff and Zaharia, Matei},
  journal={arXiv preprint arXiv:2211.15841},
  year={2022}
}

@article{nguyen2024_cosine_router,
  title={Statistical Advantages of Perturbing Cosine Router in Mixture of Experts},
  author={Nguyen, Huy and Akbarian, Pedram and Pham, Trang and Nguyen, Trang and Zhang, Shujian and Ho, Nhat},
  journal={arXiv preprint arXiv:2405.14131},
  year={2024}
}
